\documentclass{article} 
\usepackage{iclr2027_conference,times}
\usepackage{hyperref}
\usepackage{url}
\usepackage{tikz}
\usepackage{multirow}
\usepackage{xspace}
\usepackage{booktabs}
\usepackage{ulem}
\usepackage{graphicx} 
\usepackage{fancyvrb}
\usepackage{fvextra}
\usepackage{physics}
\usepackage{wrapfig}
\usepackage{bbm}
\usepackage[flushleft]{threeparttable}
\usepackage{enumitem}
\usepackage{verbatim}
\usepackage{scalerel} 
\usepackage{enumitem}
\usepackage{ragged2e}
\usepackage{tabularx}
\usepackage{array}
\usepackage{pifont}
\usepackage{graphicx}
\usepackage{makecell}
\usepackage{amssymb}
\usepackage{wrapfig}
\usepackage{needspace}

\iclrfinalcopy

\hypersetup{linkcolor=bcnavy, citecolor=bcteal, urlcolor=bcteal, filecolor=bcteal}

\usepackage[table]{xcolor}
\usepackage{pgf}

\definecolor{scoreBlue}{HTML}{C9DDEB}
\definecolor{scoreMid}{HTML}{E7E7E7}
\definecolor{scoreOrange}{HTML}{F4D2B8}
\definecolor{scoreNA}{HTML}{F2F2F2}

\newcommand{\score}[1]{%
  \ifdim #1pt<50pt
    \pgfmathparse{100-2*(#1)}%
    \edef\tempcolor{%
      \noexpand\cellcolor{scoreBlue!\pgfmathresult!scoreMid}%
    }%
    \tempcolor
  \else
    \pgfmathparse{2*(#1-50)}%
    \edef\tempcolor{%
      \noexpand\cellcolor{scoreOrange!\pgfmathresult!scoreMid}%
    }%
    \tempcolor
  \fi
  #1%
}

\newcommand{\scoreNA}{%
  \cellcolor{scoreNA}\textcolor{gray!70}{N/A}%
}

\definecolor{metricSoundness}{HTML}{EAF2F8}
\definecolor{metricAffordance}{HTML}{EBF5EE}
\definecolor{metricDesign}{HTML}{FDF1E7}
\definecolor{metricRealization}{HTML}{F2EDF7}

\newcolumntype{C}[1]{%
  >{\centering\arraybackslash}p{#1}%
}

\definecolor{checkGreen}{HTML}{4F9D69}
\definecolor{crossRed}{HTML}{C95C5C}

\newcommand{\cmark}{\textcolor{checkGreen}{\ding{51}}}
\newcommand{\xmark}{\textcolor{crossRed}{\ding{55}}}

\usepackage[table]{xcolor}

\usepackage[most]{tcolorbox}

\newtcolorbox{Toolset}{
  breakable,
  colback=white,
  colframe=gray!40,
  fonttitle=\bfseries,
  title=Detailed Description for Tool Set Available in LMBuild,
}

\newtcolorbox{Harness}{
  breakable,
  colback=white,
  colframe=gray!30,
  fonttitle=\bfseries,
  title=Agent Harness Designed for LMBuild,
}

\newtcolorbox{DataExample_1}{
  breakable,
  colback=white,
  colframe=gray!40,
  fonttitle=\bfseries,
  title=Concrete Data Examples in LMBuild (1),
}

\newtcolorbox{DataExample_2}{
  breakable,
  colback=white,
  colframe=gray!50,
  fonttitle=\bfseries,
  title=Concrete Data Examples in LMBuild (2),
}
\newtcolorbox{DataExample_3}{
  breakable,
  colback=white,
  colframe=gray!40,
  fonttitle=\bfseries,
  title=Concrete Data Examples in LMBuild (3),
}
\newtcolorbox{DataExample_4}{
  breakable,
  colback=white,
  colframe=gray!40,
  fonttitle=\bfseries,
  title=Concrete Data Examples in LMBuild (4),
}
\newtcolorbox{DataExample_5}{
  breakable,
  colback=white,
  colframe=gray!40,
  fonttitle=\bfseries,
  title=Concrete Data Examples in LMBuild (5),
}
\newtcolorbox{DataExample_6}{
  breakable,
  colback=white,
  colframe=gray!40,
  fonttitle=\bfseries,
  title=Concrete Data Examples in LMBuild (6),
}

\newtcolorbox{Metrics_1}[1]{
  enhanced,
  breakable,
  colback=blue!1.5,
  colframe=blue!30,
  fonttitle=\bfseries,
  coltitle=black,
  title={Structural Soundness -- #1},
  boxrule=0.8pt,
  arc=1.5mm,
  left=2mm,
  right=2mm,
  top=1.5mm,
  bottom=1.5mm,
}

\newtcolorbox{Metrics_2}[1]{
  enhanced,
  breakable,
  colback=green!1.5,
  colframe=green!32,
  fonttitle=\bfseries,
  coltitle=black,
  title={Functional Affordance -- #1},
  boxrule=0.8pt,
  arc=1.5mm,
  left=2mm,
  right=2mm,
  top=1.5mm,
  bottom=1.5mm,
}

\newtcolorbox{Metrics_3}[1]{
  enhanced,
  breakable,
  colback=orange!1.5,
  colframe=orange!30,
  fonttitle=\bfseries,
  coltitle=black,
  title={Design Quality -- #1},
  boxrule=0.8pt,
  arc=1.5mm,
  left=2mm,
  right=2mm,
  top=1.5mm,
  bottom=1.5mm,
}

\newtcolorbox{Metrics_4}[1]{
  enhanced,
  breakable,
  colback=purple!1.5,
  colframe=purple!27,
  fonttitle=\bfseries,
  coltitle=black,
  title={Physical Realization -- #1},
  boxrule=0.8pt,
  arc=1.5mm,
  left=2mm,
  right=2mm,
  top=1.5mm,
  bottom=1.5mm,
}

\usepackage{amsmath,amsfonts,bm}

\def\eqref#1{equation~\ref{#1}}

\def\1{\bm{1}}

\DeclareMathAlphabet{\mathsfit}{\encodingdefault}{\sfdefault}{m}{sl}
\SetMathAlphabet{\mathsfit}{bold}{\encodingdefault}{\sfdefault}{bx}{n}

\NewDocumentCommand{\heng}
{ mO{} }{\textcolor{orange}{\textsuperscript{\textit{Cheng}}\textsf{\textbf{\small[#1]}}}}

\NewDocumentCommand{\cheng}
{ mO{} }{\textcolor{red}{\textsuperscript{\textit{Cheng}}\textsf{\textbf{\small[#1]}}}}

\definecolor{citegreenblue}{RGB}{0,120,130}

\hypersetup{
    colorlinks=true,
    citecolor=citegreenblue,
    linkcolor=citegreenblue,
    urlcolor=citegreenblue
}

\title{
\raisebox{-0.25\height}{\includegraphics[height=1.3em]{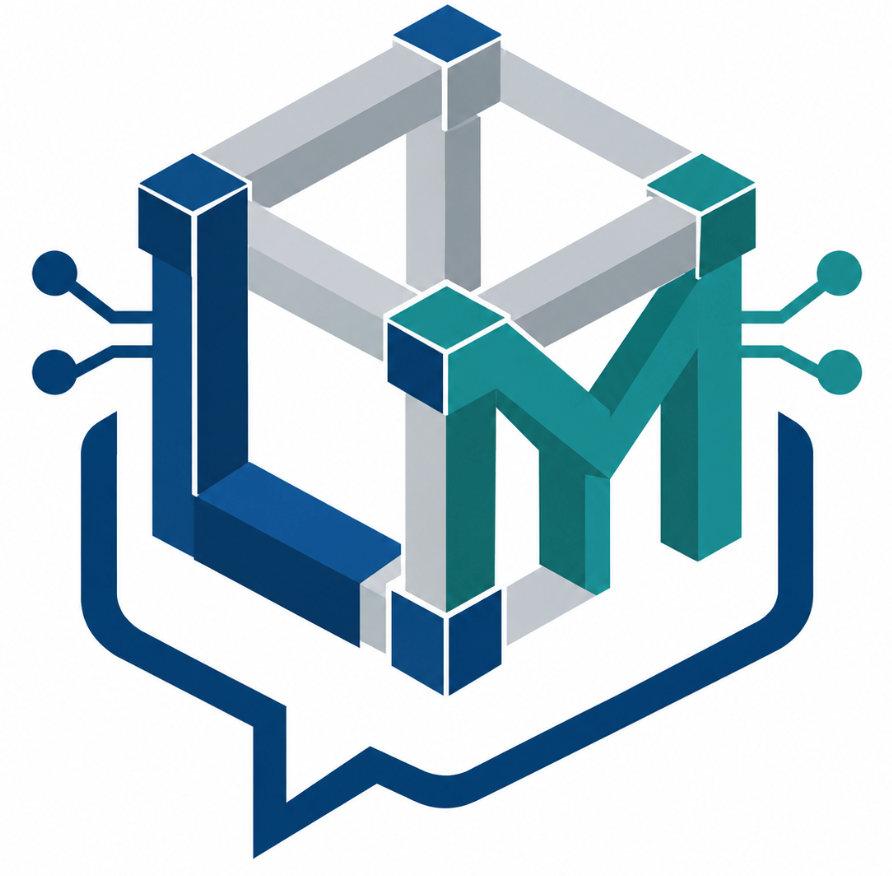}}
LMBuild: Evaluating LLM Agents for Generating Buildable and Functional Structures
}

\author{
\vspace{0.30in}
\makebox[\textwidth][c]{%
{\fontsize{11}{11.5}\selectfont
\begin{tabular}{c}
Jiateng Liu, Rushi Wang, Cheng Qian, Xuejun Zhang, Sun Li, Jiayu Liu, \\ [0.4em]
\textbf{Yifan Shen}, Xu Cao, Jiarui Yao, Bingxuan Li, Ruhi Sarikaya, Heng Ji \\ [0.6em]
University of Illinois Urbana-Champaign \\[0.5em]
{\normalfont\texttt{\{jiateng5,hengji\}@illinois.edu}} \\[0.5em]
{\normalfont \textbf{Project Page:} \url{https://lumos-jiateng.github.io/LMBuild/}}\\[0.5em]
{\normalfont \textbf{Github:} \url{https://github.com/Lumos-Jiateng/LMBuild}}\\[0.5em]
{\normalfont \textbf{Dataset:} \url{https://huggingface.co/datasets/Lumos-Jiateng/LMBuild}}
\end{tabular}
}%
}
}

\begin{document}

\maketitle

\begin{abstract}

Large Language Model (LLM)-based agents are increasingly capable of generating complex 3D structures, with the potential to reshape how objects are designed and realized in the physical world. Yet, producing elegant geometry is fundamentally different from producing objects that can actually be built and perform their intended functions. Existing evaluations largely focus on geometric or visual quality while overlooking structural soundness, functional affordances, and physical realizability. We introduce \textbf{LMBuild}, a benchmark for evaluating LLM agents on generating buildable and functional structures. LMBuild represents generated objects as assembled structures comprising meaningful part decompositions, joints, materials, and assembly sequences. To support reproducible evaluation, we provide a unified framework consisting of: (1) an interactive environment in which agents can use tools to retrieve, create, and place components to construct an object; (2) a curated benchmark that repurposes established CAD datasets and augments them with real-world object knowledge from Wikipedia; and (3) a four-level evaluation framework covering structural soundness, functional affordance, design quality, and physical realization. Evaluations across 30 domain-specific models, open-source LLMs, and frontier APIs reveal several intriguing findings: (a) Structural soundness and visual alignment are no longer the primary bottlenecks for frontier closed-source models, while functional affordance and physical operability remain substantially more challenging; (b) stronger models more effectively exploit part creation to construct structure-specific components, whereas weaker models tend to rely on retrieval and can struggle with the additional flexibility introduced by creation; and (c) providing explicit functional specifications substantially improves functional-part completeness, kinematic correctness, and overall physical operability. These results show that generating real-world structures requires capabilities beyond geometric synthesis, including deeper reasoning about functional affordances, mechanics, and physical constraints, as well as the ability to design and create novel components when existing ones are insufficient. We expect LMBuild to provide a foundation for measuring progress and incentivizing research toward agents that generate buildable and functional structures, while helping narrow the gap between open-source communities and frontier labs.

\end{abstract}

\section{Introduction}

The recent release of GPT-6 Astra~\citep{openai2026gpt6astra} has sparked an explosion of discussion around how much LLM-based agents could transform 3D structure generation. Beyond their established capabilities in language and code, LLM-based agents are increasingly able to directly manipulate complex 3D assets and scenes~\citep{hu2024scenecraft,sun20253dgpt,tang2025unrealllm,liu2026brickcomposer} and translate language-based specifications into parametric or programmatic 3D designs~\citep{li2025cadllama,wang2025cadfusion,alrashedy2025cadcodeverify}. These developments lead us to explore a more ambitious possibility: whether LLM agents could reshape not only how structures are digitally designed, but also how they are constructed and realized in the physical world~\citep{kyaw2025speech}.

However, producing visually plausible geometry is fundamentally different from producing an object that can actually be built and function in the real world. A structure may appear convincing while remaining invalid, unusable, or impossible to construct. Existing 3D-generation evaluations largely emphasize geometric quality~\citep{poole2023dreamfusion,lin2023magic3d}, visual consistency~\citep{wang2023prolificdreamer,maiti2025gen3deval,he2023t3bench,duggal2025eval3d}, and semantic alignment~\citep{wu2024gpt4veval,zhang2025multidimensional}, while overlooking critical aspects of physical realization. Yet, we still lack a systematic way to evaluate whether LLM-based agents can leverage their extensive prior knowledge and reasoning capabilities to generate structures that are not only digitally plausible, but also buildable, functional, and aligned with real-world physical constraints.

\begin{figure}[t]
\centering
\includegraphics[width=\linewidth]{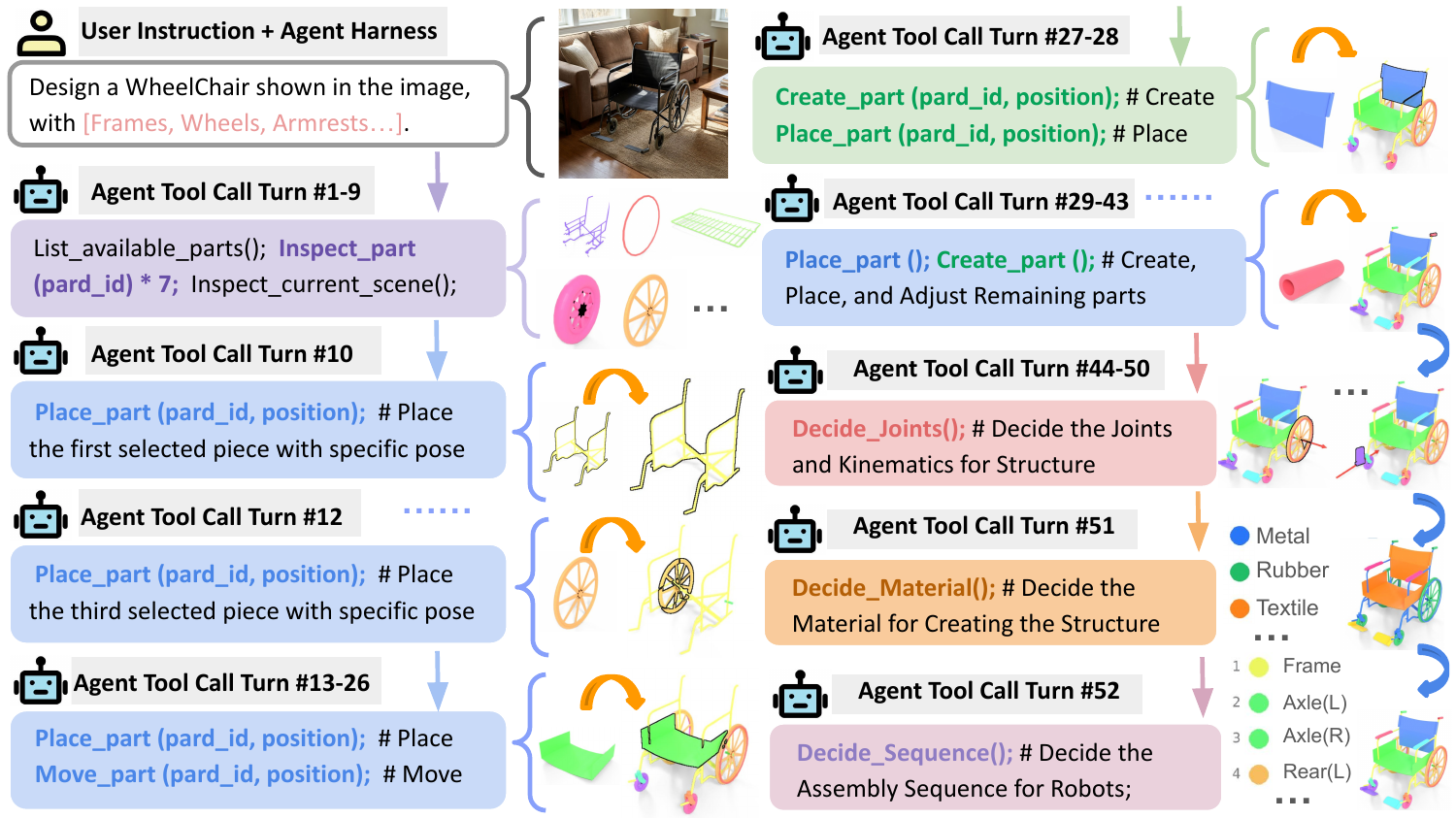}
\caption{\raisebox{-0.25em}{\includegraphics[height=1.3em]{Figures/icon.png}} \textbf{LMBuild Framework} (Using real example produced by GPT-6 Astra). Given a text prompt and conditional image, the builder LLM agent uses tools to inspect and select existing parts, create new parts, and place and adjust components to construct the  wheelchair. It then specifies joints, materials, and an assembly sequence to support kinematics and physical-world realization.}
\label{fig:_main}
\end{figure}

To this end, we propose \textbf{LMBuild}, a first step toward systematically evaluating LLM agents on the generation of buildable and functional structures, while also accommodating domain-specific structure-generation baselines. First, LMBuild provides a unified framework (Figure~\ref{fig:_main}) in which LLM agents are equipped with a standardized agent harness and can interact seamlessly with a 3D environment through a set of construction tools, including retrieving components from a sub-part pool, creating new parts with arbitrary geometry, placing components at desired poses, and refining structures through inspection and revision. Second, as summarized in Table~\ref{tab:data_stat}, we select 2,549 high-quality reference targets by repurposing existing 3D datasets containing around 368K data, filtering objects by category and model quality, and augmenting them with real-world knowledge extracted from Wikipedia, and further selected 200 objects as the core benchmark dataset for evaluation. This knowledge includes core functional parts, object attributes, joints, and kinematic relationships, which serve as textual anchors for evaluating physically grounded properties. Finally, as shown in Figure~\ref{fig:_main}, we introduce a bottom-up evaluation hierarchy covering four dimensions: (1) structural soundness, assessing whether components form valid and stable 3D assemblies; (2) functional affordance, evaluating whether functional parts, joints, and kinematics support the intended object functions; (3) design quality, assessing decomposition, aesthetics, and overall structural design; and (4) physical realization, evaluating whether the structure can be fabricated with appropriate materials and assembly sequences and ultimately operated or used in the physical world.

We evaluate 30 systems on LMBuild, including 6 closed-source APIs, 13 open-source LLMs, and 11 domain-specific systems, under diverse evaluation settings. Our results show that:

\begin{itemize}[noitemsep,topsep=0pt,parsep=0pt,partopsep=0pt,leftmargin=*]
\item Structural soundness and visual alignment are no longer primary bottlenecks for frontier closed-source models. Most achieve near-perfect soundness and strong design quality, although open-source models still show room for improvement in aesthetics, and fine-grained visual alignment.

\item Functional affordance and physical operability remain the largest capability gaps. All model families struggle with generating a complete list of functional parts, correct joints with kinematics, and artifacts that remain operable under simulation-based evaluation.

\item While frontier models can effectively create new components, weaker models rely heavily on retrieval and struggle with the additional reasoning burden introduced by part creation.

\item Enriching prompts with desired attributes and functional requirements can force the model to reason, thus improving functional-part completeness, kinematic correctness, and physical operability.
\end{itemize}

We further show that LMBuild can diagnose failure patterns for open-source LLM agents during structure construction, revealing their gap in part-creation capability as well as persistent challenges in scene-state tracking. Overall, we expect LMBuild to provide a foundation for measuring progress and incentivizing research toward agents capable of generating buildable and functional structures.

\section{
\raisebox{-0.20em}{\includegraphics[height=1.3em]{Figures/icon.png}} LMBuild
}

In this section, we describe the three core components of \textbf{LMBuild}: (1) how we construct an interactive environment for evaluating LLM agents and other systems on generating buildable and functional structures (\S~\ref{sec:2.1}); (2) how we repurpose existing datasets to support large-scale LMBuild evaluation and illustrate the interactive generation process with a concrete running example (\S~\ref{sec:2.2}); and (3) how we establish a hierarchical evaluation framework for the generated structures (\S~\ref{sec:2.3}).

\subsection{Interactive Generation of Buildable and Functional Structures}
\label{sec:2.1}

As shown in Figure~\ref{fig:_main}, LMBuild constructs an interactive environment in which agents are equipped with a manually designed harness (detailed in Appendix\ref{app:harness}) and instructed to follow user-provided construction specifications. The agents interact with a set of tools (detailed in Appendix\ref{app:environment}) to generate decomposed and buildable structures, determine appropriate joints and materials for individual components, and specify a feasible assembly sequence for real-world construction.

\paragraph{Baseline Interaction Protocol.}
LMBuild simulates a real-world setting in which users can ``speak structures into existence''~\citep{kyaw2025speech}. The user provides only a target structure label and a reference image, without geometric specifications, part-level instructions, or intermediate feedback. The agent then autonomously constructs the object in a single interaction round through tool calls (Figure~\ref{fig:_main}), either retrieving components from a candidate pool containing at least three times the required number of parts or creating new parts when needed. It selects or creates components, places and adjusts them, and specifies the final pose. For domain-specific systems, we preserve their native input and output protocols rather than introducing unsupported capabilities. When a system produces geometry without articulation information, we apply Particulate~\citep{li2026particulate} to infer part-level articulation, joints, and kinematic constraints for evaluation under LMBuild metrics.

\paragraph{Part Retrieval and Creation.}
The baseline allows agents to \textbf{either} retrieve existing components or create new ones when needed. To isolate the contribution of each mechanism, we additionally evaluate two constrained settings: \textbf{retrieval-only}, where agents must construct the target structure entirely from the provided component pool, and \textbf{creation-only}, where retrieval is disabled and every required component must be generated by the agent. Comparing these settings with the baseline reveals how much agents depend on existing parts versus their ability to design novel components.

\paragraph{Multi-Round Iterative Revision.}
LMBuild also evaluates whether agents can iteratively improve their constructions using feedback. After each round, the environment returns a six-view rendering and a rule-based diagnostic report covering collisions, disconnected components, stability, joint issues, and missing functional roles, materials, etc. The agent receives this feedback together with its previous interaction trajectory to  revises the current structure. We evaluate budgets of $K\in{1,2,3}$ rounds to measure how effectively agents use this feedback for iterative refinement.

\paragraph{Prompting with Fine-Grained Descriptions.}
Beyond the baseline category label and reference image, LMBuild evaluates whether richer task specifications improve structure generation. We augment the prompt with fine-grained descriptions of salient attributes and functional components, providing explicit guidance about what parts should be present and how they should behave or be operated. Comparison with the baseline reveals whether failures stem from insufficient specification of user intent or from limitations in translating detailed requirements into buildable structures.

\subsection{Repurposing Existing Datasets for LMBuild}
\label{sec:2.2}

{
\renewcommand{\arraystretch}{1.05}
\begin{table}[t!]
\centering
\caption{
Statistics of the source datasets repurposed for \textbf{LMBuild}, including original dataset size, objects retained in LMBuild-Core and LMBuild-Full, and average number of Wikipedia-augmented part and affordance annotations obtained. Affordances include functional attributes and kinematic information. Product-level CAD objects are complete open-source hardware assemblies from 27 projects, grouped by domain; We evaluate on the LMBuild-Core split (200 objects) in this work.
}

\vspace{1em}

\resizebox{\textwidth}{!}{
\begin{tabular}{@{} llrrrrr @{}}
\toprule
\multirow{2}{*}{\textbf{Geometry Type}}& \multirow{2}{*}{\textbf{Data Source}} & \multirow{2}{*}{\textbf{Original Size}} &
\multicolumn{2}{c}{\textbf{LMBuild-Core}} &
\multicolumn{2}{c}{\textbf{LMBuild-Full}} \\
\cmidrule(lr){4-5} \cmidrule(l){6-7}
 &
 &
 &
\textbf{Size} & \textbf{Avg. Parts / Aff.} &
\textbf{Size} & \textbf{Avg. Parts / Aff.} \\
\midrule

\multirow{2}{*}{\textbf{LEGO Toy Data}}
& BrickNet \citep{kulits2026bricknet}
& 320,808 & 5 & 13.80 / 14.60 & 888 & 8.79 / 10.82 \\

& BrickComposer \citep{liu2026brickcomposer}
& 9,977 & 50 & 10.16 / 10.34 & 50 & 10.16 / 10.34 \\

\midrule

\textbf{Static Structures}
& PartNeXt \citep{wang2025partnext}
& 23,519 & 14 & 5.07 / 8.00 & 589 & 6.00 / 7.92 \\

\midrule

\textbf{Mechanical Structures}
& Fusion 360 Gallery (Joinable) \citep{willis2022joinable}
& 8,251 & 55 & 6.85 / 8.91 & 80 & 6.20 / 8.41 \\

\midrule

\textbf{Daily Objects / Furniture}
& Artiverse \citep{iliash2026artiverse}
& 5,402 & 42 & 6.50 / 8.57 & 908 & 5.90 / 7.43 \\

\midrule

\multirow{4}{*}{\shortstack[l]{\textbf{Product-level CAD}\\\textbf{(open-source hardware)}}}
& Robotics \& vehicles \citep[][etc.]{cominelli2025astra}
& 7 & 7 & 8.29 / 9.29 & 7 & 8.29 / 9.29 \\

& Consumer devices \citep[][etc.]{ploopyElectronics}
& 10 & 10 & 6.40 / 7.80 & 10 & 6.40 / 7.80 \\

& Lab \& medical instruments \citep[][etc.]{hisakawa2024ninjapcr}
& 7 & 7 & 5.71 / 8.43 & 7 & 5.71 / 8.43 \\

& Fabrication machines \& tools \citep[][etc.]{voron2025voron24}
& 10 & 10 & 6.30 / 9.60 & 10 & 6.30 / 9.60 \\

\midrule

\textbf{Total in LMBuild}
&
& 367,991 & 200 & 7.62 / 9.25 & 2,549 & 7.03 / 8.83 \\

\bottomrule
\end{tabular}
}

\vspace{-5pt}
\label{tab:data_stat}
\end{table}
}

Evaluating LLM agents for structure generation is substantially more expensive than standard conversational evaluation. Even a simple object with fewer than 20 parts can require over 50 interactions in one construction round, with costs increasing further for complex structures and multi-round revision. Exhaustive evaluation over large 3D datasets is therefore impractical, potentially requiring millions turns of LLM generations. From around 368K candidate 3D objects, we therefore construct a full benchmark of 2,549 objects and a core set of 200 objects through stratified selection from complementary sources, as summarized in Table~\ref{tab:data_stat}. Rather than matching the frequency distribution of any individual source dataset, our selection is \textbf{coverage-oriented}: we prioritize objects that span diverse construction regimes and capabilities, have verifiable real-world references, and remain tractable for comprehensive evaluation. Due to budget constraints, all experiments in this work are conducted on the core set, while the full benchmark will be made publicly available.

\paragraph{Complementary Geometry Sources.}
No single existing dataset covers the full range of structures and construction regimes targeted by LMBuild. As summarized in Table~\ref{tab:data_stat}, we therefore combine several complementary sources. \textbf{BrickNet}~\citep{kulits2026bricknet} and \textbf{BrickComposer}~\citep{liu2026brickcomposer} provide LEGO-style structures assembled from reusable part inventories with standardized connections. \textbf{PartNeXt}~\citep{wang2025partnext} contributes fine-grained part-based objects and static structures, while the Fusion 360 Gallery assembly dataset from \textbf{JoinABLe}~\citep{willis2022joinable} provides engineering CAD assemblies with explicit part and joint structure.
\textbf{Artiverse}~\citep{iliash2026artiverse} contributes physically grounded articulated everyday objects with functional parts and kinematic annotations. Finally, we also collect product-level CAD from 27 open-source hardware projects, with each project publishes a complete, buildable product as a single assembly file. They span four domains:  (1) \textbf{Robotics and Vehicles}, including the ASTRA spinal quadruped~\citep{cominelli2025astra}, the Hubert robotic arm~\citep{gildeaHubert}, the BiDexHand dexterous hand~\citep{weng2025bidexhand}, the OpenROV underwater vehicle~\citep{openrov2020hardware}, the Ardumower lawn mower~\citep{ardumower2022}, the OpenAMR delivery robot~\citep{openamr2026} and the UPSat CubeSat~\citep{lsf2017upsat};
\textbf{Consumer Devices}, including Ploopy input and audio devices~\citep{ploopyElectronics}, the Framework desktop and keyboard~\citep{framework2026desktop,framework2026keyboard}, a DIY VR headset~\citep{dmcke52025diyvr}, a force-feedback racing pedal~\citep{chrgri2026ffbpedal}, the OLSK coffee grinder~\citep{ingrassia2023grinder}, the Shuffle-o-matic card shuffler~\citep{degonge2022shuffleomatic} and a Curta mechanical calculator~\citep{wu2026curta};
\textbf{Lab and Medical Instruments}, including the OpenRAMAN spectrometer~\citep{openraman2023}, a biaxial stretcher for live microscopy~\citep{shiwarski2019biaxial}, the ShearView rheometer~\citep{somexlab2026shearview}, the NinjaPCR thermal cycler~\citep{hisakawa2024ninjapcr}, the PolyVent ventilator~\citep{bechard2024polyvent}, a multi-channel peristaltic pump~\citep{buchhorn2025peristaltic} and the OpenAstroTracker star tracker~\citep{openastrotech2025tracker};
and \textbf{Fabrication Machines and Tools}, including the midTbot pen plotter~\citep{dring2020midtbot}, the Precious Plastic recycling machines~\citep{preciousplastic2025kit}, the OpenSlider camera slider~\citep{adamslab2021openslider}, the OLSK 3D scanner and vinyl cutter~\citep{olsk2025scanner,olsk2025vinyl}, the Voron~2.4 3D printer~\citep{voron2025voron24}, an automatic wire stripper~\citep{projectswithred2023stripper} and the Open Source Ecology CEB press~\citep{ose2026ceb}.
Together, these sources span both assembly from reusable part inventories and settings that require structure-specific part creation, and cover a broad range of real-world structures, from toys and furniture to complete machines.

\paragraph{Data Selection Strategy.}
Within the combined source pool, we stratify the benchmark data selection along four dimensions.
First, we ensure structural and kinematic diversity across six broad families, spanning static load-bearing structures, hinged mechanisms, sliding mechanisms, rotating systems, serial kinematic chains, and multi-joint articulated mechanisms.
Second, we prioritize grounded and recognizable objects whose expected components, physical properties, and functional behaviors can be verified from external Wikipedia data; each object category appears at most once in our core set. Third, we maintain difficulty coverage from simple assemblies to complex articulated systems by diversifying the number of parts, and require at least four parts per object. Fourth, we prioritize product-level CAD. These assemblies are the closest proxy for the end goal of LMBuild, building objects that work as real products, but they are orders of magnitude scarcer than part-annotated datasets. We therefore retain every product assembly that passes our screening (a complete single-file assembly, an open license, and a recognizable object category). The concrete data selection process and example data are provided in Appendix~\ref{app:data}.

\begin{figure}[t]
\centering
\includegraphics[width=\linewidth]{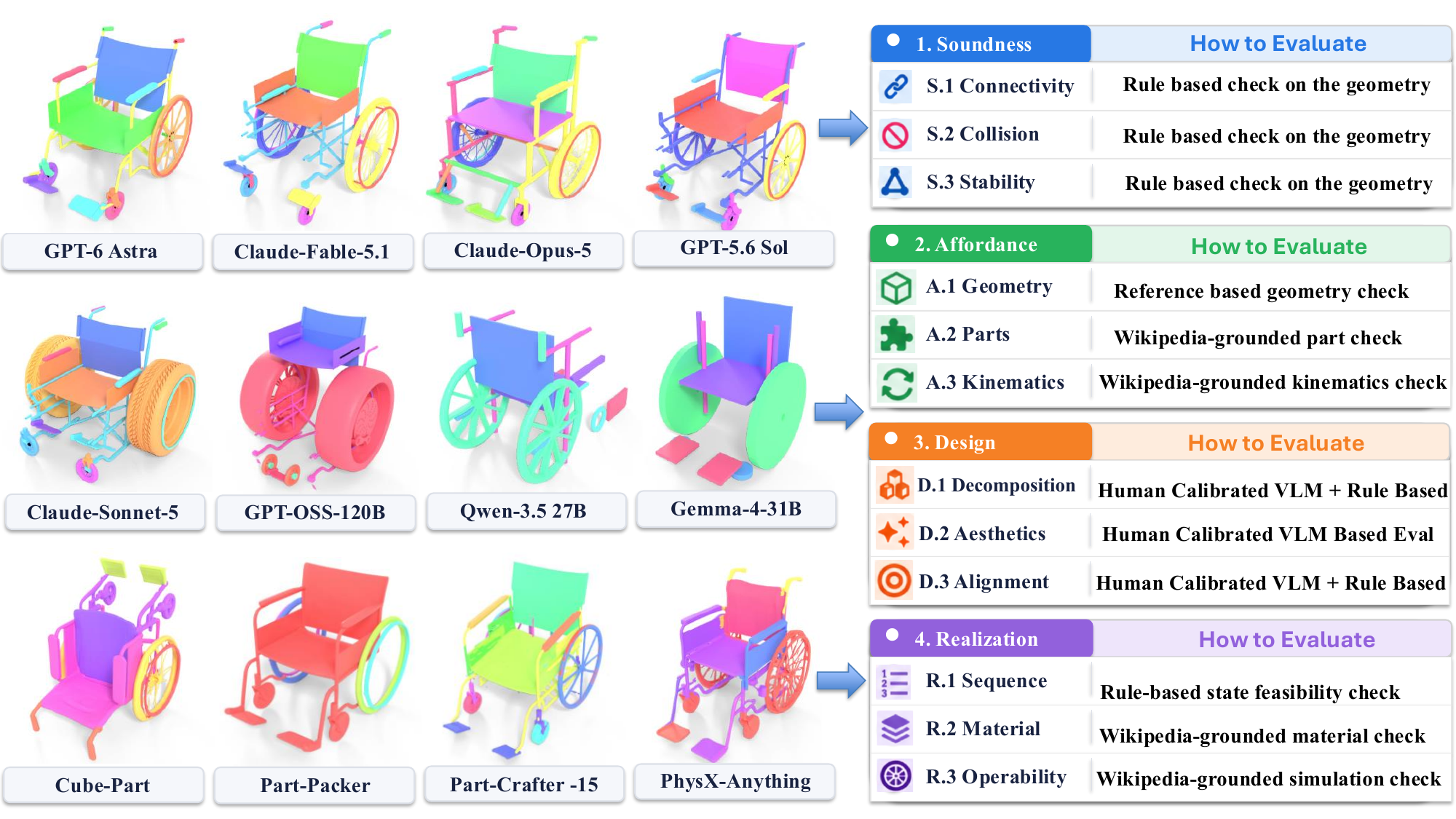}
\caption{\textbf{Left:} Generation examples produced by the LMBuild framework. Colors are used only to visually distinguish decomposed components. \textbf{Right:} The generated structures are evaluated using our comprehensive set of evaluation metrics. We briefly illustrate the reference source and general evaluation procedure for each metric; full evaluation details are provided in Appendix~\ref{app:metrics}.
}
\label{fig:example}
\end{figure}


\subsection{Hierarchical Evaluation Framework for LMBuild}
\label{sec:2.3}

Traditional evaluations of generated structures primarily textbfasize geometric quality or semantic alignment. In contrast, LMBuild evaluates whether a generated structure is buildable and functional through a hierarchical framework with four complementary dimensions. \textbf{Structural Soundness (S)} assesses whether the generated geometry is physically plausible and structurally stable, avoiding impossible configurations or severe geometric inconsistencies. \textbf{Functional Affordance (A)} evaluates whether the object exhibits the appropriate overall shape, contains the functional parts required by the target category, and specifies joints and kinematics consistent with the intended motion and functionality. \textbf{Design Quality (D)} measures whether the generated structure is appropriately decomposed into meaningful components and whether its overall form and appearance align with the reference structure and human design preferences. \textbf{Physical Realization (R)} evaluates whether the structure can be instantiated in the physical world, including whether it admits a feasible robotic assembly sequence, provides sufficient material information for fabrication, and remains operable with respect to its intended functionality. As shown in Figure~\ref{fig:example} (Right), these four dimensions are further decomposed into twelve concrete sub-metrics, together with their general corresponding evaluation strategies. To aid presentation, all scores are normalized to a common $0$--$100$ scale. To avoid overloading the main text with algorithmic details, we provide the complete evaluation criteria and implementation protocols in Appendix~\ref{app:metrics}, with illustrative examples in Table~\ref{tab:evaluation_vehicle_example}. 


\section{Experiments}

\subsection{Setup}

\paragraph{Evaluated Systems.}
We evaluate 30 systems spanning proprietary APIs, open-weight foundation models, and domain-specific generators, covering diverse model families, generations, scales, and architectures. Proprietary systems include GPT-6 Astra~\citep{openai2026gpt6astra}, GPT-5.6 Sol~\citep{openai2026gpt56}, Claude Opus 5~\citep{anthropic2026opus5}, Claude Fable 5.1~\citep{anthropic2026fable51}, Claude Sonnet 5~\citep{anthropic2026sonnet5}, and Claude Haiku 4.5~\citep{anthropic2025haiku45}. Open-weight models include six Qwen variants across Qwen3, Qwen3-VL, and Qwen3.5~\citep{yang2025qwen3,bai2025qwen3vl,qwen2026qwen35}, spanning 4B--35B parameters and both dense and MoE architectures (e.g., 30B-A3B and 35B-A3B), together with GPT-OSS-120B~\citep{openai2025gptoss}, Gemma-4-31B-IT~\citep{gemmateam2026gemma4}, Ministral-3-8B~\citep{mistralai2025mistral3}, InternVL3.5-8B/38B~\citep{wang2025internvl35}, MiniCPM-V-4.5~\citep{yu2025minicpmv45}, and ERNIE-4.5-VL-28B-A3B~\citep{ernie2025technicalreport}. We further compare with domain-specific systems for brick assembly, part-aware generation, and simulation-ready assets, including BrickGPT~\citep{pun2025brickgpt}, LegoACE~\citep{xu2025legoace}, PartCrafter~\citep{lin2025partcrafter}, PartPacker~\citep{tang2025partpacker}, Cube3D and CubePart~\citep{roblox2025cube,zhu2026cubepart}, and PhysX-Anything~\citep{cao2025physxanything}. Selected part-based generators are additionally augmented with Particulate for joint specification (Table~\ref{tab:system_capability_comparison}).

{
\renewcommand{\arraystretch}{1.18}

\begin{table*}[t]
\centering
\caption{Evaluation results across 30 systems under Baseline Interaction Protocol (\S~\ref{sec:2.1}) on core set, assessed using our hierarchical evaluation metrics (\S~\ref{sec:2.3}). We use $^{\dagger}$ (P) to denote systems integrated with Particulate to obtain joint information (Appendix~\ref{app:environment}). Background colors encode performance on a common 0--100 scale, transitioning from blue (lower performance) through gray to orange (higher performance). Systems unable to produce the required output are assigned as N/A.}
\label{tab:main_results}

\resizebox{\textwidth}{!}{
\begin{tabular}{l|ccc|ccc|ccc|ccc}
\toprule

\multirow{2}{*}{\textbf{Models / Metrics}}
& \multicolumn{3}{c|}{\textbf{Soundness}}
& \multicolumn{3}{c|}{\textbf{Affordance}}
& \multicolumn{3}{c|}{\textbf{Design}}
& \multicolumn{3}{c}{\textbf{Realization}} \\

\cmidrule(lr){2-4}
\cmidrule(lr){5-7}
\cmidrule(lr){8-10}
\cmidrule(lr){11-13}

& \textbf{S.1} & \textbf{S.2} & \textbf{S.3}
& \textbf{A.1} & \textbf{A.2} & \textbf{A.3}
& \textbf{D.1} & \textbf{D.2} & \textbf{D.3}
& \textbf{R.1} & \textbf{R.2} & \textbf{R.3} \\

\midrule

\rowcolor{gray!20}
\multicolumn{13}{c}{\textbf{State-of-the-art Closed-Source APIs}} \\

GPT-6 Astra
& \score{83.0} & \score{95.9} & \score{86.7}
& \score{63.0} & \score{45.3} & \score{41.7}
& \score{76.0} & \score{71.2} & \score{74.9}
& \score{60.9} & \score{80.4} & \score{19.2} \\

GPT-5.6 Sol
& \score{85.7} & \score{95.6} & \score{85.0}
& \score{48.9} & \score{38.8} & \score{38.0}
& \score{77.6} & \score{70.7} & \score{66.1}
& \score{65.8} & \score{87.5} & \score{19.1} \\

Claude Opus 5
& \score{86.8} & \score{95.7} & \score{84.0}
& \score{52.1} & \score{44.9} & \score{42.7}
& \score{76.4} & \score{69.3} & \score{68.2}
& \score{58.7} & \score{84.4} & \score{22.1} \\

Claude Fable 5.1
& \score{82.4} & \score{97.4} & \score{87.5}
& \score{55.3} & \score{44.2} & \score{42.5}
& \score{75.2} & \score{69.6} & \score{70.5}
& \score{69.0} & \score{76.1} & \score{20.8} \\

Claude Sonnet 5
& \score{80.5} & \score{95.2} & \score{82.5}
& \score{47.1} & \score{38.0} & \score{39.3}
& \score{74.6} & \score{60.1} & \score{64.2}
& \score{68.5} & \score{82.4} & \score{17.7} \\

Claude Haiku 4.5
& \score{81.5} & \score{90.3} & \score{70.6}
& \score{36.7} & \score{31.4} & \score{23.3}
& \score{69.3} & \score{49.2} & \score{54.9}
& \score{58.9} & \score{84.0} & \score{11.0} \\

\midrule

\rowcolor{gray!20}
\multicolumn{13}{c}{\textbf{Open-Source (Multi-modal) Large Language Models}} \\

Qwen3.5-27B
& \score{86.7} & \score{83.4} & \score{78.5}
& \score{39.5} & \score{28.8} & \score{16.7}
& \score{66.1} & \score{40.9} & \score{51.1}
& \score{74.7} & \score{86.4} & \score{10.5} \\

Qwen3.5-35B-A3B
& \score{74.5} & \score{79.3} & \score{62.5}
& \score{31.8} & \score{27.2} & \score{13.2}
& \score{60.1} & \score{33.5} & \score{47.5}
& \score{47.3} & \score{78.6} & \score{8.9} \\

GPT-OSS-120B
& \score{80.5} & \score{88.2} & \score{66.8}
& \score{28.9} & \score{27.1} & \score{21.3}
& \score{60.5} & \score{45.8} & \score{41.7}
& \score{60.1} & \score{85.4} & \score{9.6} \\

Gemma-4-31B-IT
& \score{57.7} & \score{62.2} & \score{59.2}
& \score{30.3} & \score{19.9} & \score{10.1}
& \score{47.7} & \score{34.9} & \score{40.9}
& \score{36.0} & \score{66.1} & \score{7.9} \\

Ministral-3-8B
& \score{77.0} & \score{75.7} & \score{44.5}
& \score{25.1} & \score{25.1} & \score{6.1}
& \score{55.9} & \score{41.0} & \score{43.5}
& \score{33.0} & \score{81.3} & \score{5.3} \\

InternVL3.5-38B
& \score{45.2} & \score{58.2} & \score{44.6}
& \score{19.9} & \score{11.4} & \score{2.8}
& \score{39.3} & \score{32.5} & \score{32.5}
& \score{8.5} & \score{52.5} & \score{3.6} \\

Qwen3-VL-30B-A3B
& \score{60.0} & \score{51.2} & \score{34.9}
& \score{20.1} & \score{17.2} & \score{4.2}
& \score{39.3} & \score{24.4} & \score{32.5}
& \score{35.4} & \score{58.8} & \score{4.6} \\

MiniCPM-V-4.5
& \score{45.7} & \score{42.3} & \score{32.0}
& \score{17.2} & \score{10.3} & \score{0.8}
& \score{33.8} & \score{24.7} & \score{22.8}
& \score{4.4} & \score{49.3} & \score{2.2} \\

Qwen3-4B-Instruct
& \score{81.6} & \score{52.3} & \score{62.7}
& \score{23.0} & \score{18.5} & \score{5.4}
& \score{37.6} & \score{22.9} & \score{30.4}
& \score{36.2} & \score{74.4} & \score{5.7} \\

Qwen3-VL-32B
& \score{52.3} & \score{49.8} & \score{43.3}
& \score{19.5} & \score{16.1} & \score{4.8}
& \score{36.1} & \score{24.6} & \score{29.8}
& \score{33.5} & \score{55.7} & \score{6.3} \\

ERNIE-4.5-VL-28B-A3B
& \score{22.7} & \score{20.7} & \score{16.6}
& \score{6.9} & \score{4.9} & \score{1.3}
& \score{17.6} & \score{12.6} & \score{14.8}
& \score{4.2} & \score{22.2} & \score{2.5} \\

InternVL3.5-8B
& \score{38.9} & \score{37.3} & \score{41.1}
& \score{14.4} & \score{7.1} & \score{0.7}
& \score{27.5} & \score{25.2} & \score{25.7}
& \score{0.5} & \score{51.8} & \score{2.0} \\

Qwen3-VL-8B
& \score{32.0} & \score{24.4} & \score{27.3}
& \score{9.4} & \score{8.3} & \score{2.9}
& \score{17.3} & \score{13.2} & \score{14.7}
& \score{13.6} & \score{36.0} & \score{3.9} \\

\midrule

\rowcolor{gray!20}
\multicolumn{13}{c}{\textbf{State-of-the-art Domain-Specific Models}} \\

BrickGPT
& \score{99.6} & \score{100.0} & \score{84.5}
& \score{35.5} & \score{33.5} & \scoreNA
& \score{27.6} & \score{13.0} & \score{32.1}
& \scoreNA & \scoreNA & \scoreNA \\

LegoACE
& \score{92.4} & \score{97.1} & \score{44.4}
& \score{29.9} & \score{29.3} & \scoreNA
& \score{28.1} & \score{18.0} & \score{36.9}
& \scoreNA & \scoreNA & \scoreNA \\

PartCrafter-15
& \score{90.9} & \score{78.6} & \score{14.3}
& \score{45.6} & \score{18.8} & \scoreNA
& \score{63.1} & \score{44.1} & \score{65.3}
& \scoreNA & \scoreNA & \scoreNA \\

PartCrafter-8
& \score{73.2} & \score{84.7} & \score{13.4}
& \score{42.6} & \score{13.7} & \scoreNA
& \score{58.3} & \score{38.0} & \score{62.6}
& \scoreNA & \scoreNA & \scoreNA \\

PartPacker
& \score{73.2} & \score{87.4} & \score{64.9}
& \score{52.5} & \score{20.2} & \scoreNA
& \score{55.5} & \score{61.0} & \score{67.3}
& \scoreNA & \scoreNA & \scoreNA \\

Cube3D+CubePart
& \score{92.7} & \score{77.7} & \score{37.7}
& \score{24.7} & \score{32.7} & \scoreNA
& \score{54.0} & \score{63.6} & \score{51.8}
& \scoreNA & \scoreNA & \score{13.4} \\

PartCrafter15$^{\dagger}$(P)
& \score{81.7} & \score{91.0} & \score{10.8}
& \score{45.4} & \score{6.4} & \score{2.3}
& \score{33.4} & \score{26.1} & \score{52.8}
& \scoreNA & \scoreNA & \scoreNA \\

PartCrafter8$^{\dagger}$(P)
& \score{72.1} & \score{95.1} & \score{13.5}
& \score{43.2} & \score{6.6} & \score{2.3}
& \score{36.4} & \score{33.3} & \score{59.6}
& \scoreNA & \scoreNA & \scoreNA \\

PartPacker$^{\dagger}$(P)
& \score{73.4} & \score{95.5} & \score{64.8}
& \score{52.3} & \score{9.1} & \score{3.6}
& \score{42.8} & \score{56.5} & \score{65.6}
& \scoreNA & \scoreNA & \scoreNA \\

Cube3D$^{\dagger}$(P)
& \score{69.1} & \score{96.5} & \score{73.9}
& \score{40.3} & \score{8.9} & \score{3.8}
& \score{43.1} & \score{82.2} & \score{57.7}
& \scoreNA & \scoreNA & \scoreNA \\

PhysX-Anything
& \score{52.3} & \score{88.9} & \score{49.8}
& \score{20.0} & \score{8.7} & \score{5.1}
& \score{44.9} & \score{48.3} & \score{40.8}
& \scoreNA & \score{61.4} & \score{1.9} \\

\bottomrule
\end{tabular}
}
\end{table*}
}

\subsection{Measuring Model Capabilities for Functional Structure Generation}

\paragraph{Overall Observations.}
As shown in Table~\ref{tab:main_results}, state-of-the-art closed-source LLMs achieve the strongest overall performance across most metrics, although several open-source and domain-specific systems remain competitive and occasionally surpass them on individual geometric (e.g., Cube3D$^{\dagger}$(P) on \textbf{D.2}) or realization metrics (e.g., Qwen3.5 models on \textbf{R.1}). Notably, frontier closed-source models, including GPT-6 Astra, GPT-5.6 Sol, Claude Opus 5, and Claude Fable 5.1, exhibit broadly similar performance across many evaluation dimensions, suggesting that leading model families are converging toward a comparable capability level for generating 3D structures. More broadly, overall performance is associated with stronger general reasoning capabilities, while the competitive performance of several domain-specific systems suggests that targeted training and architectural specialization can partially compensate for weaker general reasoning on structure generation. Note that although several systems achieve high performance along individual dimensions, strong performance on a single metric does not necessarily translate into a high-quality structure overall. Detailed interpretations of these results are provided in Appendix~\ref{app:interpretation}.

\paragraph{Structural Soundness and Visual Alignment Are No Longer the Primary Bottlenecks}
Across Structural Soundness, most systems perform reliably in generating interconnected (\textbf{S.1}) and collision-free (\textbf{S.2}) structures, although performance declines somewhat on structural stability (\textbf{S.3}), particularly for domain-specific generators. Similarly, frontier closed-source models already perform strongly in producing reasonable decompositions (\textbf{D.1}), aesthetically appealing designs (\textbf{D.2}), and reference-aligned structures (\textbf{D.3}), suggesting that structural plausibility and visual quality are no longer their primary bottlenecks. Several open-source and domain-specific systems are also competitive on individual design metrics, but a noticeable gap remains overall, leaving room for further improvement in generating structures with better aesthetic quality, stronger structural stability, and more fine-grained visual alignment.

\paragraph{Functional Affordance and Physical Operability Remain the Largest Capability Gaps.}
Even the strongest systems remain imperfect in generating appropriate functional geometry (\textbf{A.1}) and complete functional parts (\textbf{A.2}), while joints and kinematics (\textbf{A.3}) are particularly challenging: most open-source models score below 25, and even the strongest systems only slightly exceed 50. These limitations carry directly into physical realization. Although models are comparatively more capable of specifying plausible assembly sequences (\textbf{R.1}) and selecting materials (\textbf{R.2}), simulation-based operability (\textbf{R.3}) remains low across all systems, with the best scores reaching only the mid-30s. Together, these results suggest that generating a structurally and visually aligned artifact is substantially easier than producing the functional geometry, complete parts, joint configurations, and kinematics required for the artifact to be fabricated and operate correctly in the physical world.

\begin{figure}[t]
\centering
\includegraphics[width=\linewidth]{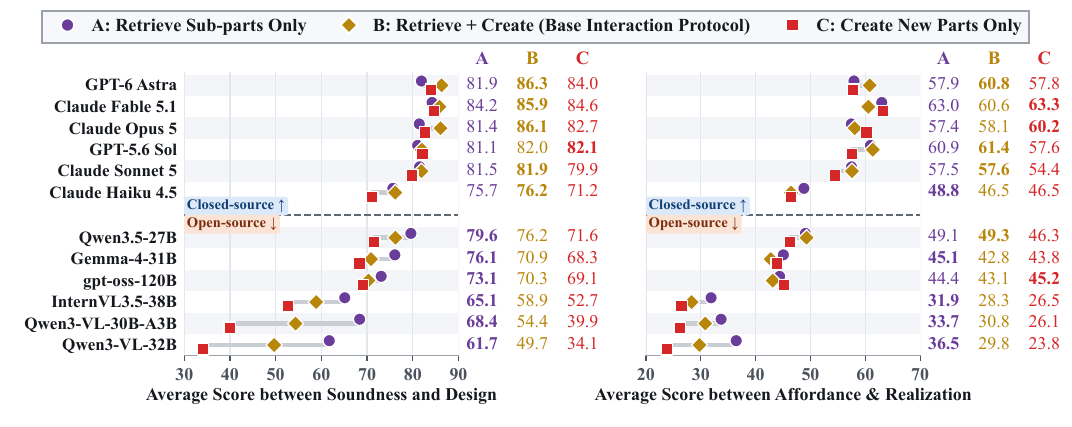}

\vspace{-1em}
\caption{\textbf{Model Performance Change} across evaluation metrics under three part-access settings with 25 sampled LMBuild objects. Allowing part creation can hurt weaker models, even when catalog parts remain available, suggesting that the added design freedom increases decision complexity. In contrast, frontier models benefit substantially from creating structure-specific parts, revealing a stronger ability to exploit flexible construction mechanisms rather than relying only on retrieval.}

\label{fig:_create_vs_retrieve}
\end{figure}

\subsection{Strong Models Create, While Weaker Models Retrieve}

We further evaluate LLM agents under three interaction protocols: our baseline setting with both retrieval and part creation, a retrieval-only setting, and a creation-only setting. As shown in Figure~\ref{fig:_create_vs_retrieve}, model behavior differs substantially across capability levels. Less capable models degrade sharply when required to create all parts themselves. For example, Qwen3-VL-32B achieves an average score of 61.7 across Soundness and Design metrics under retrieval, but drops to 34.1 under creation-only. Even simply introducing part creation alongside retrieval can reduce performance for weaker models, suggesting that additional design freedom also introduces a substantially larger reasoning and decision-making burden. In contrast, frontier models exhibit the opposite trend. They more frequently choose to create their own components when both tools are available, and maintain comparable or even stronger performance when retrieval is removed entirely. For these models, retrieval-only can instead become the more restrictive setting, as the available component pool may constrain their ability to construct geometry tailored to the target structure. This reveals an intriguing capability transition: retrieval helps weaker models by reducing the design space, whereas creation allows stronger models to exploit their generative and structural reasoning capabilities. Part creation therefore appears not merely as an additional tool, but as a capability that increasingly becomes beneficial as the underlying model becomes stronger.

\begin{figure}[t]
\centering
\includegraphics[width=\linewidth]{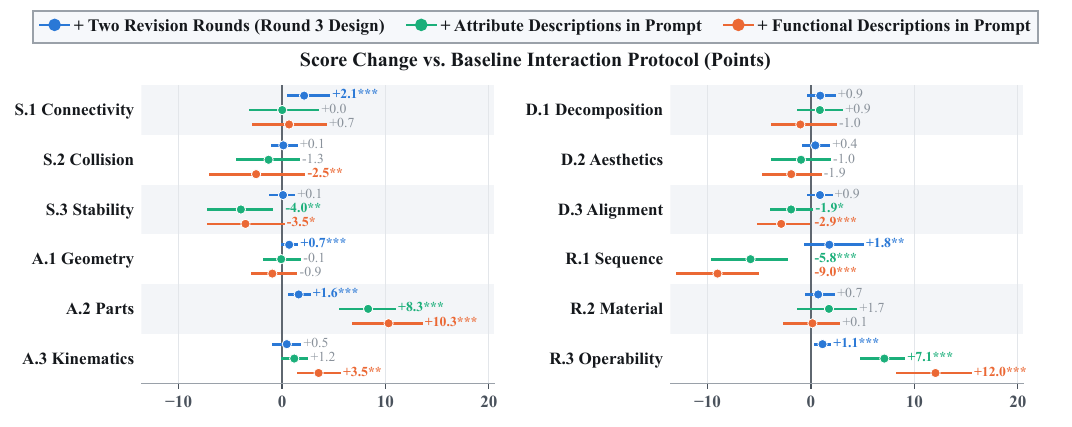}

\vspace{-1em}
\caption{\textbf{Average performance change} across models from (1) additional revision rounds, (2) explicit attribute requirements, and (3) explicit functional requirements. Revision yields small but consistent gains, while explicit prompts substantially improve functional-part completeness, kinematics, and especially operability. \textbf{Bars:} 95\% bootstrap confidence intervals over models; \textbf{Stars:} Wilcoxon signed-rank test over episodes (*$p<0.05$, ** $p<0.01$, *** $p<0.001$)}

\label{fig:_ablation}
\end{figure}

\subsection{Improving Capabilities with Iterative Revision and Prompting}

As shown in Figure~\ref{fig:_ablation}, we evaluate two common strategies for improving buildable and functional structure generation: (1) allowing models to inspect their outputs, receive error reports, and iteratively revise the structure; and (2) enriching the target description with explicit functional attributes and requirements to encourage deeper reasoning about the underlying structure. 

\paragraph{Iterative Revision Provides Only Modest Gains.} Allowing models to inspect and revise their structures over three rounds leads to steady but limited improvement, with overall gains of only about 2\%. Because agents can already interact with and inspect the environment under the baseline protocol, additional revision rounds mainly provide more opportunities to retry rather than addressing the underlying reasoning limitations required for functional structure generation.

\paragraph{Explicit Functional Specifications Produce Substantial Improvements.}
In contrast, enriching a minimal prompt with explicit attributes and functional requirements grounded in Wikipedia produces substantially larger gains on the dimensions where current models struggle most. Functional-part completeness improves by 10.3 points and physical
operability by 12.0 points ($p<0.001$), with additional improvements in
kinematic correctness. These results indicate that a substantial fraction of baseline failures can be reduced when functional requirements are made explicit. Current agents therefore appear considerably more reliable at translating specified functional constraints into structure than at inferring all such requirements from a minimal object label alone.

\paragraph{Implication.}
Together, these results suggest that improving task specification can be more effective than increasing the revision rounds.
A promising direction is therefore to develop agents that can more reliably infer, retrieve, or elicit the functional requirements of an object before construction, reducing the need for users to manually provide detailed part-level and kinematic specifications.

\subsection{LMBuild Identifies Failure Patterns of Evaluated LLMs}

\paragraph{Open Models Exhibit a Part-Creation Gap.}
Across Part Creation calls, closed-source models succeed on 95.7\% of attempts, compared with only 51.7\% for open-source models, with Qwen3.5-27B (89.4\%) as the only close exception. Most failures fall into two recurring categories. First, models violate CAD-kernel conventions, such as listing polygon vertices clockwise. Second, models generate invalid program structures by nesting nodes instead of referencing named nodes, using numeric node IDs, or referring to undefined nodes. These errors suggest that much of the creation gap stems from weak adherence to tool-specific conventions and program structure, rather than purely from deficiencies in geometric or spatial reasoning.

\paragraph{Open Models Struggle with Scene-State Tracking.}
Beyond part creation, open-source models frequently lose track of the evolving scene state. They refer to parts that were never placed at a rate of 7.2 errors per 100 calls, compared with 0.3 for closed-source models, they often attempt to attach, join, or assign properties to parts whose creation had already failed. They also redundantly add existing entities, including repeatedly placing parts instead of moving them and re-adding existing joints, at 3.2 errors per 100 calls versus 0.5 for closed-source models. Additional interface errors include invalid material names, malformed arguments, exhausted inspection calls, and even invoking part names as tools. Together, these failures reveal a broader limitation in maintaining and updating an accurate internal representation of the environment throughout long-horizon construction.

\section{Related Work}

\paragraph{3D Generation with Language Models.}
Language models increasingly support natural-language 3D creation by generating procedural scripts and scene blueprints~\citep{hu2024scenecraft,sun20253dgpt,tang2025unrealllm} or parametric CAD operation sequences~\citep{khan2024text2cad,li2025cadllama}. Visual feedback further enables training and iterative refinement of generated geometry~\citep{huang2024blenderalchemy,wang2025cadfusion,alrashedy2025cadcodeverify}. These methods connect language understanding with programmatic 3D modeling, but generally focus on producing editable or visually correct geometry rather than structures that can be physically assembled and function as intended.

\paragraph{Part-Aware Generation and Physical Assembly.}
Recent work represents objects as structured collections of parts~\citep{lin2025partcrafter,tang2025partpacker}, predicts joints and articulation~\citep{willis2022joinable,li2026particulate}, and incorporates physical attributes such as scale, materials, and motion~\citep{le2025articulate,cao2025physx,cao2025physxanything}. Complementary approaches target stable brick construction, assembly programs, component placement, and robotic execution~\citep{pun2025generatingphysicallystablebuildable,kulits2026bricknet,liu2026brickcomposer,goldberg2025bloxnet}. Together, these works motivate evaluating part structure, physical behavior, and whether the completed object can realize its intended function.

\paragraph{Evaluation of 3D Generation.}
Existing 3D evaluation primarily measures visual quality and text alignment~\citep{he2023t3bench}, often extended with multimodal judgments of appearance, surface quality, and human preference~\citep{wu2024gpt4veval,maiti2025gen3deval}, or finer-grained geometric and semantic consistency checks~\citep{duggal2025eval3d}. For assembled structures, prior work additionally evaluates stability, component selection, placement, and construction success~\citep{pun2025generatingphysicallystablebuildable,liu2026brickcomposer,goldberg2025bloxnet}. LMBuild complements these efforts by jointly evaluating soundness, affordance, design, and physical realization across heterogeneous objects using geometric checks, grounded component assessments, simulation-based motion tests, and VLM judgments.

\section{Conclusion}

In this work, we introduce LMBuild, a benchmark for evaluating LLM-based agents on the generation of structured 3D assemblies across four dimensions: structural soundness, functional affordance, design quality, and physical realization. On the LMBuild-Core benchmark, the evaluated systems perform substantially better on structural and visual criteria than on functional-part completeness, kinematic correctness, and simulation-based operability. These results highlight persistent challenges in translating object specifications into coherent components, joints, and assembly plans, as well as in creating task-specific parts when suitable components are unavailable. We hope LMBuild provides a reproducible foundation for studying these capabilities and for developing agents that generate increasingly functional and physically plausible structures. Bridging these benchmark results to real-world fabrication and operation will require further validation involving manufacturing constraints, physical assembly, and functional testing.

\subsection*{AI Use Statement}

We used large language models (LLMs) as writing assistants to improve the clarity, grammar, and presentation of the manuscript. We also used AI-based coding agents to assist with evaluation-environment construction, code refactoring, debugging, code cleaning, and code publishing. All research decisions, experimental designs, implementations, analyses, and reported results were reviewed and verified by the authors. AI tools were not used to  determine the scientific conclusions of this work.

\subsection*{Ethics Statement}

This work aims to support more transparent, accessible, and reproducible evaluation of LLM agents for real-world structure generation. By developing an open evaluation framework, we hope to lower the barriers to systematically assessing model capabilities beyond purely textual benchmarks. In particular, our benchmark emphasizes whether generated structures are not only plausible in appearance but also buildable and functionally valid under real-world constraints. We believe that open evaluation infrastructure can facilitate more rigorous comparison across models, encourage reproducible research, and contribute to the broader open-source AI ecosystem. At the same time, benchmarks can only capture a subset of the requirements involved in deploying AI-generated designs in real-world settings. Performance on LMBuild should therefore not be interpreted as sufficient evidence that a model is ready for safety-critical engineering, construction, or other physical-world applications without appropriate domain-specific validation and human oversight.

\subsection*{Reproducibility Statement}

We provide representative data samples and evaluation materials in the supplementary materials to facilitate inspection and reproduction of our methodology. We will open-source the complete codebase, data, evaluation environment, evaluation scripts, model trajectories from the reported experiments, and all the necessary artifacts required to reproduce our results. We aim to make the full evaluation pipeline transparent and executable so that future work can independently verify our findings, evaluate additional models, and extend LMBuild for constructing more powerful agents.

\bibliography{iclr2027_conference}
\bibliographystyle{iclr2027_conference}

\appendix

\section{Appendix}

\definecolor{HarnessBlue}{RGB}{239,246,255}
\definecolor{HarnessBlueFrame}{RGB}{69,113,181}
\definecolor{HarnessGreen}{RGB}{241,250,244}
\definecolor{HarnessGreenFrame}{RGB}{62,140,86}
\definecolor{HarnessOrange}{RGB}{255,247,235}
\definecolor{HarnessOrangeFrame}{RGB}{205,126,42}
\definecolor{HarnessGray}{RGB}{247,247,247}
\definecolor{HarnessGrayFrame}{RGB}{125,125,125}

\newtcolorbox{PromptBox}[1][]{
    enhanced,
    breakable,
    colback=HarnessBlue,
    colframe=HarnessBlueFrame,
    boxrule=0.7pt,
    arc=2pt,
    title=#1,
    fonttitle=\bfseries,
    left=5pt,
    right=5pt,
    top=5pt,
    bottom=5pt
}

\newtcolorbox{ProtocolBox}[1][]{
    enhanced,
    breakable,
    colback=HarnessGreen,
    colframe=HarnessGreenFrame,
    boxrule=0.7pt,
    arc=2pt,
    title=#1,
    fonttitle=\bfseries,
    left=5pt,
    right=5pt,
    top=5pt,
    bottom=5pt
}

\newtcolorbox{NoteBox}[1][]{
    enhanced,
    breakable,
    colback=HarnessOrange,
    colframe=HarnessOrangeFrame,
    boxrule=0.7pt,
    arc=2pt,
    title=#1,
    fonttitle=\bfseries,
    left=5pt,
    right=5pt,
    top=5pt,
    bottom=5pt
}

\subsection{Agent Harness for LMBuild}
\label{app:harness}

\paragraph{Overview.}
LMBuild exposes a common interactive assembly environment to heterogeneous model families through a lightweight model-specific harness. General-purpose LLM (VLM) agents receive the same environment semantics, task specification, tool definitions, action budgets, and multi-round revision protocol. The harness changes only the mechanism by which observations are delivered and actions are returned: closed-source agents interact with the environment through shell commands, whereas locally served open-source models use an OpenAI-compatible chat interface with text-serialized tool calls. Domain-specific structure-generation systems do not support the common tool interface and are therefore evaluated using their native generation pipelines. This separation allows LMBuild to preserve a common task and environment while accommodating substantial differences in model APIs. In particular, the closed- and open-source agent settings expose equivalent task information, although image transport, tool-call serialization, and conversation-state management necessarily differ across interfaces.

\paragraph{Environment tiers.}
All general-purpose agents are evaluated in one of three environment specifications. We refer to them as Tier A, Tier B, and Tier C for simplicity.
\textbf{Tier A} permits retrieval and placement of catalog parts only.
\textbf{Tier B}, our primary baseline interaction protocol, provides the same catalog while additionally allowing the agent to create a new component when no suitable catalog part is available. \textbf{Tier C} removes the catalog entirely and requires the agent to create every component before placing it. Except for the corresponding catalog and part-creation instructions, the environment prompt and tool semantics are unchanged across tiers. The environment enforces a maximum of 160 tool calls per episode, including at most 6 calls to \texttt{check}, 4 calls to \texttt{render}, and 80 placed parts. Part creation is limited to 30 calls in Tier B and 60 calls in Tier C. Both \texttt{check} and \texttt{render} are explicitly described to the agent as metered operations so that they are used for verification rather than unrestricted search.

\paragraph{Multi-round interaction.}
An episode may contain multiple revision rounds. Calling \texttt{submit} ends
the current round. When another revision round remains, the environment returns a structured diagnostic produced by \texttt{check} together with condition, isometric, front, and side-view images of the current assembly. The agent is asked to inspect this feedback, revise the same scene, and submit again. A final \texttt{submit} returns \texttt{episode\_done=true}. The assembly state is persistent across rounds; therefore, revisions modify the existing object rather than restarting generation.

\subsubsection{Shared LMBuild Agent Prompt}
\label{app:harness_shared_prompt}

The following system prompt is shared by all general-purpose LLM/VLM agents.
The placeholder \texttt{\{PARTS\_SOURCE\}} is instantiated according to the
tier, after which LMBuild appends the tier-specific part-creation instructions
and the generated tool documentation.

\begin{PromptBox}[Shared environment system prompt]
\begin{Verbatim}[fontsize=\scriptsize,breaklines=true,breakanywhere=true]
You assemble a physical object out of 3D parts by calling tools. The result is judged as a real, buildable, working product: it must hold together, stand, look like what was asked for, move where the real object moves, be made of sensible materials, and be buildable in the order you give.

WORLD FRAME
- Units are metres. +Z is up. The ground is the plane z = 0. The FRONT of the object faces -Y (a person sitting on a chair faces -Y). Standing in front of the object (at -Y) and looking at it, +X is to your right.
- Real-world size matters. Build the object at its real size.

PARTS
{PARTS_SOURCE}
{TOOLS_DOCUMENTATION}
- Every part's local origin is the centre of its bounding box. place_part puts that centre at `position` and applies `rotation` = [rx, ry, rz] degrees about the world X, then Y, then Z axes. So an unrotated part placed at z = size_z / 2 rests exactly on the ground.
- attach moves a part so that one face centre of its bounding box (top, bottom, left, right, front, back, center) lands on a face centre of another part's box, plus an optional offset. It is the easiest way to stack parts exactly: attach(child, "bottom", parent, "top").
- Parts count as connected when their real surfaces touch (gap under about 4 mm), or when a joint joins them at a place where both parts are: the joint's origin within 2 cm of each part's surface. A joint whose origin is far from either part connects nothing. Parts that interpenetrate a lot collide. Anything not connected to the rest is floating.

DECLARATIONS (all are scored)
- role: say what each part is, in plain words ("seat", "backrest", "caster wheel", ...).
- material: choose one per part from list_materials.
- joints: declare every place the real object moves. add_joint(joint_id, type, parent, child, axis, origin, limits). type is one of fixed, revolute, continuous, prismatic, cylindrical, ball. axis is a direction in world coordinates; origin is a world point on the axis where parent and child meet;
limits = [low, high] in radians (revolute) or metres (prismatic, cylindrical = the sliding range); omit limits for continuous. The child moves relative to the parent, carrying whatever is attached to the child.
- assembly sequence: set_assembly_sequence with every part once, in the order a person would build the object, each step optionally with the direction the part travels as it is put in place (e.g. [0, 0, -1] for lowering it from above).

ROUNDS AND BUDGETS
- submit ends a round. If review rounds remain you are shown a check of the design and renders, and you may revise it and submit again. After the final round the episode ends.
- check and render are metered; use them to verify, not to search. Tool calls are budgeted.

Keep reasoning brief. Prefer making tool calls over explaining.
\end{Verbatim}
\end{PromptBox}


\paragraph{Task prompt.}
The user message specifies the target category, optional reference image,
optional fine-grained attributes and functional constraints, and the current
environment tier. For example, the main Tier-B multimodal condition for an
office chair is:

\begin{PromptBox}[Example user prompt: Tier B with reference image]
\begin{Verbatim}[fontsize=\scriptsize,breaklines=true]
TASK: According to image W2, build an office chair. The condition image W2 is attached.
Tier B: Use catalog parts, and create new parts where the catalog has none that fit. A picture of every catalog part, labelled with id, name and size, is attached. view_catalog_sheet shows it again. Start with list_parts and list_materials.
\end{Verbatim}
\end{PromptBox}

For name-only conditions, the first line is simply ``Build an office chair.''
For image-conditioned tasks, the corresponding image identifier is prepended.
Attribute-conditioned tasks append descriptions of required components,
materials, and other object properties. Functional conditions additionally
provide subsystem and kinematic constraints, such as specifying that an office
chair seat rotates relative to its base about a vertical axis. Tier A instructs
the model to use catalog parts only, while Tier C states that no catalog exists,
requires every placed component to be created, and begins with
\texttt{list\_materials} followed by \texttt{create\_part}. For models without visual capability, image-conditioned conditions are preserved rather than silently converted to text: the message explicitly states that the task refers to an image that the model cannot see. This allows the same benchmark condition to be retained while exposing the model's actual input limitation.

\subsubsection{Closed-Source Agent Harness}
\label{app:harness_closed}

\paragraph{Claude models.}
Claude Haiku 4.5, Sonnet 5, Opus 5, and Fable 5.1 are evaluated through the Claude command-line interface. Each episode is placed in an isolated working directory and the model is given only the session interface and the image files returned by that session. The agent first requests the complete LMBuild prompt and subsequently invokes environment tools through shell commands of the form 

\begin{Verbatim}[fontsize=\scriptsize,breaklines=true] 

.venv_eval/bin/python -m ppbench.v2.session call --state <STATE> <TOOL_NAME> '<JSON_ARGUMENTS>'

\end{Verbatim}

Each invocation returns one JSON object. Tools that produce images return PNG paths, which Claude may inspect using its image-reading capability. The outer wrapper is:

\begin{PromptBox}[Claude shell wrapper]
\begin{Verbatim}[fontsize=\scriptsize,breaklines=true,breakanywhere=true]
You are the model being evaluated in a 3D assembly benchmark. Work only through the session command below; do not edit, create or delete any files, and do not run any other programs except reading image files.

Your session state file: {state}

1. Run: {cmd} prompt --state {state}
   Read the whole output: it is your task, your tools and how to call them. Open every image file it names with your file-reading tool.
   
2. Build the object with the tools. Tool calls look like:
   {cmd} call --state {state} <tool_name> '<JSON arguments>'
   Each call prints one JSON result; image results are PNG paths you open to look at.
   
3. The episode may have several rounds: each submit ends a round and returns a review. Keep working until a submit result says "episode_done": true. If a call returns an error, read it and correct the call. 

Reply at the end with one line: the number of parts and joints in your final design.
\end{Verbatim}
\end{PromptBox}

\paragraph{GPT models.}
GPT-5.6 Sol and GPT-6 Astra use the same session environment but are driven
through codex. Their outer wrapper is slightly more restrictive
with respect to filesystem access:

\begin{PromptBox}[GPT shell wrapper]
\begin{Verbatim}[fontsize=\scriptsize,breaklines=true,breakanywhere=true]
You are the model being evaluated in a 3D assembly benchmark. Work only through the session command below. Do not edit, create, or delete files directly. Do not inspect the reference assembly, evaluator data, other results, or repository code. You may read only the condition image, catalog sheets, and renders named by your session.

Your session state file is {state}.

Run `{cmd} prompt --state {state}` first and read its full output. Open the image files it names. Build the object using only session tool calls such as `{cmd} call --state {state} <tool_name> '<JSON arguments>'`. Each successful submit saves a round and returns feedback. Continue revising the same design until a submit returns `episode_done: true`. If the saved session is already partly complete, call get_scene and continue from that state; never start over. Inspect your renders and correct errors before submitting. At the end, report the final part and joint counts.
\end{Verbatim}
\end{PromptBox}

Thus, Claude and GPT receive the same LMBuild environment prompt and tool
interface; their outer wrappers differ only in provider-specific execution and
isolation instructions. GPT episodes permit up to three execution attempts.

\subsubsection{Open-Source LLM/VLM Agent Harness}
\label{app:harness_open}

Open-source LLMs and VLMs are served using vLLM through an OpenAI-compatible
chat endpoint. This group includes Qwen3-VL, Qwen3.5, Gemma-4,
InternVL3.5, MiniCPM-V, GPT-OSS, Ministral-3, ERNIE-4.5-VL,
Llama-3.1-8B, etc. Unlike the shell-based agents, these models
interact with LMBuild through an in-process agent loop.

The system message is constructed as

\[
\texttt{system\_prompt(tier)}
\;+\;
\texttt{tool\_docs(tier)}
\;+\;
\texttt{tool-call protocol},
\]

while the user message contains the task text, the condition image when the
model supports vision, and the catalog sheet for Tier A/B. Condition images
are resized to a maximum side length of 1024 pixels, while catalog sheets are
resized to a maximum side length of 1600 pixels. Because the shared vLLM deployment does not require model-specific native tool parsers, LMBuild uses the following text protocol for all locally served agents:

\begin{ProtocolBox}[Open-source tool-call protocol]
\begin{Verbatim}[fontsize=\scriptsize,breaklines=true,breakanywhere=true]
Write each tool call as a block on its own: <tool_call>{"name": "place_part", "arguments": {"part_id": "P001", "instance_id": "wheel_1", "position": [0, 0, 0.03]}}</tool_call>

You may write several blocks in one reply; they run in order and you then see
every result. Arguments are strict JSON (double quotes, no comments, no trailing commas).
\end{Verbatim}
\end{ProtocolBox}

Tool outputs are returned as user messages using the form

\begin{Verbatim}[fontsize=\scriptsize,breaklines=true]
<tool_result name="TOOL_NAME">{...JSON result...}</tool_result>
\end{Verbatim}

The parser first searches for explicit \texttt{<tool\_call>} blocks and then supports a small number of model-specific fallback formats, including
ERNIE-style \texttt{<name>}/\texttt{<arguments>} outputs, standalone or fenced JSON, and function-like expressions. Automatic repair is deliberately limited to evaluating simple arithmetic expressions in numeric arguments and closing truncated JSON structures. When parsing fails, the model receives the concrete parsing error and is instructed to resend a strict-JSON tool call. If a model produces no environment action, it is reminded that the interaction contains no human operator and that it should continue through tools or call \texttt{submit} once the object is complete. An episode is terminated after
three consecutive responses containing no executable tool call. All locally served models use temperature $0.6$ and random seed 0. Model-specific serving modifications are used only when required by the underlying model implementation.

\subsubsection{Domain-Specific Generation Systems}
\label{app:harness_domain}

Domain-specific structure generators are not forced into the LMBuild agent
interface when their native architectures do not support iterative tool use.
Instead, each system receives the subset of the benchmark condition supported
by its original inference pipeline and performs a single generation. We retain
each method's native representation and prompting procedure rather than adding
capabilities unavailable to the original system. For these methods, Tier B is operationally identical to Tier A because their native generation pipelines do not expose LMBuild's optional
\texttt{create\_part} action.

\begin{table*}[t]
\centering
\caption{Comparison of domain-specific structure-generation systems and LLM agents in terms of the information received and produced throughout the construction pipeline. While LLM agents interact seamlessly with LMBuild, domain-specific systems cover only a subset.}
\label{tab:system_capability_comparison}

\renewcommand{\arraystretch}{1.18}
\setlength{\tabcolsep}{4pt}

\resizebox{\textwidth}{!}{
\begin{tabular}{
    C{0.45cm}
    p{2.7cm}
    p{5.0cm}
    C{1.4cm}
    C{1.3cm}
    C{1.0cm}
    C{1.0cm}
    C{1.1cm}
    C{1.35cm}
    C{1.35cm}
}
\toprule

\textbf{\#} &
\textbf{System} &
\textbf{Pipeline} &
\textbf{Input} &
\textbf{Geometry} &
\textbf{Parts} &
\textbf{Roles} &
\textbf{Joints} &
\textbf{Materials} &
\textbf{Sequence} \\

\midrule

1 & BrickGPT & Generates LEGO Brick Structures & Text &
\cmark & \cmark & \xmark & \xmark & \xmark & \xmark \\

2 & LegoACE & Generates LEGO Brick Structures & Text &
\cmark & \cmark & \xmark & \xmark & \xmark & \xmark \\

3 & PartCrafter-15 & Generates a mesh of 15 parts. & Image &
\cmark & \cmark & \xmark & \xmark & \xmark & \xmark \\

4 & PartCrafter-8 & Generates a mesh of 8 parts. & Image &
\cmark & \cmark & \xmark & \xmark & \xmark & \xmark \\

5 & PartPacker & Generates part meshes. & Image &
\cmark & \cmark & \xmark & \xmark & \xmark & \xmark \\

6 & Cube3D+CubePart & Generate geometry and decompose & Text &
\cmark & \cmark & \cmark$^{*}$ & \xmark & \xmark & \xmark \\

7 & Particulate +(3) & Re-segment PartCrafter-15 Mesh & Image &
\cmark & \cmark & \xmark & \cmark & \xmark & \xmark \\

8 & Particulate +(4) & Re-segment PartCrafter-8 Mesh & Image &
\cmark & \cmark & \xmark & \cmark & \xmark & \xmark \\

9 & Particulate +(5) & Re-segment PartPacker Mesh & Image &
\cmark & \cmark & \xmark & \cmark & \xmark & \xmark \\

10 & Particulate +(6) & Re-segment Cube-3D Mesh & Text &
\cmark & \cmark & \xmark & \cmark & \xmark & \xmark \\

11 & PhysX-Anything & Generation aligned with Physics & Image &
\cmark & \cmark & \cmark & \cmark & \cmark & \xmark \\

\midrule

-- & \textbf{LLM Agents (19)} & \textbf{Interact with LMBuild} & \textbf{Multimodal} &
\cmark & \cmark & \cmark & \cmark & \cmark & \cmark \\

\bottomrule
\end{tabular}
}

\end{table*}

\subsection{Environment Details for LMBuild}
\label{app:environment}

\paragraph{Implementation details of the LMBuild environment.} LMBuild is a stateful, pure-Python tool environment. Entire episode state coming from LMBuild (placed instances with poses, created-part programs, joints, assembly sequence, counters and a full call trace) is treated as one JSON-serialisable dictionary, so the same episode can be driven in-process or from a shell, and it can be checkpointed after every round and replayed exactly.

\paragraph{World frame and parts} All geometry of LMBuild lives in one metric world frame: $+Z$ up, ground plane at $z = 0$, object back facing $Y$. Catalog parts are triangle meshes taken from segmented reference assets and stored as glTF. On loading, trimesh converts each mesh with the defined convention and re-centers it on its bounding box. Therefore reduce to exact rigid transforms: a translation plus $XYZ$ Euler rotations, or face-centered alignment of bounding boxes.

\textbf{Part creation} The part creation tool accepts a declarative CSG program. The program is a DAG of at most 96 nodes over box, cylinder, sphere, torus, extrude and revolve primitives. These can be combined with union, difference, intersection and hull, and moved with transform and mirror. The program is compiled with manifold3d (v3.5), which guarantees watertight 2-manifold output. Every created part is therefore a valid solid with an exact volume. Parameters are range-checked (0.5 mm–3 m), curved primitives are tessellated with 48 segments, and parts that fall apart into disconnected pieces are flagged.

\textbf{Physical feedback} All physical feedback comes from one geometry kernel, built on NumPy and SciPy, that is shared with the evaluator. Each placed part is voxelised on a 4-mm grid: area-proportional surface sampling, then scipy gives a solid occupancy grid.

\textbf{Material Definitions} Declared materials are restricted to 23 submaterials in 9 classes. Their densities are medians over 5,510 Artiverse annotations.

\textbf{Rendering} Renders are produced out-of-process by a headless Blender (bpy 5.2 LTS) worker. The posed scene is exported as a GLB, with each part in a distinct hue so that parts can be told apart. Each image is rendered with Cycles on the CPU (448 px, 16 samples with denoising, AgX view transform) under three sun lights and above a ground plane. The camera has a 50 mm lens and auto-framing from the design's bounding sphere. Six fixed (azimuth, elevation) viewpoints are available, including one matching the condition image's camera. The rendered views are tiled with PIL. The catalog contact sheet is pre-rendered once in the same way.

\textbf{Running restrictions} The environment never repairs a model's action. Invalid calls return explicit error messages and are counted.
Budgets limitation include: 160 tool calls per round to restrict unlimited interactions. 6 check calls and 4 render calls per episode; the review returned by submit is free. An episode consists of up to three cumulative rounds (R1-R3). Each submit returns a free review (a check plus four views), and the design is checkpointed.

\paragraph{Comparability Across System Families} LMBuild includes both general-purpose LLM agents and domain-specific structure-generation systems to provide a broad view of current capabilities for structured 3D generation. Comparisons should nevertheless be interpreted with respect to the outputs natively supported by each system. General-purpose LLM agents receive the same LMBuild task specification, environment semantics, tool definitions, and interaction budgets, and are evaluated end-to-end across geometry, parts, joints, materials, and assembly sequence. In contrast, domain-specific systems are evaluated using their native inference pipelines and may produce only a subset of these outputs, as summarized in Table~\ref{tab:system_capability_comparison}. We therefore report a metric as N/A when the information required by that metric lies outside a system's native output space, rather than treating the absence of an unsupported output as a generation failure. Comparisons between LLM agents and domain-specific systems should consequently be interpreted primarily on their shared applicable metrics, while the complete LMBuild evaluation for LLM agents reflects their ability to solve the full end-to-end construction task. We include domain-specific systems not to imply identical interfaces or capabilities, but to contextualize how specialized generation methods compare with general-purpose agents on the portions of the structure-generation pipeline they support.

\textbf{Information exposed to agents.} The environment communicates generic, system-independent construction rules and diagnostic feedback needed to use its tools, while target-specific evaluation sheets and final metric outcomes are not exposed to agents. We view these generic rules as part of the benchmark interface rather than hidden test labels. The functional-prompt experiments intentionally expose additional task requirements and should be interpreted as measuring an agent’s ability to translate explicit specifications into structures, whereas the baseline also tests whether those requirements can be inferred from a category label and reference image. Future work may separately study settings with weaker feedback, hidden interface tolerances, or interactive requirement elicitation.

We provide the complete list of 19 tools implemented in the LMBuild environment that models can invoke when constructing structures, together with description of the functionality of each tool. 

\begin{Toolset}

\textbf{Tool: list-parts(query)}; \ \newline

\textbf{Function:} Lists the parts currently available to the agent, including catalog parts and any parts created by the model, together with their IDs, names, and dimensions. \ \newline

\textbf{Tool: inspect-part(part-id)}; \ \newline

\textbf{Function:} Inspects a specified part and returns its information, including dimensions, watertightness, surface area, and thumbnail path. \ \newline

\textbf{Tool: view-catalog-sheet()}; \ \newline

\textbf{Function:} Displays a labeled visual sheet containing all parts available in the catalog. This tool is unavailable when there is no sub-part pool. \ \newline

\textbf{Tool: list-materials()}; \ \newline

\textbf{Function:} Lists the available construction materials, including their material classes and physical densities. \ \newline

\textbf{Tool: place-part(part-id, instance-id, position, rotation, role-description)}; \ \newline 

\textbf{Function:} Places a selected part into the 3D scene at a specified position and optional rotation,  and specify its role if possible. It also returns feedback about its bounding box, contacts, collisions, and ground penetration. \ \newline

\textbf{Tool: set-material(part-id, instance-id)}; \ \newline 

\textbf{Function:} Specify the material of a selected part. \ \newline

\textbf{Tool: move-part(part-id, position, rotation)}; \ \newline

\textbf{Function:} Moves or rotates an existing part using either an absolute target position or a relative translation, while returning updated contact, collision, and ground-penetration information. \ \newline

\textbf{Tool: attach(instance-id, anchor, target, target-anchor, offset)}; \ \newline

\textbf{Function:} Aligns an anchor of one part with an anchor of another part or with the ground, enabling structured placement without manually specifying exact coordinates. \ \newline

\textbf{Tool: measure(part a, part b)}; \ \newline

\textbf{Function:} Measures the spatial relationship between two parts, including their minimum surface gap, closest points, touching status, and overlap volume. \ \newline

\textbf{Tool: remove-part(instance-id)}; \ \newline

\textbf{Function:} Removes a placed part from the current structure, together with its associated joints and its entry in the assembly sequence. \ \newline

\textbf{Tool: set-part-info(instance-id, role, material)}; \ \newline

\textbf{Function:} Updates the information associated with a placed part, such as its functional role and material assignment. \ \newline

\textbf{Tool: remove-joint(joint-id)}; \ \newline

\textbf{Function:} Removes an existing joint declaration between two parts. \ \newline

\textbf{Tool: add-joint(joint-id, type, parent, child, axis, origin)}; \ \newline

\textbf{Function:} Defines a joint relationship between two parts. Supported joint types include fixed, revolute, continuous, prismatic, cylindrical, and ball joints. \ \newline

\textbf{Tool: set-assembly-sequence(steps)}; \ \newline
 
\textbf{Function:} Specifies the intended assembly order of the generated structure, with an optional insertion direction for each assembly step. \ \newline

\textbf{Tool: create-part(name, program)}; \ \newline

\textbf{Function:} Creates a new custom part from a constructive solid geometry program. This tool is available when create new parts is enabled. \ \newline

\textbf{Tool: get-scene()}; \ \newline

\textbf{Function:} Returns the current state of the construction environment, including all placed parts, their poses, bounding boxes, roles, materials, joints, assembly sequence, and remaining interaction budgets. \ \newline

\textbf{Tool: check()}; \ \newline

\textbf{Function:} Use rule-based implementations to check the current structure for potential issues, including collisions, disconnected components, ground contact, stability problems, joint issues, and missing declarations. This tool is metered. \ \newline
 
\textbf{Tool: render(view)}; \ \newline
 
\textbf{Function:} Generates visual renderings of the current structure from one or more specified viewpoints, allowing the agent to inspect and refine its design. This tool is metered. \ \newline

\textbf{Tool: submit()}; \ \newline

\textbf{Function:} Ends the current interaction round. If additional rounds remain, it returns structural feedback and multiple rendered views before advancing to the next round; after the final round, the task is completed. \

\end{Toolset}

\subsection{Data Selection and Data Examples in LMBuild}
\label{app:data}

\paragraph{Data Selection Strategy.}

LMBuild is constructed from a heterogeneous pool of approximately \textsc{367K} raw 3D assets. Because these datasets differ substantially in geometry quality, semantic annotation, part decomposition, and kinematic information, we do not directly sample benchmark objects from the combined pool. Instead, we apply a five-stage selection and grounding pipeline designed to ensure that retained objects are structurally valid, sufficiently articulated, semantically identifiable, diverse in their underlying construction, and grounded in independently retrievable real-world knowledge. 

\paragraph{Candidate Admission and Structural Validity.}
We first remove assets that are unsuitable for evaluating part-based structure generation. An object is admitted into the candidate catalog only if it contains at least four separately identifiable components, its geometry can be successfully loaded and rendered, and any available joint annotations can be resolved to parts belonging to the same object. These requirements exclude corrupted meshes, effectively monolithic objects, incomplete assemblies, and assets whose kinematic annotations refer to missing or external components. After this initial filtering step, approximately \textsc{74.7K} objects remain as structurally valid candidates.

\paragraph{Source-aware and Diversity-aware Sampling.}
We next construct a manageable candidate subset while preserving structural diversity within each source and semantic category. Because the source datasets differ substantially in their characteristic level of decomposition, we use source-specific part-count ranges rather than imposing a single global threshold. We retain Artiverse objects with 4--60 parts, PartNeXt objects with 5--40 parts, LEGO assemblies containing 25--150 bricks, and Fusion~360 assemblies with 6--60 components and at least one movable joint. These ranges remove both trivial instances and extremely large assemblies whose complexity is dominated by dataset-specific decomposition rather than meaningful object structure. Within each source-category pair, we then perform diversity-aware sampling instead of random selection. Each candidate object is represented by a structural descriptor containing counts of semantic part roles together with counts of moving-part and joint-type combinations. We first select the instance closest to the component-wise median of these descriptors, yielding a representative medoid-like example. Subsequent instances are chosen using farthest-point sampling under $\ell_1$ distance, repeatedly selecting the object that is most dissimilar from the examples already retained. This procedure favors coverage of distinct structural configurations rather than repeatedly selecting near-duplicate instances of the same design. Dataset-specific metadata are additionally used when necessary: LEGO categories are recovered from captions and assemblies are prioritized according to movable connectors such as hinges, axles, and ball joints, whereas Fusion~360 categories are assigned from rendered views using a VLM.

\paragraph{Semantic Concept Grounding and Visual Verification.}
Structural validity alone does not guarantee that a CAD instance corresponds to a well-defined real-world object category. We therefore ground every retained category to both Wikidata and WordNet. For each category name, we retrieve the top Wikidata search candidates and the corresponding WordNet noun synsets. A language model is permitted only to select among these retrieved candidates, rather than freely generating an entity identifier. The selected Wikidata item and WordNet synset must denote the general physical object type represented by the category; entities corresponding to brands, named products, artworks, locations, or other non-category concepts are rejected. Categories for which no suitable Wikidata entity can be identified are removed. We additionally verify the category assignment at the individual-instance level. A VLM observes two rendered views of each candidate object and determines whether the asset both matches the intended category and represents a reasonably complete instance of that object. This step removes fragments, isolated subassemblies, severely incomplete models, and semantically mislabelled examples that may have passed the metadata-based filters.

\paragraph{Kinematic Coverage and Core-set Construction.}
From the grounded candidate pool, we construct the LMBuild-Core benchmark by explicitly covering qualitatively different forms of mechanical organization. Rather than selecting the 50 core tasks solely according to object category frequency, we organize them into six broad kinematic families: \textbf{(A)} static load-bearing structures, \textbf{(B)} hinged mechanisms, \textbf{(C)} sliding mechanisms, \textbf{(D)} rolling or rotating systems, \textbf{(E)} serial kinematic chains, and \textbf{(F)} structures containing many joints attached to a common body. These families are intended to expose models to substantially different requirements for connectivity, motion, joint specification, and physical realization. The final 50 core objects span a wide range of complexity, from small assemblies containing only four parts to highly decomposed structures containing as many as 417 parts. The remaining selected data form LMBuild-Full, containing 2,349 additional objects from 152 semantic categories. LMBuild-Full may share object categories with LMBuild-Core, but it contains no overlapping object instances, allowing the extended set to preserve broader intra-category structural variation without duplicating the tasks used in our primary evaluation.

\paragraph{Retrieval-based Real-world Knowledge Grounding.}
LMBuild further augments each selected object with external knowledge describing its expected components, materials, functional attributes, and mechanisms. Importantly, these annotations are not manually authored from the benchmark designers' prior knowledge and are not inferred directly from the CAD geometry. Instead, we construct a reproducible evidence bundle from Wikidata, Wikipedia, WordNet, and, only as a corroborating source, ConceptNet. For each category's Wikidata entity, we retrieve structured statements including has part, material, use, uses, powered by, and inverse part of relations. To recover information stated at a more general taxonomic level, we additionally traverse up to three levels of subclass of ancestors, retaining at most three ancestors per level and terminating at overly generic concepts such as physical object, product, or equipment. Part and material information may be inherited from these ancestors, but inherited parts are retained only when they have an English Wikipedia article and their own Wikidata entry independently confirms an appropriate part of relation. This conservative inheritance rule reduces the risk of propagating overly broad taxonomic properties to specific object categories. We also pin specific Wikipedia revisions for the target object, identified components, and mechanism-level concepts relevant to the object's kinematic family. For example, hinged structures may retrieve evidence from articles such as Hinge, rotating systems from Wheel or Caster, and articulated mechanisms from Kinematic pair. For identified moving parts, we additionally retrieve a relevant Wikipedia article when available. WordNet part and substance meronyms are incorporated as supplementary structured evidence and may be inherited over up to two hypernym levels. ConceptNet is used only to corroborate evidence already supported by stronger sources and is never treated as the sole basis for benchmark annotations.

\paragraph{Evidence-constrained Claim Extraction and Verification.}
The retrieved evidence is converted into benchmark annotations through a two-model extraction and verification procedure. Qwen3.5-27B first proposes candidate claims describing object parts, attributes, subsystems, and kinematic relationships. Every proposed claim must include a supporting quotation from the retrieved evidence bundle, and claims whose quoted evidence cannot be matched verbatim to the source text are automatically discarded. This prevents the extraction model from introducing unsupported world knowledge not present in the retrieved sources. A second model, Gemma-4-31B, independently evaluates each surviving claim along two dimensions: whether it is supported by the cited evidence and whether it is typical of the general object class rather than an unusual property of a particular instance. Core properties and more detailed attributes are assessed separately. When a claim fails because its supporting evidence is incomplete or poorly localized, we permit a single re-grounding step that retrieves at most three additional sentences from the same evidence bundle; no unrestricted external generation is allowed during this stage. The final annotation therefore retains only claims that can be traced to explicit external evidence and judged to describe characteristic properties of the object category. This pipeline produces the 50-task LMBuild-Core benchmark together with the 2,349-object, 152-category extended set. Each retained knowledge claim remains traceable to its source through either a Wikidata entity and statement identifier or a pinned Wikipedia page revision and supporting quotation. This provenance is important for LMBuild because the benchmark evaluates not only whether a generated structure visually resembles an object, but also whether its parts, materials, affordances, and kinematics are consistent with independently grounded knowledge about how that object is constructed and functions.

\begin{DataExample_1}

\begin{center}
\includegraphics[width=\linewidth]{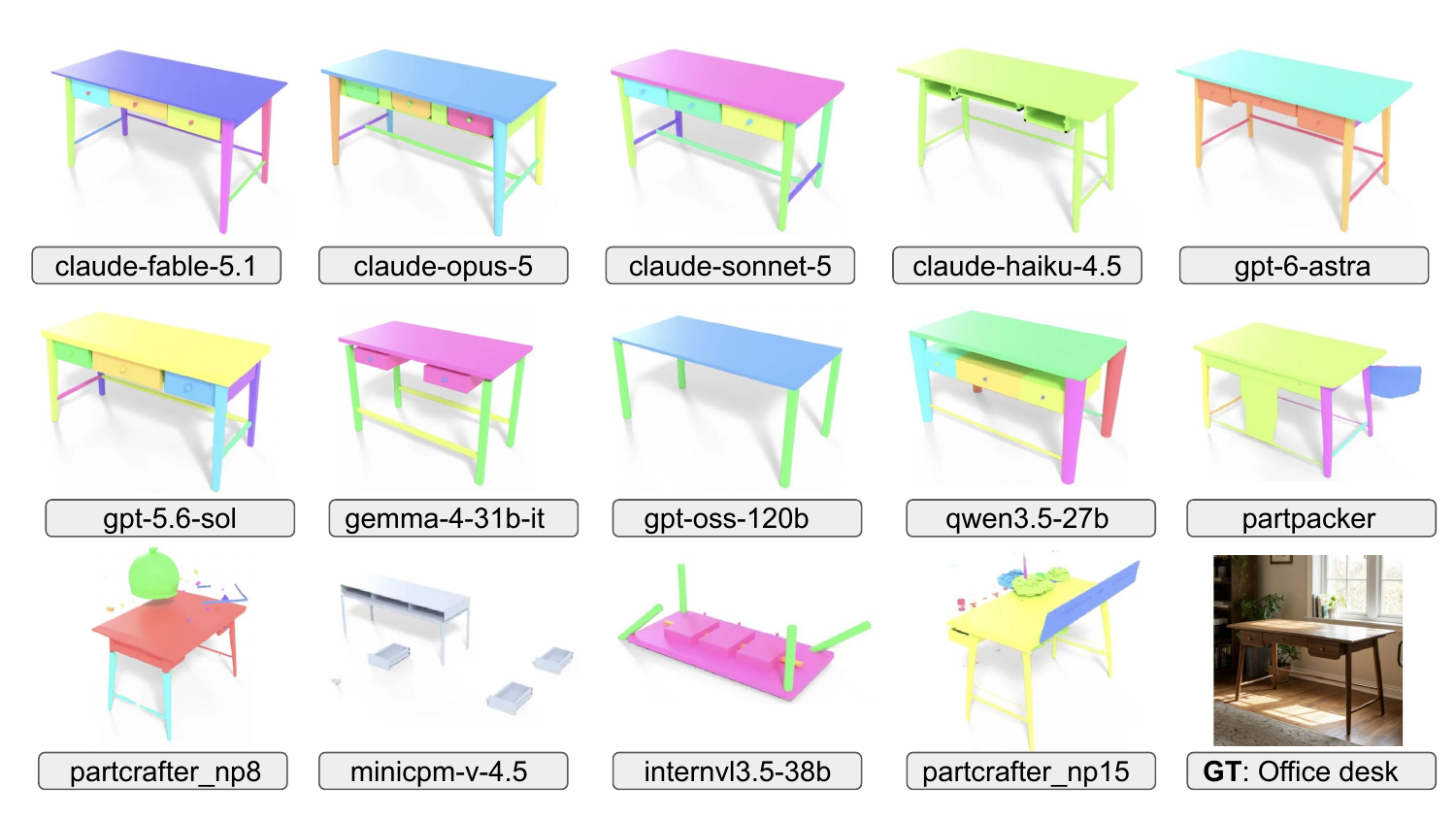}
\end{center}

\noindent\textbf{Specification the evaluator checks}

Verified claims from the pinned Wikipedia and Wikidata sources; 6 further extracted claims were dropped by the verifier. The small tag says how each claim is checked on a design; hover a source for its quote. Full sources and verdicts are in the panel below.

\medskip
\noindent\textbf{Parts 2}

\begin{itemize}[leftmargin=*,nosep]
    \item P1 \textbf{work surface} --- provides a flat area for reading, writing, or using equipment

    \item P2 \textbf{drawers} --- stores items such as office supplies and papers

    \item Wikidata ``has part'' (P527): \href{https://www.wikidata.org/wiki/Q30902154}{builders hardware} via Q14745, \href{https://www.wikidata.org/wiki/}{drawer}, \href{https://www.wikidata.org/wiki/Q1397443}{fiberboard} via Q14745, \href{https://www.wikidata.org/wiki/Q56056049}{furniture leg} via Q14748, \href{https://www.wikidata.org/wiki/Q898776}{particle board} via Q14745, \href{https://www.wikidata.org/wiki/Q1355652}{plank} via Q14745
\end{itemize}

\medskip
\noindent\textbf{Attributes 6}
\begin{itemize}[leftmargin=*,nosep]
    \item A1 \textbf{material} Desks are usually made of wood or metal, though glass is sometimes used.

    \item A2 \textbf{material} Modern mass-produced desks may be constructed from engineered wood products like fiberboard or particle board.

    \item A3 \textbf{property} A desk must be sturdy to support users who lean on it while working.

    \item A4 \textbf{use} Desks are used for academic, professional, or domestic activities such as reading, writing, or using computers.

    \item A5 \textbf{operating\_environment} Desks are found in schools, offices, homes, and libraries.

    \item A6 \textbf{user\_interaction} Users typically sit at a desk, either on a separate chair or a built-in chair.
\end{itemize}

\medskip
\noindent\textbf{Functional subsystems 1}

\begin{itemize}[leftmargin=*,nosep]
    \item F1 \textbf{Storage subsystem}: drawers, compartments, pigeonholes\\
    Stores office supplies, papers, and documents to keep the work surface clear.
\end{itemize}

\medskip
\noindent\textbf{Kinematics 1}

\begin{itemize}[leftmargin=*,nosep]
    \item K1 \textbf{drawer} slides relative to desk frame \textit{prismatic} range: from fully enclosed to fully extended
\end{itemize}

\medskip
\noindent\textbf{Attributes prompt}

\begin{quote}
\small
Build a desk. Include a work surface to provide a flat area for reading, writing, or using equipment and drawers to store items such as office supplies and papers. Use wood, metal, or glass, or engineered wood products like fiberboard or particle board. Ensure the desk is sturdy to support users who lean on it while working. Design it for academic, professional, or domestic activities such as reading, writing, or using computers. Ensure it is suitable for schools, offices, homes, and libraries. Design it so users typically sit at the desk, either on a separate chair or a built-in chair.
\end{quote}

\noindent\textbf{Functional prompt}

\begin{quote}
\small
Build a desk. Include a work surface to provide a flat area for reading, writing, or using equipment and drawers to store items such as office supplies and papers. Use wood, metal, or glass, or engineered wood products like fiberboard or particle board. Ensure the desk is sturdy to support users who lean on it while working. Design it for academic, professional, or domestic activities such as reading, writing, or using computers. Ensure it is suitable for schools, offices, homes, and libraries. Design it so users typically sit at the desk, either on a separate chair or a built-in chair. Incorporate a storage subsystem consisting of drawers, compartments, and pigeonholes to store office supplies, papers, and documents to keep the work surface clear. Install drawer slides relative to the desk frame that range from fully enclosed to fully extended.
\end{quote}

\end{DataExample_1}

\begin{DataExample_2}

\begin{center}
\includegraphics[width=\linewidth]{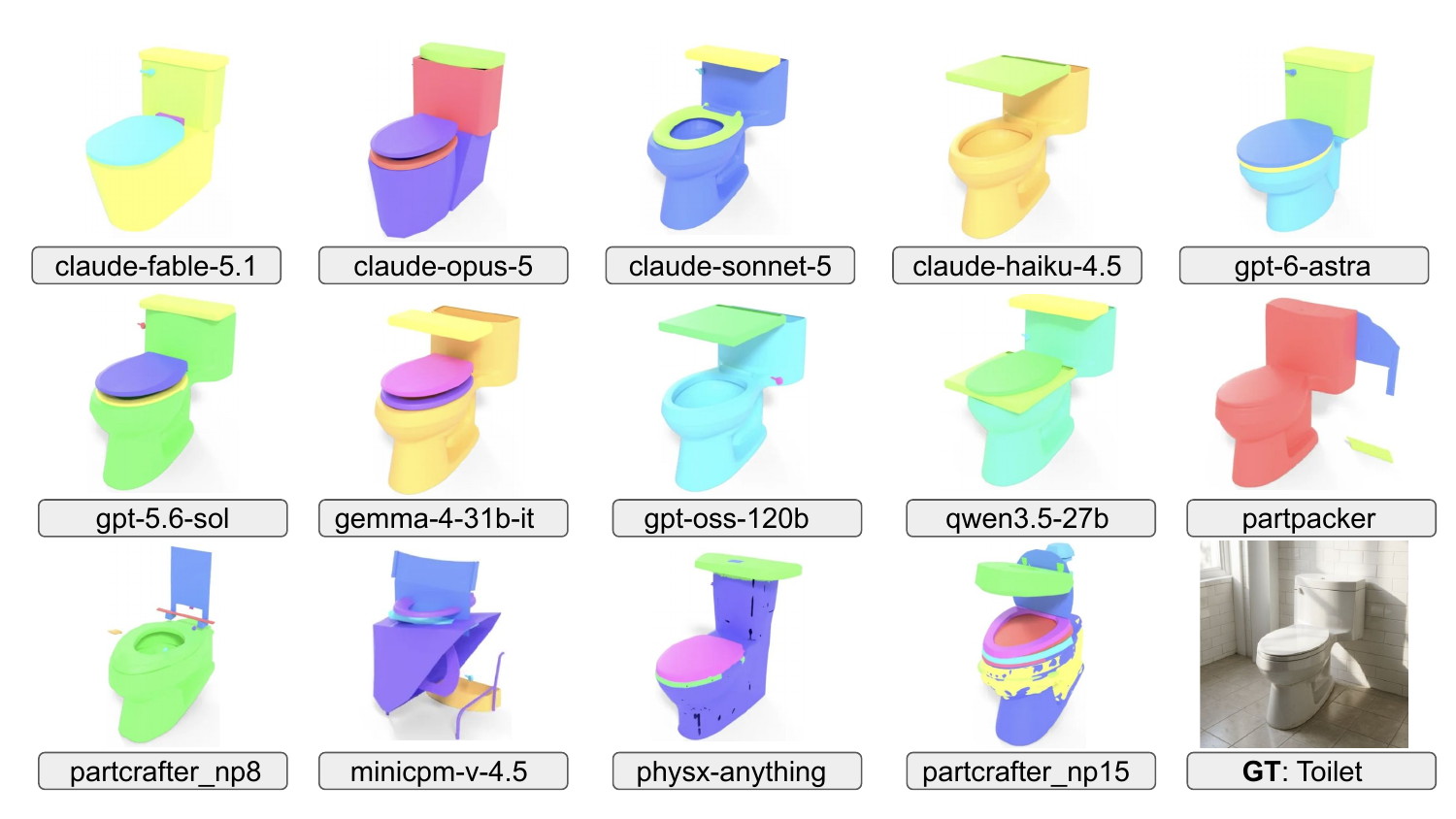}
\end{center}

\noindent\textbf{Specification the evaluator checks}

Verified claims from the pinned Wikipedia and Wikidata sources; 5 further extracted claims were dropped by the verifier. The small tag says how each claim is checked on a design; hover a source for its quote. Full sources and verdicts are in the panel below.

\medskip
\noindent\textbf{Parts 5}

\begin{itemize}[leftmargin=*,nosep]
    \item P1 \textbf{bowl} --- collects human waste and holds water to seal against sewer gases

    \item P2 \textbf{trap} --- holds water in the bowl to act as a seal against noxious sewer gases

    \item P3 \textbf{drain pipe} --- removes effluent from the bowl to a sewer or septic tank

    \item P4 \textbf{flush mechanism} --- releases a large volume of water rapidly to sweep waste into the drain

    \item P5 \textbf{cistern} --- stores water for flushing in tank-style toilets

    \item Wikidata ``has part'' (P527): \href{https://www.wikidata.org/wiki/Q18352682}{dry toilet}, \href{https://www.wikidata.org/wiki/Q268534}{flush toilet}, \href{https://www.wikidata.org/wiki/Q7813355}{toilet}, \href{https://www.wikidata.org/wiki/}{toilet bowl}, \href{https://www.wikidata.org/wiki/Q126127572}{toilet building}, \href{https://www.wikidata.org/wiki/}{toilet seat}, \href{https://www.wikidata.org/wiki/Q646411}{toilets in japan}
\end{itemize}

\medskip
\noindent\textbf{Attributes 7}

\begin{itemize}[leftmargin=*,nosep]
    \item A1 \textbf{material} Toilets are commonly made of ceramic (porcelain), concrete, plastic, or wood.

    \item A2 \textbf{use} The primary function is to collect human waste (urine and feces) and sometimes toilet paper for disposal.

    \item A3 \textbf{operating\_environment} Flush toilets are usually connected to a sewer system in urban areas or a septic tank in isolated areas.

    \item A4 \textbf{power\_source} Flush toilets use water as the primary medium to move waste, while dry toilets use no water.

    \item A5 \textbf{user\_interaction} Users operate the flush by pressing a button, pushing a handle, pulling a lever, or pulling a chain.

    \item A6 \textbf{property} The water in the bowl acts as a barrier to prevent sewer gas from entering the building.

    \item A7 \textbf{use} Toilets can be designed for sitting or squatting postures depending on regional preferences.
\end{itemize}

\medskip
\noindent\textbf{Functional subsystems 3}

\begin{itemize}[leftmargin=*,nosep]
    \item F1 \textbf{Waste Collection and Disposal}: bowl, trap, drain pipe\\
    Collects waste and uses a water seal to prevent gas entry while channeling waste to the sewer.

    \item F2 \textbf{Water Storage and Delivery}: cistern, fill valve, flush valve\\
    Stores water and releases it rapidly into the bowl to flush waste.

    \item F3 \textbf{User Interface}: flush handle, flush button, seat\\
    Allows the user to initiate the flush and provides a seating surface.
\end{itemize}

\medskip
\noindent\textbf{Kinematics 2}

\begin{itemize}[leftmargin=*,nosep]
    \item K1 \textbf{flush handle} swings relative to cistern \textit{revolute}

    \item K2 \textbf{flush valve (flapper)} swings relative to cistern outlet \textit{revolute} range: lifts clear of the seat to allow flow
\end{itemize}

\medskip
\noindent\textbf{Attributes prompt}

\begin{quote}
\small
Build a toilet. Use a bowl to collect human waste and hold water to seal against sewer gases. Include a trap to hold water in the bowl as a seal against noxious sewer gases. Attach a drain pipe to remove effluent from the bowl to a sewer or septic tank. Incorporate a flush mechanism to release a large volume of water rapidly to sweep waste into the drain. Add a cistern to store water for flushing in tank-style toilets. Construct it from ceramic, concrete, plastic, or wood. Design it to collect human waste and sometimes toilet paper for disposal. Ensure it can connect to a sewer system in urban areas or a septic tank in isolated areas. Use water as the primary medium to move waste. Allow users to operate the flush by pressing a button, pushing a handle, pulling a lever, or pulling a chain. Ensure the water in the bowl acts as a barrier to prevent sewer gas from entering the building. Design it for sitting or squatting postures.
\end{quote}

\noindent\textbf{Functional prompt}

\begin{quote}
\small
Build a toilet. Use a bowl to collect human waste and hold water to seal against sewer gases. Include a trap to hold water in the bowl as a seal against noxious sewer gases. Attach a drain pipe to remove effluent from the bowl to a sewer or septic tank. Incorporate a flush mechanism to release a large volume of water rapidly to sweep waste into the drain. Add a cistern to store water for flushing in tank-style toilets. Construct it from ceramic, concrete, plastic, or wood. Design it to collect human waste and sometimes toilet paper for disposal. Ensure it can connect to a sewer system in urban areas or a septic tank in isolated areas. Use water as the primary medium to move waste. Allow users to operate the flush by pressing a button, pushing a handle, pulling a lever, or pulling a chain. Ensure the water in the bowl acts as a barrier to prevent sewer gas from entering the building. Design it for sitting or squatting postures. Create a Waste Collection and Disposal subsystem where the bowl connects to the trap and the trap connects to the drain pipe to channel waste and prevent gas entry. Create a Water Storage and Delivery subsystem where a fill valve carries fluid to the cistern, the cistern carries fluid to the flush valve, and the flush valve carries fluid to the bowl. Create a User Interface subsystem where a flush handle or flush button carries motion to the flush valve. Ensure the flush handle swings relative to the cistern. Ensure the flush valve swings relative to the cistern outlet and lifts clear of the seat to allow flow.
\end{quote}

\end{DataExample_2}

\begin{DataExample_3}

\begin{center}
\includegraphics[width=\linewidth]{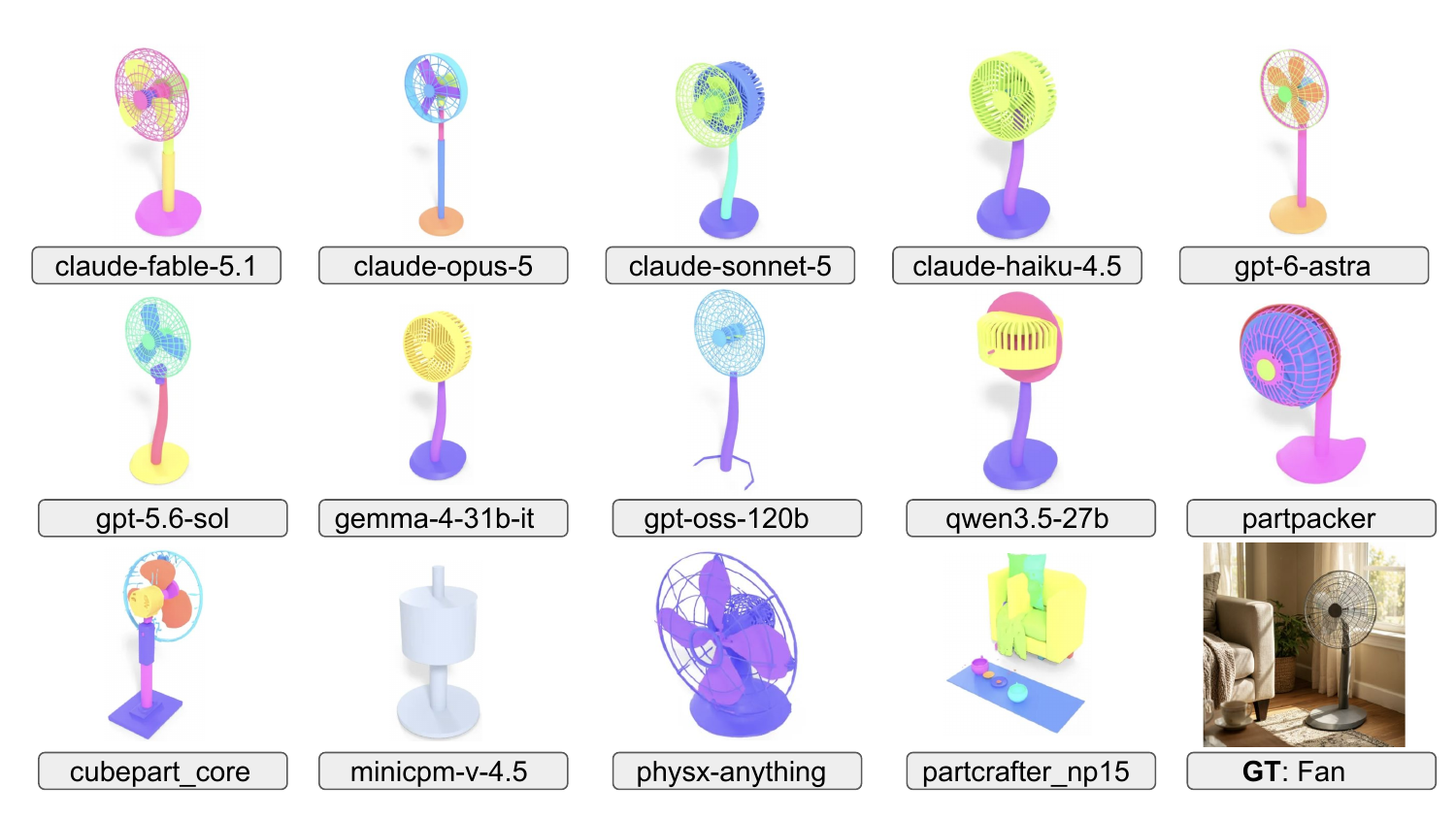}
\end{center}

\noindent\textbf{Specification the evaluator checks}

Verified claims from the pinned Wikipedia and Wikidata sources; 3 further extracted claims were dropped by the verifier.

\medskip
\noindent\textbf{Parts 7}

\begin{itemize}[leftmargin=*,nosep]
    \item P1 \textbf{blades} --- to act on the air and create airflow

    \item P2 \textbf{electric motor} --- to convert electrical energy into mechanical rotation

    \item P3 \textbf{hub} --- to connect the blades to the rotating shaft

    \item P4 \textbf{housing} --- to direct airflow and prevent contact with blades

    \item P5 \textbf{blade guard} --- to increase safety by preventing objects from contacting the fan blades

    \item P6 \textbf{oscillator gearbox} --- to motion the fan from side to side

    \item P7 \textbf{base} --- to support the fan and provide stability

    \item Wikidata ``has part'' (P527): \href{https://www.wikidata.org/wiki/Q193466}{rotor}, \href{https://www.wikidata.org/wiki/Q1698700}{impeller}, \href{https://www.wikidata.org/wiki/Q72313}{electric motor}, \href{https://www.wikidata.org/wiki/Q931447}{electrical load} via Q2425052
\end{itemize}

\medskip
\noindent\textbf{Attributes 4}

\begin{itemize}[leftmargin=*,nosep]
    \item A1 \textbf{material} Blades are generally made of wood, plastic, or metal.

    \item A2 \textbf{power\_source} Most fans are powered by electric motors using AC or DC electricity.

    \item A3 \textbf{operating\_environment} Fans are effective for cooling people but do not lower air temperature directly.

    \item A4 \textbf{property} Fans produce air flows with high volume and low pressure.
\end{itemize}

\medskip
\noindent\textbf{Functional subsystems 3}

\begin{itemize}[leftmargin=*,nosep]
    \item F1 \textbf{Airflow generation}: blades, hub, electric motor\\
    Rotates blades to move air parallel to the shaft in axial fans or radially in centrifugal fans.

    \item F2 \textbf{Oscillation mechanism}: oscillator gearbox, oscillator shaft, base\\
    Moves the fan head from side to side to distribute airflow.

    \item F3 \textbf{Safety enclosure}: housing, blade guard\\
    Prevents objects from contacting the fan blades while allowing air to pass through.
\end{itemize}

\medskip
\noindent\textbf{Kinematics 1}

\begin{itemize}[leftmargin=*,nosep]
    \item K1 \textbf{fan head} swings relative to base \textit{revolute} range: side to side
\end{itemize}

\medskip
\noindent\textbf{Attributes prompt}

\begin{quote}
\small
Build an electric fan. Use blades to act on the air and create airflow, an electric motor to convert electrical energy into mechanical rotation, and a hub to connect the blades to the rotating shaft. Include a housing to direct airflow and prevent contact with blades, a blade guard to increase safety by preventing objects from contacting the fan blades, an oscillator gearbox to motion the fan from side to side, and a base to support the fan and provide stability. The blades are generally made of wood, plastic, or metal. Most fans are powered by electric motors using AC or DC electricity. Fans are effective for cooling people but do not lower air temperature directly. Fans produce air flows with high volume and low pressure.
\end{quote}

\noindent\textbf{Functional prompt}

\begin{quote}
\small
Build an electric fan. Use blades to act on the air and create airflow, an electric motor to convert electrical energy into mechanical rotation, and a hub to connect the blades to the rotating shaft. Include a housing to direct airflow and prevent contact with blades, a blade guard to increase safety by preventing objects from contacting the fan blades, an oscillator gearbox to motion the fan from side to side, and a base to support the fan and provide stability. The blades are generally made of wood, plastic, or metal. Most fans are powered by electric motors using AC or DC electricity. Fans are effective for cooling people but do not lower air temperature directly. Fans produce air flows with high volume and low pressure. For airflow generation, rotate blades to move air using the electric motor and hub, where the hub to blades carries motion. For the oscillation mechanism, use the oscillator gearbox and base to move the fan head from side to side to distribute airflow, where the oscillator gearbox to oscillator shaft carries motion and the oscillator shaft to base carries motion. For the safety enclosure, use the housing and blade guard to prevent objects from contacting the fan blades while allowing air to pass through, where the blade guard to motor case carries load. Ensure the fan head swings relative to base with a side to side range.
\end{quote}

\end{DataExample_3}

\begin{DataExample_4}

\begin{center}
\includegraphics[width=\linewidth]{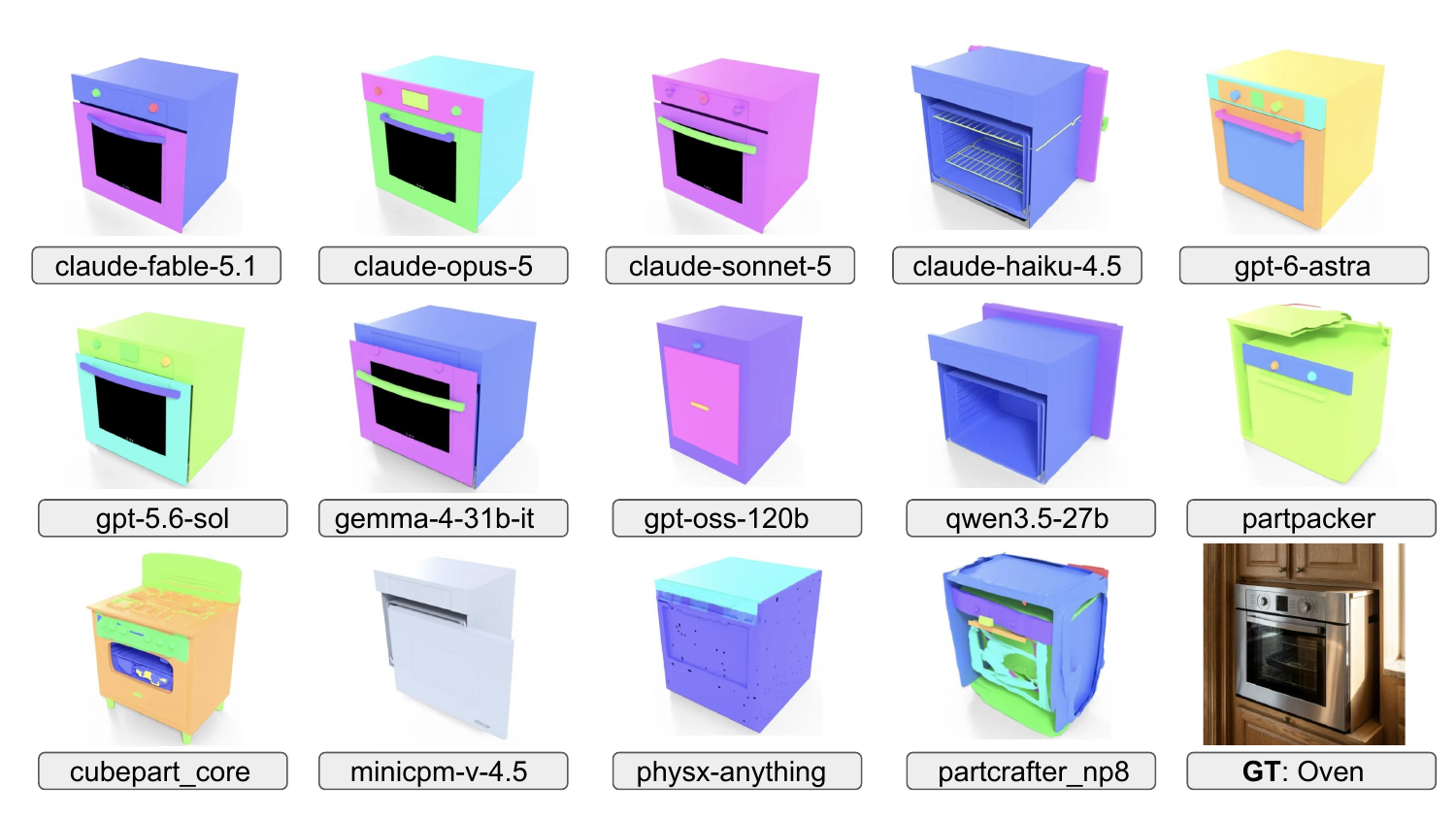}
\end{center}

\noindent\textbf{Specification the evaluator checks}

Verified claims from the pinned Wikipedia and Wikidata sources; 3 further extracted claims were dropped by the verifier. The small tag says how each claim is checked on a design; hover a source for its quote. Full sources and verdicts are in the panel below.

\medskip
\noindent\textbf{Parts 5}

\begin{itemize}[leftmargin=*,nosep]
    \item P1 \textbf{heating chamber} --- encloses the space where materials are exposed to heat

    \item P2 \textbf{heating element} --- generates heat to raise the temperature inside the chamber

    \item P3 \textbf{rack} --- supports food items inside the chamber for even heating

    \item P4 \textbf{thermostat} --- regulates the temperature by turning the heating source on or off

    \item P5 \textbf{control interface} --- allows the user to set temperature and operating modes

    \item Wikidata ``has part'' (P527): \href{https://www.wikidata.org/wiki/Q1249476}{oven}, \href{https://www.wikidata.org/wiki/Q96279691}{grate}, \href{https://www.wikidata.org/wiki/Q228255}{zug}
\end{itemize}

\medskip
\noindent\textbf{Attributes 6}

\begin{itemize}[leftmargin=*,nosep]
    \item A1 \textbf{use} Ovens are used to expose materials to a hot environment for cooking, baking, or industrial processing.

    \item A2 \textbf{power\_source} Ovens are typically powered by electricity, natural gas, or combustion of solid fuels like wood or coal.

    \item A3 \textbf{property} The interior of an oven reaches high temperatures to cook or process materials.

    \item A4 \textbf{operating\_environment} Ovens are commonly found in kitchens, restaurants, and industrial facilities.

    \item A5 \textbf{user\_interaction} Users interact with ovens via controls to set temperature, time, and cooking modes.

    \item A6 \textbf{material} Oven chambers are constructed from fireproof materials such as brick, stone, clay, or metal.
\end{itemize}

\medskip
\noindent\textbf{Functional subsystems 4}

\begin{itemize}[leftmargin=*,nosep]
    \item F1 \textbf{Heating System}: heating element, fuel source, combustion chamber\\
    Generates and distributes heat within the chamber to cook or process materials.

    \item F2 \textbf{Temperature Control System}: thermostat, control interface, sensors\\
    Maintains the chamber temperature at a user-defined setpoint by regulating the heating source.

    \item F3 \textbf{Access System}: door, hinge, handle\\
    Provides a sealable opening for loading and unloading materials while retaining heat.

    \item F4 \textbf{Support System}: rack, shelf\\
    Holds food or materials at specific heights within the chamber for optimal heat exposure.
\end{itemize}

\medskip
\noindent\textbf{Kinematics 2}

\begin{itemize}[leftmargin=*,nosep]
    \item K1 \textbf{door} swings relative to oven frame \textit{revolute}

    \item K2 \textbf{rotisserie spit} rotates relative to oven interior \textit{revolute}
\end{itemize}

\medskip
\noindent\textbf{Attributes prompt}

\begin{quote}
\small
Build an oven. Use it to expose materials to a hot environment for cooking, baking, or industrial processing. Power it with electricity, natural gas, or combustion of solid fuels like wood or coal. Ensure the interior reaches high temperatures to cook or process materials. Place it in kitchens, restaurants, or industrial facilities. Allow users to interact via controls to set temperature, time, and cooking modes. Construct the heating chamber from fireproof materials such as brick, stone, clay, or metal. Include a heating chamber to enclose the space where materials are exposed to heat. Include a heating element to generate heat to raise the temperature inside the chamber. Include a rack to support food items inside the chamber for even heating. Include a thermostat to regulate the temperature by turning the heating source on or off. Include a control interface to allow the user to set temperature and operating modes.
\end{quote}

\noindent\textbf{Functional prompt}

\begin{quote}
\small
Build an oven. Use it to expose materials to a hot environment for cooking, baking, or industrial processing. Power it with electricity, natural gas, or combustion of solid fuels like wood or coal. Ensure the interior reaches high temperatures to cook or process materials. Place it in kitchens, restaurants, or industrial facilities. Allow users to interact via controls to set temperature, time, and cooking modes. Construct the heating chamber from fireproof materials such as brick, stone, clay, or metal. Include a heating chamber to enclose the space where materials are exposed to heat. Include a heating element to generate heat to raise the temperature inside the chamber. Include a rack to support food items inside the chamber for even heating. Include a thermostat to regulate the temperature by turning the heating source on or off. Include a control interface to allow the user to set temperature and operating modes. Implement a Heating System using a heating element, fuel source, and combustion chamber to generate and distribute heat within the chamber. Implement a Temperature Control System using a thermostat, control interface, and sensors to maintain the chamber temperature at a user-defined setpoint. Implement an Access System using a door, hinge, and handle to provide a sealable opening, where the door to the heating chamber carries load. Implement a Support System using a rack and shelf to hold materials, where the rack to the heating chamber carries load. Ensure the door swings relative to the oven frame. Ensure the rotisserie spit rotates relative to the oven interior.
\end{quote}

\end{DataExample_4}

\begin{DataExample_5}

\begin{center}
\includegraphics[width=\linewidth]{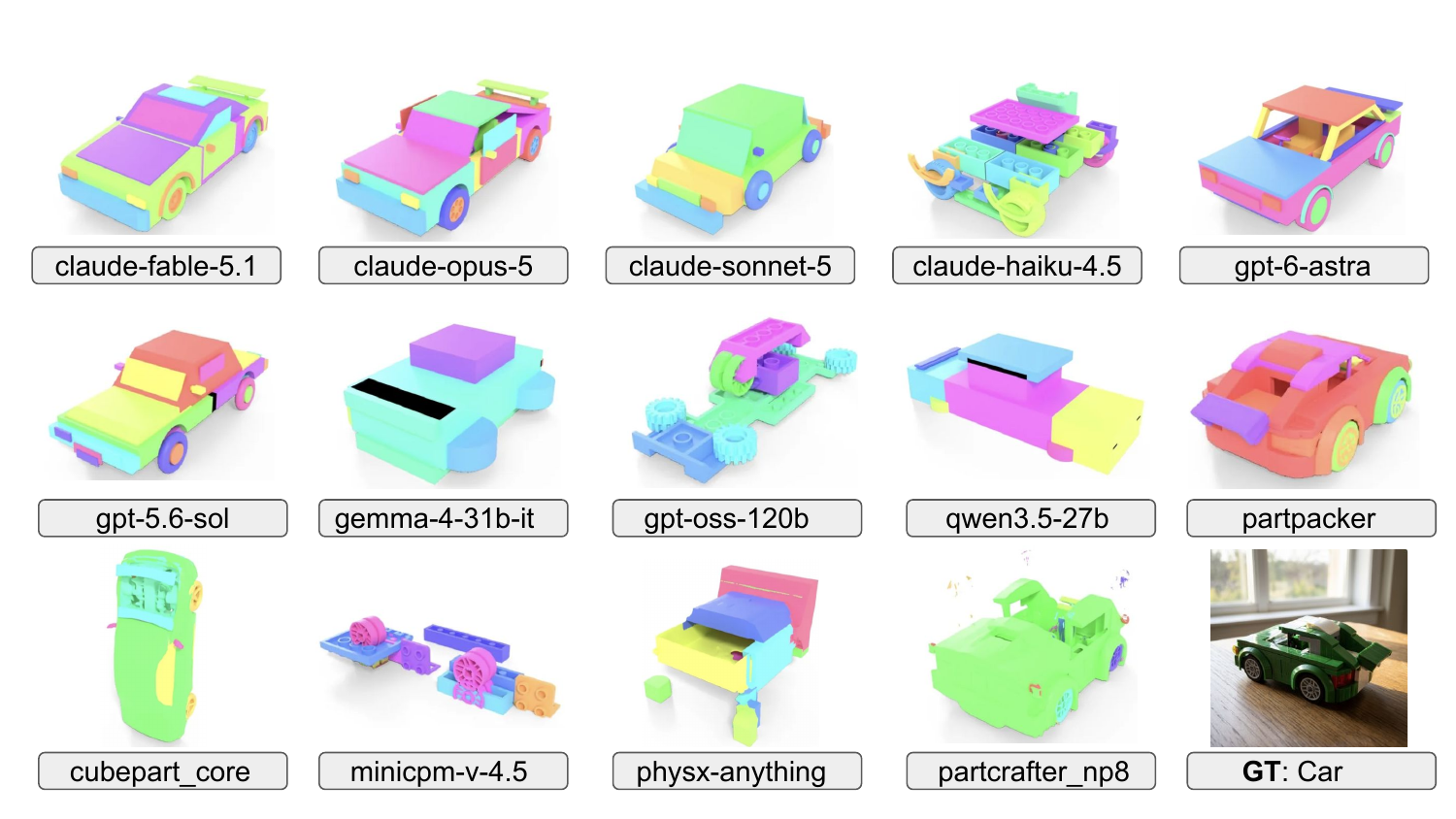}
\end{center}

\noindent\textbf{Specification the evaluator checks}

Verified claims from the pinned Wikipedia and Wikidata sources; 11 further extracted claims were dropped by the verifier. The small tag says how each claim is checked on a design; hover a source for its quote. Full sources and verdicts are in the panel below.

\medskip
\noindent\textbf{Parts 9}

\begin{itemize}[leftmargin=*,nosep]
    \item P1 \textbf{Engine} --- converts fuel energy into mechanical motion to propel the vehicle

    \item P2 \textbf{Wheels and Tires} --- support the vehicle load and provide traction on the road surface

    \item P3 \textbf{Steering System} --- allows the driver to control the direction of the vehicle

    \item P4 \textbf{Brakes} --- inhibit motion to slow or stop the vehicle

    \item P5 \textbf{Transmission} --- transfers engine power to the wheels and adjusts torque and speed

    \item P6 \textbf{Fuel Tank} --- stores the flammable fluid required for the engine to operate

    \item P7 \textbf{Seats} --- provide a place for occupants to sit while the vehicle is in motion

    \item P8 \textbf{Doors} --- allow ingress and egress for passengers and secure the vehicle interior

    \item P9 \textbf{Suspension} --- absorbs shock impulses from the road to maintain tire contact and ride comfort
\end{itemize}

\medskip
\noindent\textbf{Attributes 6}

\begin{itemize}[leftmargin=*,nosep]
    \item A1 \textbf{power\_source} Most cars are powered by internal combustion engines burning fossil fuels, though electric motors are increasingly common.

    \item A2 \textbf{material} Car bodies are typically constructed from steel, aluminum, or increasingly from composite materials like carbon fiber.

    \item A3 \textbf{use} Cars are primarily used to transport people rather than cargo on roads.

    \item A4 \textbf{operating\_environment} Cars are designed to operate primarily on paved roads.

    \item A5 \textbf{property} Modern cars typically weigh between 1 and 3 tonnes.

    \item A6 \textbf{user\_interaction} Drivers control the vehicle using a steering wheel, pedals for braking and acceleration, and a gear shift.
\end{itemize}

\medskip
\noindent\textbf{Functional subsystems 2}

\begin{itemize}[leftmargin=*,nosep]
    \item F1 \textbf{Propulsion System}: Engine, Fuel Tank, Transmission, Clutch, Drive Shaft, Differential, Axle\\
    Generates power from fuel and transmits it to the wheels to move the vehicle.

    \item F2 \textbf{Steering System}: Steering Wheel, Steering Column, Rack and Pinion, Tie Rods, Steering Knuckle, Ball Joint\\
    Translates driver input into the turning of the front wheels to change direction.
\end{itemize}

\medskip
\noindent\textbf{Kinematics 5}

\begin{itemize}[leftmargin=*,nosep]
    \item K1 \textbf{Steering Wheel} rotates relative to Steering Column \textit{revolute}

    \item K2 \textbf{Front Wheels} swings relative to Vehicle Chassis \textit{revolute}

    \item K3 \textbf{Car Door} swings relative to Vehicle Body \textit{revolute}

    \item K4 \textbf{Side Windows} slides relative to Door Frame \textit{prismatic} axis: vertical axis

    \item K5 \textbf{Brake Pedal} swings relative to Dashboard/Floor \textit{revolute}
\end{itemize}

\medskip
\noindent\textbf{Attributes prompt}

\begin{quote}
\small
Build a car. Use an engine, wheels and tires, a steering system, brakes, a transmission, a fuel tank, seats, doors, and suspension. The vehicle is powered by internal combustion engines or electric motors and constructed from steel, aluminum, or composite materials. It is primarily used to transport people on roads, specifically paved roads. The car typically weighs between 1 and 3 tonnes. Drivers interact with it using a steering wheel, braking and acceleration pedals, and a gear shift.
\end{quote}

\noindent\textbf{Functional prompt}

\begin{quote}
\small
Build a car. Use an engine, wheels and tires, a steering system, brakes, a transmission, a fuel tank, seats, doors, and suspension. The vehicle is powered by internal combustion engines or electric motors and constructed from steel, aluminum, or composite materials. It is primarily used to transport people on roads, specifically paved roads. The car typically weighs between 1 and 3 tonnes. Drivers interact with it using a steering wheel, braking and acceleration pedals, and a gear shift. Integrate a propulsion system where the engine to transmission carries power and the differential to axle carries motion. Include a steering system where the steering wheel to steering column carries motion, the steering column to rack and pinion carries motion, the rack and pinion to tie rods carries motion, the tie rods to steering knuckle carries force, and the steering knuckle to ball joint carries force. Ensure the steering wheel rotates relative to the steering column, front wheels swing relative to the vehicle chassis, the car door swings relative to the vehicle body, side windows slide vertically relative to the door frame, and the brake pedal swings relative to the dashboard or floor.
\end{quote}

\end{DataExample_5}

\begin{DataExample_6}

\begin{center}
\includegraphics[width=\linewidth]{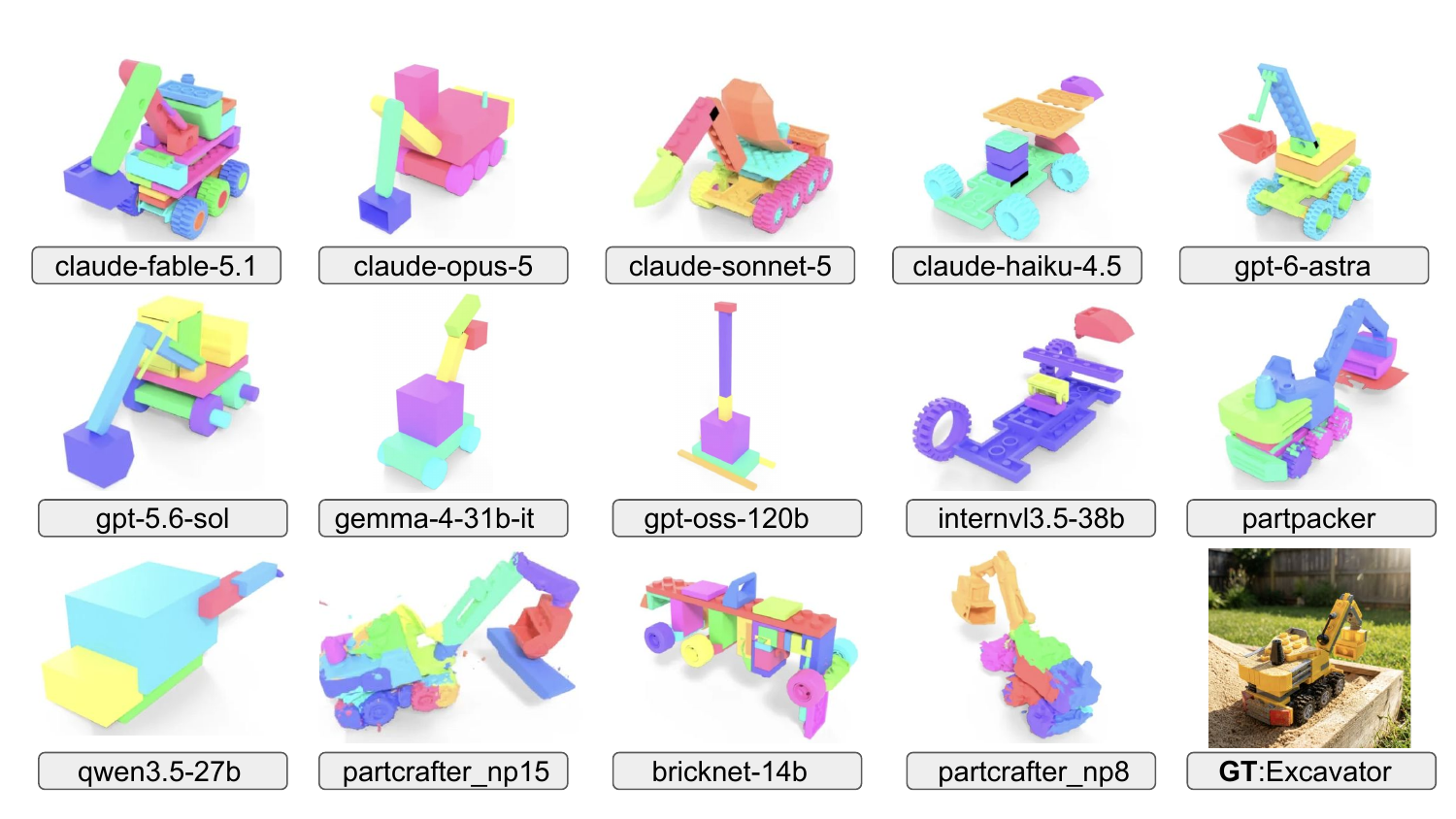}
\end{center}

\noindent\textbf{Specification the evaluator checks}

Verified claims from the pinned Wikipedia and Wikidata sources; 5 further extracted claims were dropped by the verifier. The small tag says how each claim is checked on a design; hover a source for its quote. Full sources and verdicts are in the panel below.

\medskip
\noindent\textbf{Parts 9}

\begin{itemize}[leftmargin=*,nosep]
    \item P1 \textbf{Undercarriage} --- Provides the mobile base with tracks or wheels for locomotion

    \item P2 \textbf{House} --- Contains the engine, operator cab, and hydraulic systems on a rotating platform

    \item P3 \textbf{Boom} --- Provides the primary up-and-down movement for the digging arm

    \item P4 \textbf{Stick (Dipper)} --- Provides the digging movement to pull the bucket through the ground

    \item P5 \textbf{Bucket} --- Serves as the container for scooping and holding excavated material

    \item P6 \textbf{Engine} --- Converts fuel energy into mechanical power to drive hydraulic pumps

    \item P7 \textbf{Hydraulic Pumps} --- Convert mechanical engine power into high-pressure fluid flow for actuators

    \item P8 \textbf{Hydraulic Cylinders} --- Convert fluid pressure into linear force to move the boom, stick, and bucket

    \item P9 \textbf{Swing Motor} --- Rotates the house 360 degrees relative to the undercarriage
\end{itemize}

\medskip
\noindent\textbf{Attributes 5}

\begin{itemize}[leftmargin=*,nosep]
    \item A1 \textbf{power\_source} Excavators typically use an internal combustion engine to drive hydraulic pumps.

    \item A2 \textbf{use} Excavators are used for digging trenches, holes, foundations, and material handling.

    \item A3 \textbf{operating\_environment} Excavators operate in construction sites, mining operations, and confined spaces.

    \item A4 \textbf{property} Hydraulic systems in excavators operate at high pressures up to 5000 psi for main functions.

    \item A5 \textbf{user\_interaction} Operators control the machine using joysticks that distribute digging functions across two x-y controls.
\end{itemize}

\medskip
\noindent\textbf{Functional subsystems 3}

\begin{itemize}[leftmargin=*,nosep]
    \item F1 \textbf{Hydraulic Power System}: Engine, Hydraulic Pumps, Hydraulic Oil Tanks, Hydraulic Cylinders, Hydraulic Motors\\
    Converts engine mechanical energy into hydraulic fluid pressure to actuate all moving parts.

    \item F2 \textbf{Arm Assembly}: Boom, Stick, Bucket, Hydraulic Cylinders\\
    Performs the digging, lifting, and dumping motions required for excavation.

    \item F3 \textbf{Travel System}: Undercarriage, Tracks, Track Motors, Final Drives\\
    Propels the machine forward, backward, and turns via independent track drive.
\end{itemize}

\medskip
\noindent\textbf{Kinematics 4}

\begin{itemize}[leftmargin=*,nosep]
    \item K1 \textbf{House} rotates relative to Undercarriage \textit{revolute} axis: vertical center pin range: 360 degrees

    \item K2 \textbf{Stick} swings relative to Boom \textit{revolute} range: closer and further

    \item K3 \textbf{Bucket} swings relative to Stick \textit{revolute}

    \item K4 \textbf{Piston Rod} slides relative to Cylinder Barrel \textit{prismatic} range: extension and retraction
\end{itemize}

\medskip
\noindent\textbf{Attributes prompt}

\begin{quote}
\small
Build an excavator. Use an undercarriage for the mobile base. Build a house containing the engine, operator cab, and hydraulic systems on a rotating platform. Attach a boom for primary up-and-down movement, a stick for digging movement, and a bucket for scooping material. Include an engine to convert fuel energy into mechanical power, hydraulic pumps to convert that power into high-pressure fluid flow, and hydraulic cylinders to convert fluid pressure into linear force. Add a swing motor to rotate the house 360 degrees relative to the undercarriage. The excavator uses an internal combustion engine to drive hydraulic pumps and is used for digging trenches, holes, foundations, and material handling. It operates in construction sites, mining operations, and confined spaces. The hydraulic systems operate at high pressures up to 5000 psi. Operators control the machine using joysticks.
\end{quote}

\noindent\textbf{Functional prompt}

\begin{quote}
\small
Build an excavator. Use an undercarriage for the mobile base. Build a house containing the engine, operator cab, and hydraulic systems on a rotating platform. Attach a boom for primary up-and-down movement, a stick for digging movement, and a bucket for scooping material. Include an engine to convert fuel energy into mechanical power, hydraulic pumps to convert that power into high-pressure fluid flow, and hydraulic cylinders to convert fluid pressure into linear force. Add a swing motor to rotate the house 360 degrees relative to the undercarriage. The excavator uses an internal combustion engine to drive hydraulic pumps and is used for digging trenches, holes, foundations, and material handling. It operates in construction sites, mining operations, and confined spaces. The hydraulic systems operate at high pressures up to 5000 psi. Operators control the machine using joysticks. Create a Hydraulic Power System using the engine, hydraulic pumps, hydraulic oil tanks, hydraulic cylinders, and hydraulic motors to convert mechanical energy into fluid pressure, where the engine connects to the hydraulic pumps. Assemble an Arm Assembly using the boom, stick, bucket, and hydraulic cylinders, where the boom carries load to the stick and the stick carries load to the bucket. Build a Travel System using the undercarriage, tracks, track motors, and final drives, where track motors and final drives carry motion to the tracks. Ensure the house rotates 360 degrees relative to the undercarriage on a vertical center pin. The stick must swing relative to the boom, and the bucket must swing relative to the stick. The piston rod must slide relative to the cylinder barrel.
\end{quote}

\end{DataExample_6}

\paragraph{Examples of Declared Joint Motion}
\label{app:declared-joint-motion}

\begin{figure}[p]
    \centering
    \includegraphics[width=\linewidth]{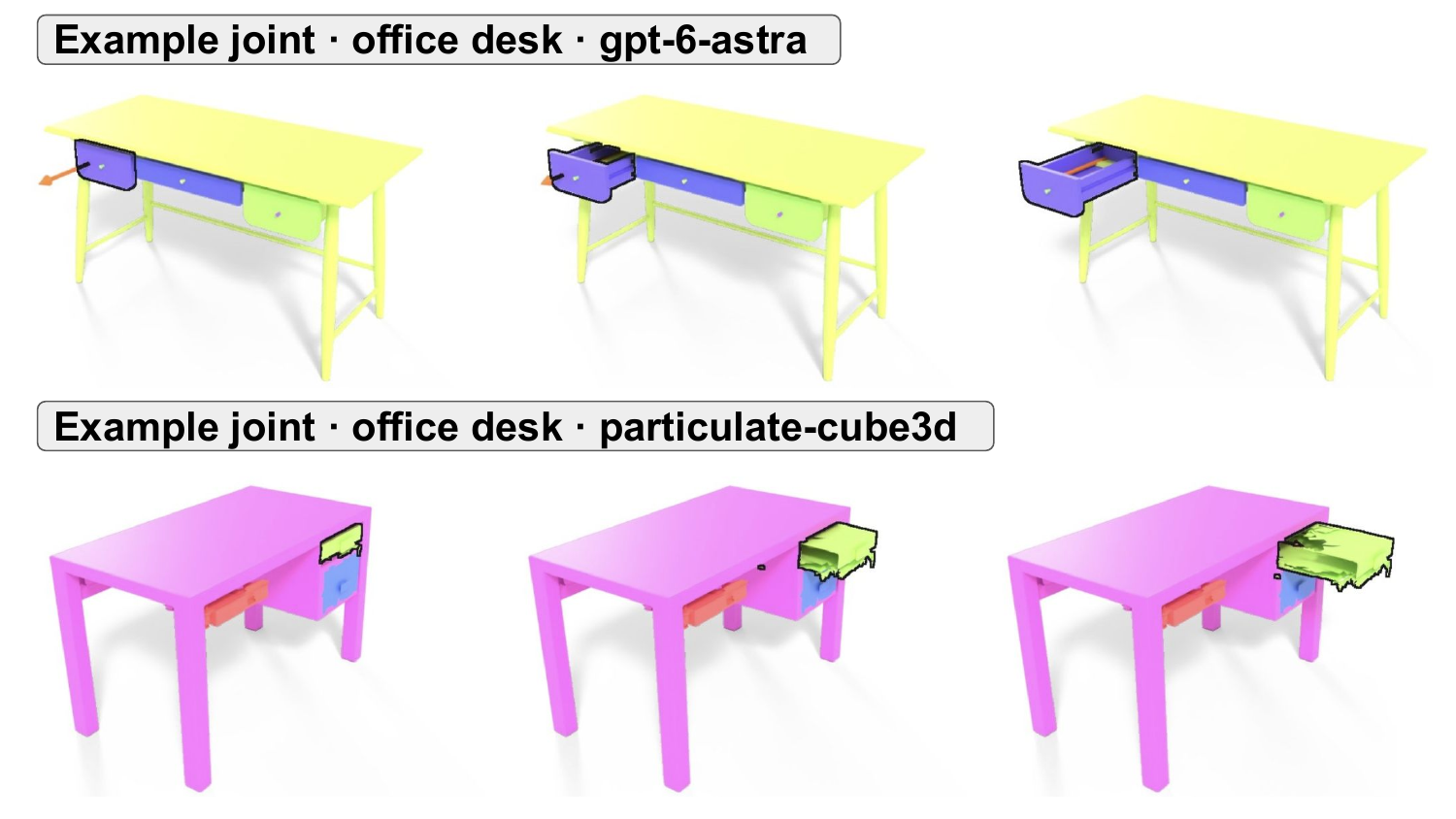}
    \caption{Declared joint motion in office desk examples. Each row shows a generated structure in three poses under a fixed camera. Outlined parts belong to the moving set, and the orange rod marks the declared joint axis.}
    \label{fig:joint-office-desk}
\end{figure}

\begin{figure}[p]
    \centering
    \includegraphics[width=\linewidth]{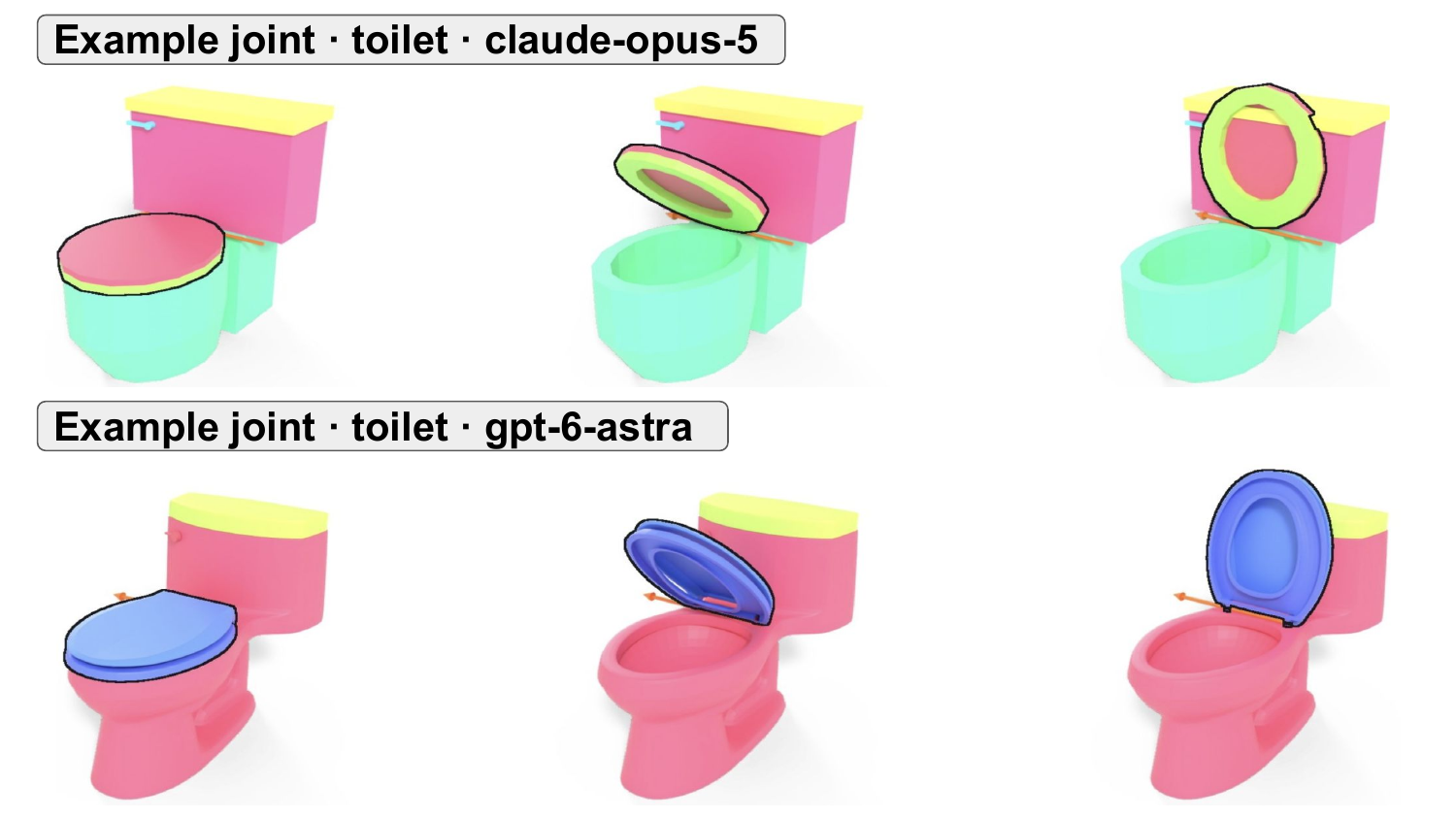}
    \caption{Declared joint motion in toilet examples. Each row shows a generated structure in three poses under a fixed camera. Outlined parts belong to the moving set, and the orange rod marks the declared joint axis.}
    \label{fig:joint-toilet}
\end{figure}

\begin{figure}[p]
    \centering
    \includegraphics[width=\linewidth]{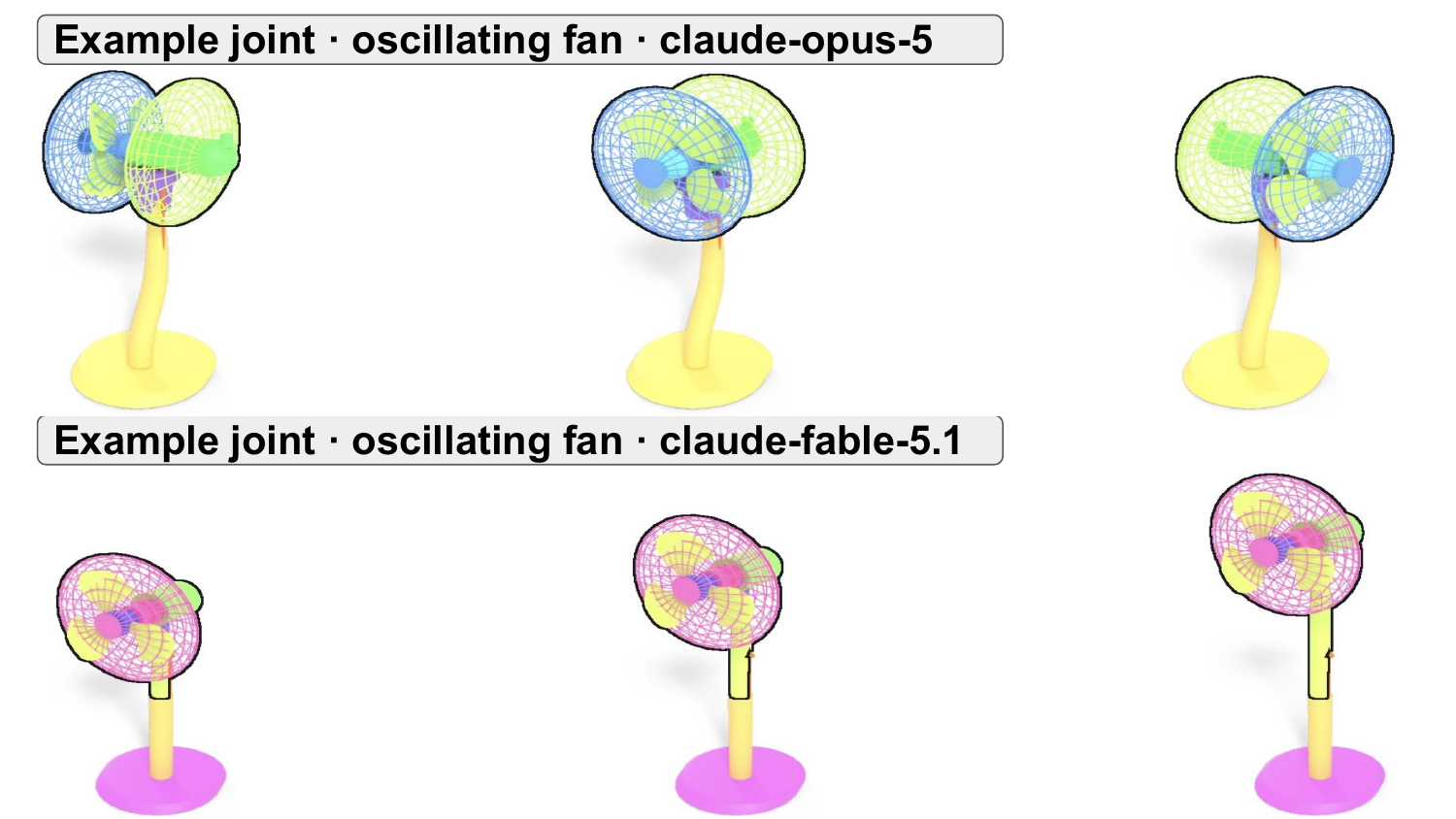}
    \caption{Declared joint motion in oscillating fan examples. Each row shows a generated structure in three poses under a fixed camera. Outlined parts belong to the moving set, and the orange rod marks the declared joint axis.}
    \label{fig:joint-oscillating-fan}
\end{figure}

\begin{figure}[p]
    \centering
    \includegraphics[width=\linewidth]{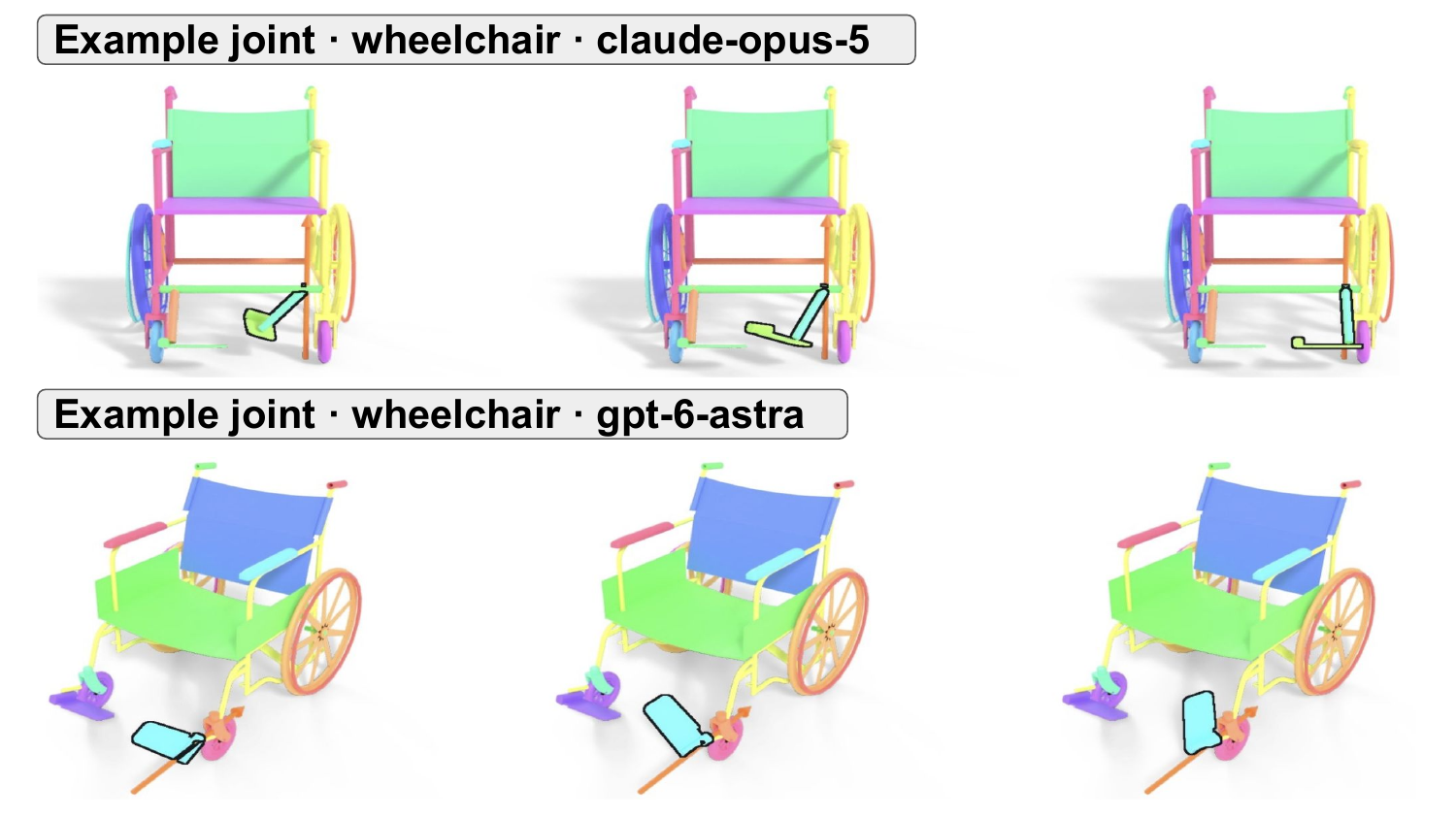}
    \caption{Declared joint motion in wheelchair examples. Each row shows a generated structure in three poses under a fixed camera. Outlined parts belong to the moving set, and the orange rod marks the declared joint axis.}
    \label{fig:joint-wheelchair}
\end{figure}

\begin{figure}[p]
    \centering
    \includegraphics[width=\linewidth]{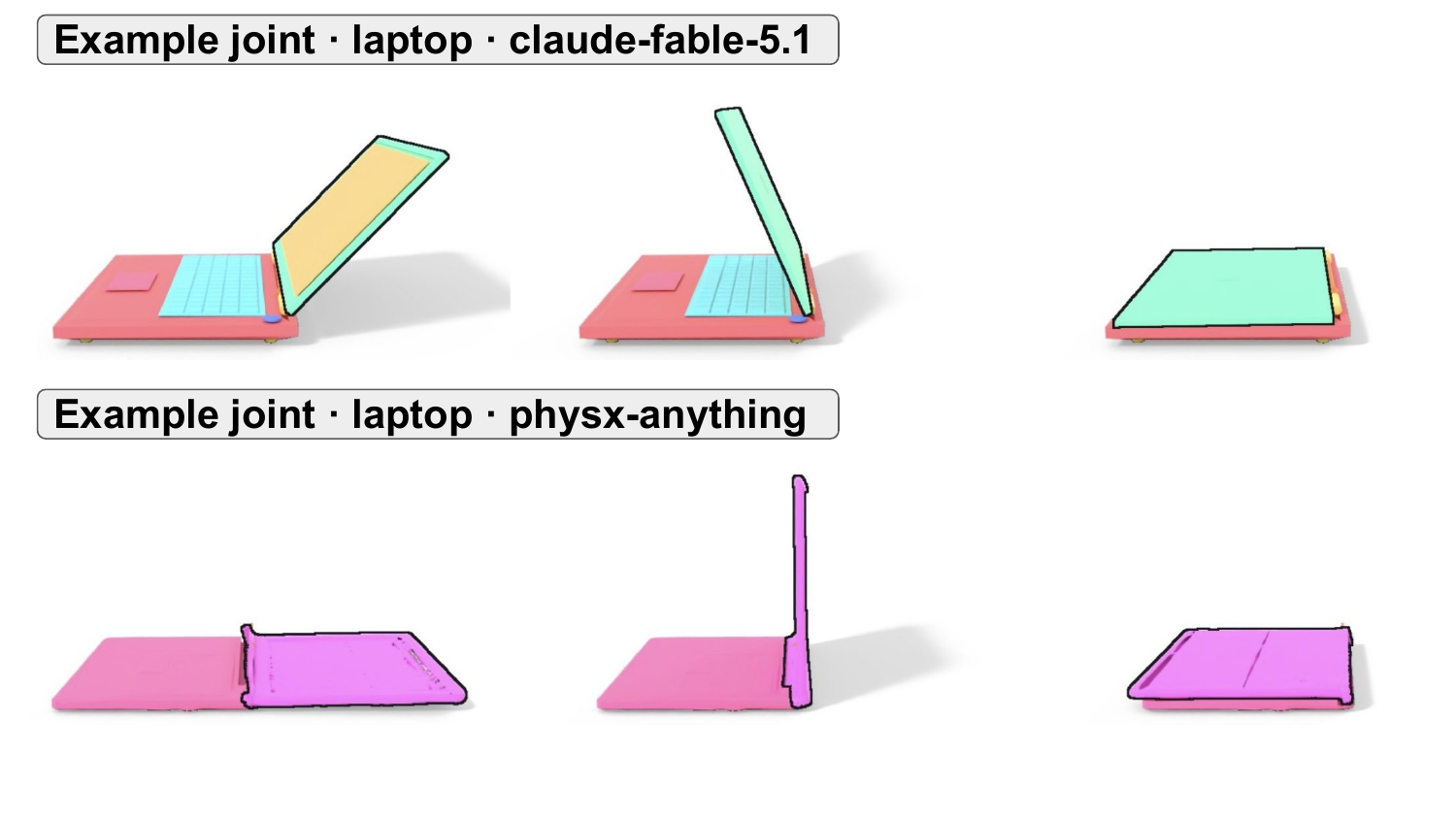}
    \caption{Declared joint motion in laptop examples. Each row shows a generated structure in three poses under a fixed camera. Outlined parts belong to the moving set, and the orange rod marks the declared joint axis.}
    \label{fig:joint-laptop}
\end{figure}

\begin{figure}[p]
    \centering
    \includegraphics[width=\linewidth]{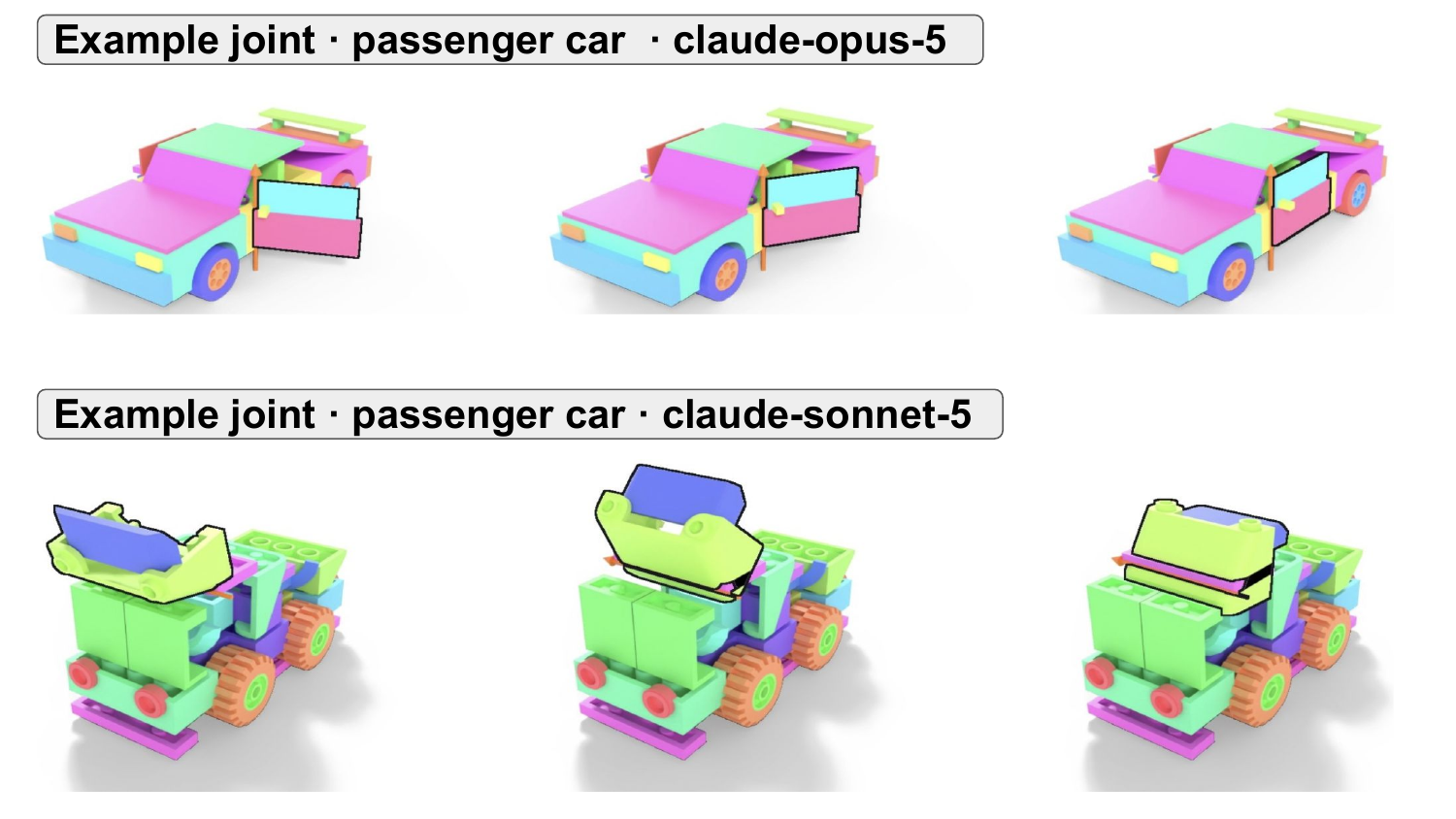}
    \caption{Declared joint motion in passenger car examples. Each row shows a generated structure in three poses under a fixed camera. Outlined parts belong to the moving set, and the orange rod marks the declared joint axis.}
    \label{fig:joint-passenger-car}
\end{figure}

\begin{figure}[p]
    \centering
    \includegraphics[width=\linewidth]{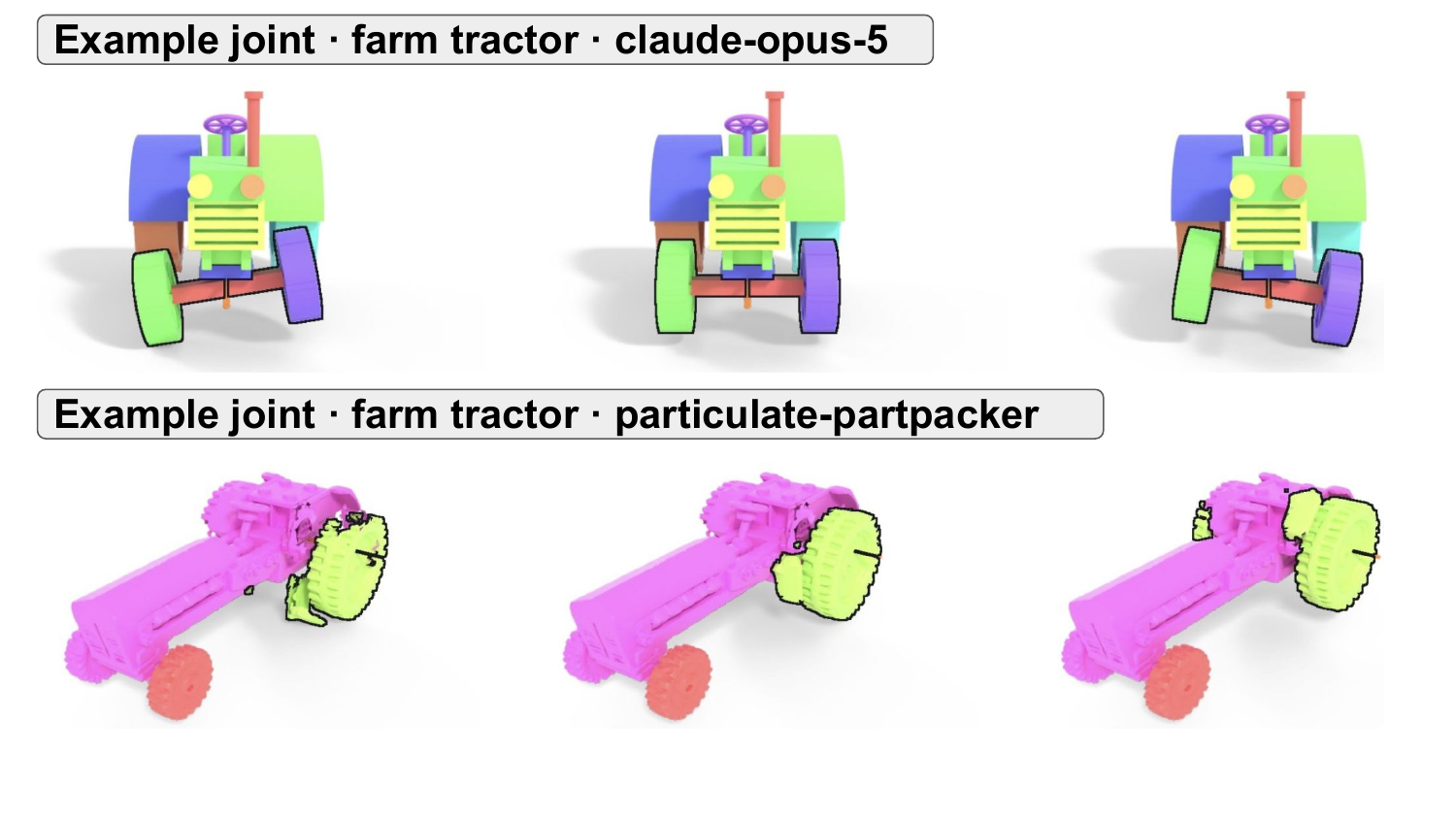}
    \caption{Declared joint motion in farm tractor examples. Each row shows a generated structure in three poses under a fixed camera. Outlined parts belong to the moving set, and the orange rod marks the declared joint axis.}
        \label{fig:joint-farm-tractor}
\end{figure}

\begin{figure}[p]
    \centering
    \includegraphics[width=\linewidth]{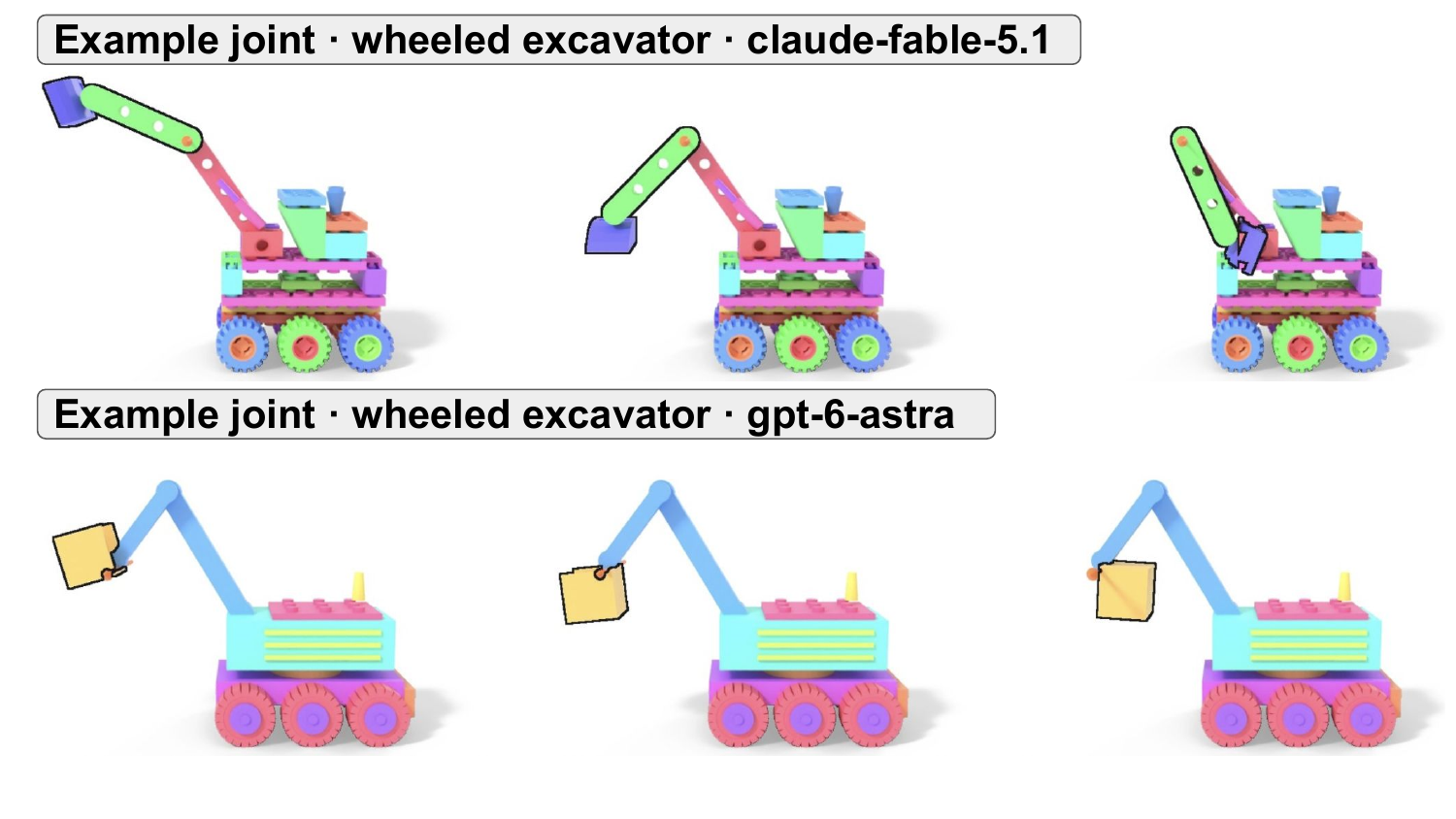}
    \caption{Declared joint motion in wheeled excavator examples. Each row shows a generated structure in three poses under a fixed camera. Outlined parts belong to the moving set, and the orange rod marks the declared joint axis.}
    \label{fig:joint-wheeled-excavator}
\end{figure}

Figures~\ref{fig:joint-office-desk}--\ref{fig:joint-wheeled-excavator} visualize joints declared in generated structures. Each row shows three poses of a generated structure under a fixed camera, illustrating the motion associated with one declared joint. Outlines identify the parts that move with the joint, and the orange rod indicates its declared axis. The moving set is determined by traversing the declared joint graph from the child part without passing through the parent part.

\subsection{Detailed Evaluation Metrics for LMBuild}
\label{app:metrics}

\begin{table*}[t]
\centering
\caption{Illustrative questions used to interpret the LMBuild evaluation metrics, using the construction of an off-road vehicle as a running example.}
\label{tab:evaluation_vehicle_example}

\begingroup
\renewcommand{\arraystretch}{1.25}
\setlength{\tabcolsep}{7pt}

\begin{tabularx}{\textwidth}{
    l
    >{\raggedright\arraybackslash}X
}
\toprule

\textbf{Metric} &
\textbf{Question to Ask for Evaluation} \\

\midrule

\rowcolor{metricSoundness}
\textbf{S.1 Connectivity}
&
Are the wheels, axles, chassis, and other connected components properly attached rather than floating, detached, or disconnected? \\

\cline{2-2}
\rowcolor{metricSoundness}
\textbf{S.2 Collision}
&
Do the wheels and surrounding components avoid unintended interpenetration with suspension, axles, or neighboring parts? \\

\cline{2-2}
\rowcolor{metricSoundness}
\textbf{S.3 Stability}
&
Can the vehicle remain stably supported on the ground without tipping, collapsing, or losing appropriate contact through its wheels? \\

\midrule

\rowcolor{metricAffordance}
\textbf{A.1 Geometry}
&
Are the wheels appropriately shaped, sized, oriented, and positioned so that they can contact the ground and support effective driving? \\

\cline{2-2}
\rowcolor{metricAffordance}
\textbf{A.2 Parts}
&
Does the vehicle contain the functional components required for driving, such as wheels, axles, steering elements, and appropriate supports? \\

\cline{2-2}
\rowcolor{metricAffordance}
\textbf{A.3 Kinematics}
&
Can the wheels rotate around their axles and, where required, steer or articulate along the intended motion axes? \\

\midrule


\cline{2-2}
\rowcolor{metricDesign}
\textbf{D.1 Decomposition}
&
Is the vehicle divided into meaningful granular components rather than overly monolithic or unnecessarily fragmented parts? \\

\cline{2-2}
\rowcolor{metricDesign}
\textbf{D.2 Aesthetics}
&
Does the vehicle exhibit aesthetically elegant proportions, coherent overall form, and sufficiently refined geometry as a standalone design? \\

\cline{2-2}
\rowcolor{metricDesign}
\textbf{D.3 Alignment}
&
Are the wheels, cabin, and other major components arranged consistently with the intended off-road vehicle design or reference? \\

\midrule

\rowcolor{metricRealization}
\textbf{R.1 Sequence}
&
Can the components be assembled in a feasible order without becoming unstable during assembly or blocking subsequent assembly steps? \\

\cline{2-2}
\rowcolor{metricRealization}
\textbf{R.2 Material}
&
Are the materials appropriate for each component's physical role, such as compliant tires and sufficiently rigid and durable structural parts? \\

\cline{2-2}
\rowcolor{metricRealization}
\textbf{R.3 Operability}
&
Once assembled, can the vehicle actually roll and steer through its intended mechanisms, with the required components usable?  \\

\bottomrule
\end{tabularx}

\endgroup
\end{table*}

Unless otherwise specified, all metrics are reported on a normalized scale of $[0,100]$. We use $\operatorname{clip}(x)=\min(1,\max(0,x))$ throughout the definitions below. Our evaluation framework comprises four hierarchical dimensions that assess the \textbf{Structural Soundness}, \textbf{Functional Affordance}, \textbf{Design Quality}, and \textbf{Physical Realization} of the generated structure. A concise overview of the evaluation question addressed by each metric is provided in Table~\ref{tab:evaluation_vehicle_example}. 

\paragraph{Evaluation Framework Positioning.}
Most prior evaluation protocols for generated structures focus primarily on semantic alignment or geometric consistency. In contrast, our framework evaluates buildable structure generation across complementary dimensions of structural soundness, functional affordance, design quality, and physical realization. Importantly, our benchmark spans heterogeneous geometry domains, including LEGO assemblies, mechanical structures, everyday objects, and freely generated meshes, whose representations, part semantics, and physical constraints differ substantially. This cross-domain setting makes many existing domain- or representation-specific metrics difficult to apply uniformly, motivating evaluation criteria that can generalize across structure types. Several metrics necessarily introduce heuristic thresholds, tolerances, discretization choices, or weighting coefficients. These hyperparameters were selected based on geometric and physical reasoning, empirical inspection of representative structures, and practical considerations of the evaluation setting. AI tools were additionally consulted during metric development to help examine candidate formulations, identify potential edge cases, and improve the clarity of the evaluation rules; they were not used to determine benchmark outcomes or tune parameters to individual models or results. The resulting hyperparameters should therefore be interpreted as standardized evaluation parameters rather than theoretically optimal values or exact models of real-world physical behavior. All hyperparameters are fixed before benchmark-wide evaluation and applied identically across systems and outputs. We therefore view the proposed metrics as practical, interpretable, and reproducible comparative proxies rather than definitive solutions. Our goal is to provide a consistent basis for evaluating heterogeneous generated structures while making the assumptions and limitations of each metric explicit. Several dimensions, particularly design quality, human-use geometry, assembly feasibility, and operability, remain difficult to quantify in a fully unbiased and model-independent manner. We hope this framework serves as a useful baseline and motivates further work toward more robust, generalizable, and standardized evaluation protocols for buildable structure generation.


\textbf{Shared Evaluation Conventions.}
Each generated design consists of a collection of triangle-mesh parts represented in a coordinate system, with the vertical axis aligned to $z$. A design is required to additionally provide semantic role for each part, materials, declared joints with their type, axis, origin, and an assembly sequence. Evaluation uses a task-specific reference object together with curated factual claims describing required parts, functional subsystems, kinematics, and attributes extracted from \textbf{Wikipedia} knowledge. Unless otherwise specified, geometric tests use a common voxel representation with pitch $r=4$~mm. For a part $i$, let $G_i$ denote the set of voxels occupied by the part. We define $G_i^{\ominus}$ as its eroded core, obtained by removing the outermost one-voxel-thick boundary of $G_i$; equivalently, a voxel remains in $G_i^{\ominus}$ only when its required neighboring voxels also belong to $G_i$. We define $G_i^{\oplus}$ as its one-cell halo, obtained by expanding $G_i$ outward by one voxel, such that $G_i^{\oplus}$ contains $G_i$ together with its immediately neighboring voxels. Using this representation, two parts form an interfacing pair when

$$
\mathcal{I}
=
\left\{
(i,k): G_i^{\oplus}\cap G_k\neq\varnothing
\right\}.
$$

A declared joint between two incident parts is considered \textbf{origin-valid} when its declared origin lies within \(2\)~cm of the surface of each part. This tolerance accommodates small discrepancies introduced by voxelization, mesh discretization, and coordinate quantization while excluding joint declarations that are spatially inconsistent with the corresponding part interface. We measure proximity to the part surfaces rather than their centroids because physical joints are expected to occur at or near the interface between connected components. We distinguish physical connectivity from mere geometric contact. Two parts are considered \textbf{physically connected} if at least one of three conditions holds: (i) their geometry forms a valid interlocking relation that mechanically constrains separation; (ii) they are connected by a geometrically valid declared fastening relation, such as a hinge, pin, screw, or other supported joint type; or (iii) for brick-based structures, they form a valid stud--receptacle connection. This hybrid definition allows the evaluation to accommodate heterogeneous construction paradigms without treating simple surface contact or proximity as sufficient evidence of attachment. These definitions serve as shared geometric primitives and are reused consistently across the structural, decomposition, assembly, and operability evaluations described below.

All per-design metric scores lies in $[0,1]$ and will be re-sacled to $[0,100]$. Main-table values are reported as the mean score over benchmark tasks. Models which provide empty or failed generations receive a score of zero.

\begin{Metrics_1}{S.1 Connectivity}

Although the target structures in our benchmark are decomposable into multiple constituent parts, the resulting generation should ultimately represent an integrated assembled artifact rather than a collection of independently plausible but disconnected components. Connectivity therefore measures the extent to which the generated parts are physically integrated into a coherent assembly. We first construct an assembly graph in which each generated part corresponds to a node and an edge is introduced only when the relationship between two parts is supported by explicit physical evidence. Such evidence may include a valid declared joint located at a physically plausible contact region, a geometric interlock, a brick--stud connection, or another direct part-to-part attachment supported by the generated geometry. This graph-based representation allows us to distinguish nominally related parts from components that are actually supported by a plausible physical connection. Let $n$ denote the number of generated parts, $m$ the number of connected components in the resulting assembly graph, $C_1$ the largest connected component, $V(C_1)$ the total volume of the parts contained in $C_1$, $V$ the total volume of the generated object, and $n_{\mathrm{fast}}$ the number of physically supported connections identified in the assembly. We compute
\[
S_{1.1}
=
\frac{1}{3}
\left(
\frac{V(C_1)}{V}
+
\left(1-\frac{m-1}{n-1}\right)
+
\min\left(1,\frac{n_{\mathrm{fast}}}{n-1}\right)
\right).
\]

The three terms capture complementary aspects of assembly connectivity. The first term, $V(C_1)/V$, measures the fraction of the total object volume contained in the largest connected component. This volume-weighted formulation captures whether the dominant physical mass of the generated object belongs to a common assembly, while preventing a small detached decorative or auxiliary component from disproportionately reducing the score. The second term, $1-(m-1)/(n-1)$, measures the degree of fragmentation at the part level. A fully connected assembly has $m=1$ and therefore receives the maximum value, while the score decreases as the generated object is divided into an increasing number of disconnected components. In contrast to the volume-based term, this component remains sensitive to the number of disconnected groups even when those groups contain relatively small parts. The final term, $\min(1,n_{\mathrm{fast}}/(n-1))$, measures whether the generated structure contains a sufficient number of connections that are explicitly supported by physical evidence. A connected graph of $n$ nodes requires at least $n-1$ edges; we therefore use $n-1$ as a natural normalization factor. The value is capped at one because additional redundant connections may improve structural redundancy but should not increase a metric whose primary purpose is to assess whether the parts are sufficiently connected to constitute a single assembly. These three signals intentionally address different failure modes. A structure may contain one large connected body while leaving several smaller parts detached, producing a high largest-component score but a lower fragmentation score. Conversely, a structure may appear graph-connected through numerous declared relationships while those relationships are not supported by plausible physical contacts, which is reflected by the physically supported connection term. Their combination therefore evaluates not only whether most of the generated object belongs to a common component, but also whether the assembly is globally integrated and whether its connectivity is grounded in explicit physical evidence. For a valid single-part structure ($n=1$), no inter-part connection is required; we therefore define the connectivity score as one.

\end{Metrics_1}

\begin{Metrics_1}{S.2 Collision}

A physically valid assembly should allow its constituent parts to occupy compatible regions of space without substantial unintended interpenetration. Collision therefore measures whether the generated components can coexist geometrically in a plausible assembled configuration. We do not require meshes to be strictly intersection-free, since small overlaps may arise from mesh discretization, floating-point imprecision, approximate geometric representations, or intentional connection geometry. Instead, we identify only overlaps that are sufficiently large to indicate a likely structural conflict.

For each geometrically relevant pair of parts $(i,k)\in\mathcal{I}$, let $O_{ik}$ denote their overlap volume and $V_i$ and $V_k$ their respective part volumes. We classify the pair as an invalid collision only when both an absolute and a relative overlap criterion are satisfied:
\[
O_{ik}>1\,\mathrm{cm}^{3}
\qquad\text{and}\qquad
\frac{O_{ik}}{\min(V_i,V_k)}>\tau_{ik},
\]
where
\[
\tau_{ik}=
\begin{cases}
0.25, & \text{if a valid joint is declared between parts } i \text{ and } k,\\
0.10, & \text{otherwise}.
\end{cases}
\]

The collision score is then defined as
\[
S_{1.2}
=
1-
\frac{
\left|
\left\{
(i,k)\in\mathcal{I}:
O_{ik}>1\,\mathrm{cm}^{3}
\wedge
\frac{O_{ik}}{\min(V_i,V_k)}>\tau_{ik}
\right\}
\right|
}{
|\mathcal{I}|
}.
\]

The two-stage criterion avoids over-penalizing harmless geometric artifacts. The absolute threshold filters negligible intersections caused by numerical tolerance or mesh approximation, while the relative threshold identifies overlaps that occupy a meaningful fraction of either participating part. A more permissive threshold is used for explicitly jointed parts, since valid connection geometries may naturally introduce localized overlap. This formulation distinguishes plausible attachment geometry from substantial unintended interpenetration, such as components occupying the same physical region or overlapping far beyond what is necessary for a connection. The metric therefore measures the proportion of geometrically plausible part interactions rather than enforcing a strict zero-intersection requirement. When no geometrically relevant part pair exists, we define the collision score as one.

\end{Metrics_1}

\begin{Metrics_1}{S.3 Reliability}

A connected and collision-free structure may still be physically implausible if it cannot maintain a stable configuration under small external perturbations. Reliability therefore evaluates whether the generated artifact can remain upright under quasi-static loading. In this work, we focus specifically on \textbf{tipping stability}, which can be estimated from the relationship between the object's center of mass and its support region without requiring a full rigid-body dynamic simulation. Let $d$ denote the horizontal stability margin between the projection of the object's center of mass and the nearest boundary of its support footprint, $h_c$ the height of the center of mass above the supporting surface, and $h_{\mathrm{top}}$ the total height of the generated structure. A larger support margin relative to the object's height generally indicates greater resistance to tipping.

We derive two complementary stability proxies. The first is the critical tilt angle,
\[
\theta
=
\arctan\left(\frac{d}{h_c}\right),
\]
which approximates the angular perturbation required for the projected center of mass to cross the boundary of the support region. A larger $\theta$ therefore indicates that the object can tolerate greater angular displacement before becoming unstable. The second proxy is
\[
f
=
\frac{d}{h_{\mathrm{top}}},
\]
which corresponds to the horizontal force applied at the top of the structure, expressed as a fraction of the object's weight, required to initiate tipping under quasi-static torque balance. This term captures stability with respect to lateral loading and complements the angular interpretation of $\theta$. Because appropriate stability requirements vary substantially across object categories, we normalize both quantities relative to the corresponding real product rather than imposing one universal target. We define
\[
\theta^\star
=
\operatorname{clip}_{[5^\circ,30^\circ]}
\left(\theta_{\mathrm{ref}}\right)
\]
and
\[
f^\star
=
\operatorname{clip}_{[0.05,0.35]}
\left(f_{\mathrm{ref}}\right),
\]
where $\theta_{\mathrm{ref}}$ and $f_{\mathrm{ref}}$ are computed from the reference structure. The final reliability score is
\[
S_{1.3}
=
\frac{1}{2}
\left(
\operatorname{clip}
\left(
\frac{\theta}{\theta^\star}
\right)
+
\operatorname{clip}
\left(
\frac{f}{f^\star}
\right)
\right).
\]

Reference-relative normalization is important because different artifact categories naturally exhibit different stability profiles. For example, a tall articulated structure and a wide low-profile object should not be expected to share the same tipping margin. Using the corresponding real product as a reference therefore provides a category-sensitive target. At the same time, clipping $\theta^\star$ and $f^\star$ prevents unusually stable or unusually unstable reference objects from defining excessively demanding or permissive evaluation criteria. The two terms also capture complementary failure modes. A structure with a narrow support footprint and a high center of mass may tip under only a small angular perturbation, while a tall structure may also be vulnerable to relatively small lateral forces applied near its top. Averaging the two normalized proxies reduces dependence on any single geometric characteristic and provides a more balanced measure of quasi-static stability. This metric is intentionally limited to tipping reliability. It does not attempt to model sliding, elastic deformation, dynamic impact, material fracture, joint strength, or fatigue, which require more detailed material models and dynamic simulation. These aspects are therefore outside the scope of the current geometric stability evaluation.

\end{Metrics_1}


\textbf{Evaluation Basis for Functional Affordance.}
Functional Affordance evaluates whether a generated structure contains the geometry, constituent components, and articulation needed to support the intended object category. The evaluation is fully deterministic and does not rely on perceptual judging. Instead, its evaluation targets are derived from structured functional knowledge extracted from Wikipedia together with class-level statistics obtained from real object instances. For each task, we construct a component sheet from required-part ($P$), functional-subsystem ($F$), and kinematic ($K$) claims. Canonical component phrases are normalized through tokenization, removal of positional modifiers, singularization, spelling normalization, and merging of equivalent alternatives. Roles in the generated structure are then grounded to these canonical components using deterministic phrase matching, predefined lexical mappings, and component-specific fallback rules. To avoid treating a single reference decomposition as the only valid realization of an object category, we estimate component relevance and expected multiplicity from a population consisting of the task reference and up to eleven additional real instances from the same object class. A component is retained as evaluation-relevant if it occurs in the reference or appears in at least $40\%$ of the available population instances. For plural components, we define the minimum expected multiplicity as

$$
m_c
=
\min
\left(
8,\,
\max
\left(
1,\left\lfloor Q_{25}(\mathrm{count}_c)\right\rfloor
\right)
\right),
$$

where $Q_{25}$ denotes the lower quartile of the observed component counts. Using the lower quartile provides a conservative class-level expectation that accommodates legitimate variation across realizations, while the upper cap of eight prevents highly repetitive components from disproportionately dominating the evaluation. Singular components use $m_c=1$.

\begin{Metrics_2}{A.1 Geometry}

Functional objects must not only contain recognizable components, but should first, exhibit a global geometry consistent with the intended object category. Geometry therefore evaluates correspondence between the generated structure and the task reference at both the whole-object and, when semantic roles are available, component levels. Importantly, the metric is designed to measure shape rather than absolute physical scale. We uniformly sample $20{,}000$ surface points from both the generated design and the reference object. Each point cloud is independently centered using its horizontal bounding-box center and minimum height and then divided by its own bounding-box diagonal. Let $p_\tau$ denote the fraction of generated samples lying within distance $\tau$ of the reference surface and $r_\tau$ the corresponding fraction of reference samples lying within $\tau$ of the generated surface. We compute the bidirectional geometric F-score
\[
F_\tau
=
\frac{2p_\tau r_\tau}{p_\tau+r_\tau}.
\]
To account for ambiguity in the canonical front direction, we evaluate the design under four $90^\circ$ yaw orientations and retain the best whole-object match:
\[
F_{\mathrm{shape}}
=
\max_{\psi\in\{0^\circ,90^\circ,180^\circ,270^\circ\}}
F_{0.025}
\left(
R_z(\psi)D,\,
R
\right).
\]

When the design provides sufficiently informative part roles, we additionally evaluate component-level correspondence. A design is considered role-aware when at least $50\%$ of its parts have non-generic semantic names. For each visible component $c$ represented by at least 20 reference samples, we collect the generated samples belonging to parts whose roles ground to $c$ and compute
\[
F_c
=
F_{0.05}(D_c,R_c).
\]
If the generated design contains no part grounded to component $c$, we set $F_c=0$. Component comparison remains in the common normalized coordinate frame, such that the correct component geometry at an incorrect spatial location is appropriately penalized.

The final score is
\[
A_{2.1}
=
\frac{
0.6F_{\mathrm{shape}}
+
0.4\overline{F_c}
}{
0.6
+
0.4\,
\mathbf{1}
[\text{component term defined}]
}.
\]
Thus, role-less designs are evaluated solely on overall shape, while designs that explicitly expose semantic part structure are additionally evaluated on whether those components occupy geometrically appropriate regions. Created parts and library-retrieved parts are treated identically. The metric therefore does not reward a particular construction mechanism; it evaluates the geometry ultimately produced by that mechanism.

\end{Metrics_2}

\begin{Metrics_2}{A.2 Parts}

Global geometric similarity alone does not establish whether a generated object contains the distinct functional components expected of the target category. The Parts metric therefore evaluates both component coverage and multiplicity, while explicitly allowing valid generated decompositions to differ from the segmentation of any single reference instance. Generated parts are first assigned to canonical components. When an informative semantic role is available and can be grounded to a canonical component, this assignment is made directly from the declared role. Otherwise, we fall back to spatial correspondence: a generated part is assigned to the reference component that covers at least $50\%$ of its surface samples within normalized distance $0.05$, provided that at least five samples are available. This fallback allows systems without explicit semantic labels to remain evaluable while requiring substantial geometric agreement before a component identity is inferred. Because a single functional component may itself be represented by multiple generated parts, parts assigned to the same canonical component are grouped into a common instance when they are connected by fixed joints or when their axis-aligned bounding boxes overlap by more than $20\%$ of the smaller box. This prevents decomposition granularity alone from affecting component counts: for example, a wheel represented by separate rim, tire, and hub geometries may still correspond to a single functional wheel instance. Reference segmentations do not necessarily expose all legitimate internal functional structure. To avoid systematically penalizing such alternative realizations, we additionally identify anonymous interior modules. The generated object is voxelized using

$$
r_{\mathrm{int}}
=
\operatorname{clip}
\left(
\frac{\operatorname{diag}(D)}{150},
0.5\text{ mm},
4\text{ mm}
\right),
$$

and exterior air is recovered through a 6-connected flood fill. An unassigned, non-brick part is treated as interior when at most $5\%$ of its boundary cells are adjacent to exterior air. Parts satisfying this criterion are grouped into anonymous internal modules. This mechanism does not assume that such modules correspond to a particular semantic component; rather, it provides conservative evidence that the generated design contains nontrivial internal structure that may implement a required but unobserved function. For each canonical component $c$, an identified instance receives score $1$ when it participates in the connected assembly and score $1/2$ when all of its constituent parts are floating. This distinction ensures that mere geometric presence is not treated as equivalent to functional integration into the artifact. Given the task-specific minimum expected count $m_c$, the component score is

$$
s_c
=
\frac{
\sum_{j=1}^{m_c}
s^{(j)}_c
}{
m_c
},
$$

where $s^{(j)}_c$ are the $m_c$ highest instance scores. Thus, a design receives full credit only when the required number of component instances is present and integrated, while additional redundant instances do not inflate the score. For a non-visible component for which no explicit instance can be identified, one anonymous interior module may be consumed to provide partial credit of $1/2$. This deliberately conservative treatment acknowledges plausible hidden functional structure without equating an unlabeled internal module with an explicitly identified component.

The final Parts score is

$$
A_{2.2}
=
\frac{2}{3}
\overline{s_c}^{\,\mathrm{visible}}
+
\frac{1}{3}
\overline{s_c}^{\,\mathrm{internal}},
$$

with the weights renormalized when either group is empty. The larger weight on visible components reflects the stronger evidence available for externally identifiable structure, while the internal term preserves sensitivity to hidden functional organization. Overall, the metric is designed to evaluate whether the expected functional parts are present in sufficient number and meaningfully integrated, without requiring the generated artifact to reproduce a particular reference decomposition.

\end{Metrics_2}

\begin{Metrics_2}{A.3 Kinematics}

Many buildable objects are defined not only by the presence and arrangement of their parts, but also by the relative motions that enable their intended function. Kinematics therefore evaluates whether generated moving joints provide the expected motion type, orientation, placement, range, moving-part extent, and collision-free articulation required by the target object. The metric is deliberately stricter than simply checking whether a non-fixed joint has been declared: a useful articulation must be both semantically appropriate and geometrically realizable. Kinematic targets are constructed from two complementary sources. First, each non-fixed reference articulation whose moving child can be grounded to a canonical component defines a reference-based motion target. Revolute, continuous, cylindrical, and screw joints are grouped as rotational motion; prismatic joints define translational motion; and ball joints define spherical motion. Second, each curated $K$ claim whose moving phrase grounds to a canonical component defines a claim-based target. When a claim describes the same articulation as a reference joint, the two are merged into a single target rather than counted independently. Claims associated with non-visible components are treated as \textbf{declared-only}: they may justify a generated joint when computing precision, but they do not introduce an additional recall requirement. This distinction prevents incomplete observability from creating unsupported mandatory motions while still allowing semantically plausible articulations to be recognized. Where sufficient real-instance evidence is available, expected kinematic properties are derived from the population rather than imposed as universal assumptions. These properties include the permissible joint type, whether the motion axis should align with a symmetry or principal axis, whether its direction should be vertical or horizontal, whether the joint should lie near the component center or edge, its expected angular span or translation ratio, and the maximum plausible fraction of the artifact that should move with the joint. When population evidence is insufficient, the corresponding generic expectation is used. This combines category-specific evidence with deterministic fallback rules, allowing the metric to accommodate heterogeneous object classes without requiring a single kinematic template for all structures. For a generated joint $j$, we define joint quality as

$$
q_j
=
q_j^{\mathrm{type}}
q_j^{\mathrm{axis}}
q_j^{\mathrm{offset}}
q_j^{\mathrm{range}}
q_j^{\mathrm{moveset}}
q_j^{\mathrm{free}}.
$$

The multiplicative form is intentional: these terms represent complementary conditions that are jointly required for a useful articulation, such that a severe failure in any one dimension should substantially reduce the validity of the overall motion. The type term is $1$ for a compatible joint, $0.75$ when a cylindrical joint is used for an otherwise compatible rotational or translational target, and $0$ for an incompatible type. Axis consistency is evaluated using

$$
g(e)
=
\operatorname{clip}
\left(
\frac{30^\circ-e}{20^\circ},
0,1
\right),
$$

where $e$ denotes the relevant angular deviation from the expected symmetry, principal, vertical, or horizontal axis. The continuous falloff avoids introducing a brittle binary decision near a single angular threshold while still assigning no credit to substantially misaligned motion axes. Joint placement is evaluated relative to the geometry of the corresponding component. Let $o$ denote the distance from the component-instance center to the joint axis, normalized by the component extent in the relevant direction. A center-oriented joint receives

$$
q_j^{\mathrm{offset}}
=
\operatorname{clip}
\left(
\frac{0.7-o}{0.4},
0,1
\right),
$$

whereas an edge-oriented joint receives

$$
q_j^{\mathrm{offset}}
=
\operatorname{clip}
\left(
\frac{o-0.2}{0.3},
0,1
\right).
$$

Offsets larger than $1.5$ receive zero credit. These two cases capture common functional distinctions such as rotation about a wheel center versus articulation about the edge of a door-like component. The range term evaluates whether the declared motion is sufficiently large to support the intended articulation. For rotational targets,

$$
q_j^{\mathrm{range}}
=
\min
\left(
1,
\frac{\operatorname{span}_j}{\operatorname{span}^{\star}}
\right),
$$

while for translational targets,

$$
q_j^{\mathrm{range}}
=
\min
\left(
1,
\frac{
|\Delta\operatorname{limit}_j|
}{
\rho^\star
L_{\mathrm{move}}
}
\right),
$$

where $\rho^\star$ is the expected travel ratio and $L_{\mathrm{move}}$ is the moving set extent along the joint axis. Ball joints receive full range credit. These ratios measure whether a declared joint merely permits motion in principle or provides a motion extent consistent with the intended function.

The moving set contains the joint child together with geometry reachable through subsequent joints and, for otherwise joint-less parts, physical contact, without traversing through the parent. A motion receives no moving set credit if this set contains the target's relative component or degenerates into motion of the entire object. Otherwise, the score decreases linearly when the moving fraction exceeds the target-specific expected maximum. This term is important because a formally valid joint is not functionally meaningful if it causes both sides of the intended articulation—or nearly the entire artifact to move together. Finally, $q_j^{\mathrm{free}}$ evaluates whether the declared articulation is geometrically realizable rather than merely specified in metadata. We voxel-sweep the moving set through multiple poses over the declared motion range and measure the fraction of poses that do not introduce substantial deep overlap with static geometry. Full rotations use 11 poses at $30^\circ$ increments; other motions use eight poses over their declared interval. A moving set that is already deeply embedded in static geometry at rest receives zero free-motion credit. This discrete sweep is intended as a deterministic geometric feasibility test rather than a full dynamic simulation. We aggregate joint-level measurements through both recall and precision. Each required target takes the best joint quality obtained by each of its component instances, and the target receives the average of its best $m_c$ instances. Averaging over all targets gives recall $R$, which measures whether the expected articulations are successfully realized. For precision, every generated moving joint receives the best quality it achieves on any required target or declared-only claim. A joint unmatched to any such target may still receive half of its quality under the generic expectation for its own joint type, allowing mechanically plausible but unclaimed articulations to receive limited credit without being treated as equivalent to expected motions. Letting $P$ denote the mean score over generated moving joints, the final metric is

$$
A_{2.3}
=
\frac{2RP}{R+P}.
$$

If no moving joint is generated, $P=0$. The harmonic aggregation prevents systems from obtaining a high score either by omitting required articulations or by introducing large numbers of unsupported moving joints. The metric is not applicable to tasks for which no kinematic target exists; in the current benchmark, this only includes: \texttt{bed-frame}, \texttt{dining-chair}, \texttt{dining-table}, and \texttt{step-ladder}.

\end{Metrics_2}


\textbf{Design Quality: Perceptual Evaluation and Human Calibration.}
\label{app:human_annotation}

Design Quality captures aspects of generated structures for which no single deterministic geometric criterion provides an adequate evaluation. We therefore use a \textbf{Calibration-then-Evaluation} protocol: human judgments on a fixed, pre-declared calibration subset are used only to calibrate the scale and relative contribution of automated evaluation signals; the resulting calibration parameters are then frozen and applied uniformly to the remaining benchmark outputs.

Our automatic perceptual signal is obtained from two independent VLM judges, Qwen2.5-VL-32B-Instruct and Gemma-3-27B-IT. Each judge is provided with explicitly defined, metric-specific judging criteria and assigns a raw score $s\in{1,\ldots,10}$ to each criterion. We normalize each raw judgment as $(s-1)/9$. Let $J\in[0,1]$ denote the mean of these normalized scores across the defined criteria and the two VLM judges. Where an objective geometric or programmatic measurement is available, we denote the corresponding computed signal by $C$. Thus, for a given Design Quality metric, the available automated evaluation signals are drawn from $h\in{J,C}$.

Because the raw perceptual and computed signals need not naturally share the same scale as human judgments, each available signal $h$ is independently calibrated. Let $h'$ denote its calibrated value, and let $\ell_h$ and $u_h$ denote the raw values of $h$ that map to calibrated scores $0$ and $1$, respectively. We define

$$
h'
=
\operatorname{clip}
\left(
\frac{h-\ell_h}{u_h-\ell_h},
0,1
\right).
$$

Human calibration targets are collected on a 1--5 rating scale. Let $r\in{1,\ldots,5}$ denote a raw human rating, and let

$$
H=\frac{r-1}{4}
$$

denote its normalized value in $[0,1]$. To determine the calibration range for each automated signal $h$, we fit an affine mapping from $h$ to the human score distribution. Let $b$ denote the fitted scale factor and $a$ the fitted offset. These parameters are obtained by matching the mean and standard deviation of the transformed automated scores to those of the normalized human ratings:

$$
b=\frac{\operatorname{sd}(H)}{\operatorname{sd}(h)},
\qquad
a=\overline H-b\overline h.
$$

The corresponding raw-score endpoints that map to calibrated values $0$ and $1$ are then

$$
\ell_h=-\frac{a}{b},
\qquad
u_h=\frac{1-a}{b}.
$$

Accordingly, $\ell_h$ and $u_h$ define the metric-specific calibration interval for signal $h$.

Calibration is fit on 48 pre-declared items spanning eight objects. When both perceptual and computed signals are available, we combine their calibrated values using nonnegative weights $\alpha$ and $\beta$, where $\alpha$ controls the contribution of the perceptual signal $J'$ and $\beta$ controls the contribution of the computed signal $C'$. The ratio $\alpha:\beta$ is selected on the calibration set using leave-one-object-out mean absolute error (MAE), such that all items associated with the held-out object are excluded when evaluating each candidate blending ratio. After this fitting stage, the calibration ranges and blending weights are fixed and reused unchanged for benchmark evaluation.

The resulting metric-specific calibration parameters are

$$
\begin{array}{c|c|c|c}
\text{Metric}
&
\ell_J\rightarrow u_J
&
\ell_C\rightarrow u_C
&
\alpha:\beta
\\ \hline
D.1 & 0.3611\rightarrow0.7644 & 0.1276\rightarrow1.0247 & 1:1\\
D.2 & 0.3741\rightarrow0.7251 & \text{--} & 1:0\\
D.3 & 0.4500\rightarrow0.7766 & 0.1536\rightarrow0.6057 & 3:1
\end{array}
$$

where $J'$ and $C'$ denote the calibrated perceptual and computed signals obtained from the transformation above. When both signals are available, the final Design Quality metric is

$$
D
=
\frac{\alpha J'+\beta C'}{\alpha+\beta}.
$$

When only one signal is defined, its calibrated value is used directly.

\textbf{Human Annotation Details for Design Evaluation.}
We recruited four university lab volunteers to annotate 90 objects generated by six systems. Annotators provided 1--5 ratings for decomposition quality, aesthetics, and reference alignment. We used human ratings from a subset of eight objects to calibrate our Design evaluation metrics, and evaluated the resulting metrics on a disjoint held-out set, as reported in Table~\ref{tab:human_alignment}. The calibrated metrics show strong agreement with human judgments at the system level, with Spearman correlations of 0.77, 0.83, and 0.94 for decomposition, aesthetics, and alignment, respectively. Item-level correlations are more moderate but remain positive across all three dimensions. Overall, these results suggest that our Design metrics capture human-perceived differences between systems reasonably well, while still leaving room for disagreement at the individual-object level.


\begin{Metrics_3}{D.1 Decomposition}

A buildable structure should be decomposed into components that correspond to meaningful and independently realizable parts, rather than represented as either a single monolithic mesh or an arbitrary collection of fragments. At the same time, many real objects admit multiple valid decompositions, and the segmentation of a single reference instance should not be treated as the unique ground truth. Decomposition therefore combines a perceptual assessment of whether the proposed part organization is meaningful with conservative programmatic signals that capture specific decomposition failure modes. The perceptual and computed signals are calibrated independently using the common Calibration-then-Evaluation protocol described above and are then combined using the pre-determined D.1 blending ratio.

\paragraph{Perceptual decomposition quality.}
To expose the generated part organization to the perceptual judges, each part is rendered using a distinct color so that part boundaries remain visually identifiable. For designs containing at most 40 parts, the decomposition sheet contains two assembled views and two exploded views, with each part displaced outward from the object center by $0.25$ times the object diagonal. For designs containing more than 40 parts, non-exploded front and top views are used instead to avoid excessive visual clutter. The judges additionally receive the task condition image for semantic context and are explicitly informed of the generated part count.

Each judge evaluates three complementary criteria: \textbf{boundary quality}, \textbf{granularity}, and \textbf{separability}. Boundary quality measures whether part divisions occur at structurally or semantically meaningful locations. Granularity measures whether the decomposition uses an appropriate level of detail rather than being excessively fragmented or overly coarse. Separability measures whether the resulting components correspond to plausible independently realizable parts. Let $x_{q,c}\in{1,\ldots,10}$ denote the raw score assigned by judge $q$ to criterion $c$. Let $\mathcal{C}={\mathrm{boundary},\mathrm{granularity},\mathrm{separability}}$ denote the three perceptual criteria, and let $x_{q,c}\in{1,\ldots,10}$ be the score assigned by judge $q$ to criterion $c$. The perceptual decomposition signal is

$$
J_{3.1}
=
\frac{1}{6}
\sum_{q\in\{\mathrm{Qwen},\mathrm{Gemma}\}}
\sum_{c\in\mathcal{C}}
\frac{x_{q,c}-1}{9}.
$$

\paragraph{Programmatic decomposition quality.}
We complement the perceptual assessment with a computed signal $C_{3.1}$ consisting of the programmatic terms that are defined for the generated design. These terms are not intended to prescribe a unique decomposition; instead, they provide conservative evidence for three concrete failure modes: substantial disagreement in decomposition granularity, unnecessary fragmentation of otherwise continuous components, and geometrically unsupported joint placement.

The first term measures reference-relative decomposition granularity. Let $n$ denote the number of generated parts and $n_{\mathrm{ref}}$ the number of parts in the segmented reference. We define

$$
C_{\mathrm{gran}}
=
\frac{
\min(n,n_{\mathrm{ref}})
}{
\max(n,n_{\mathrm{ref}})
}
=
2^{-|\log_2(n/n_{\mathrm{ref}})|}.
$$

This formulation symmetrically penalizes proportional over- and under-decomposition: for example, generating twice as many parts as the reference receives the same score as generating half as many. Importantly, the reference part count serves only as supporting evidence of decomposition granularity rather than as a required target, since alternative physically meaningful realizations may legitimately contain different numbers of components. The second term detects unnecessary fragmentation of a component that otherwise appears continuous. This signal is defined when material declarations are available. Let $V_{\mathrm{split}}$ denote the number of interfacing pairs $(i,k)\in\mathcal{I}$ that simultaneously satisfy three conditions: the two parts ground to the same canonical semantic role, use the same material, and have no declared joint between them. Such a configuration provides evidence that geometry corresponding to a common functional component may have been unnecessarily divided. We define

$$
C_{\mathrm{consol}}
=
\operatorname{clip}
\left(
1-
\frac{
V_{\mathrm{split}}
}{
\max(n-1,1)
},
0,1
\right).
$$

The normalization by $\max(n-1,1)$ scales the penalty relative to the size of the generated decomposition, while clipping ensures that the term remains within the common $[0,1]$ score range even when the number of flagged interfacing pairs exceeds $n-1$. This can occur because a part may interface with multiple neighboring parts. The criterion is intentionally conservative: geometric contact alone is insufficient to identify unnecessary fragmentation. Parts must additionally share the same canonical semantic role and material, and the absence of a declared articulation prevents intentionally movable interfaces from being penalized. The third term evaluates whether declared joints are supported by the corresponding geometric interfaces. Among interfacing part pairs carrying a declared joint, let $N_{\mathrm{decl}}$ denote the number of such declarations and $N_{\mathrm{rej}}$ the number that fail the shared fastening criterion defined in the common evaluation conventions. When at least one such declaration is available, we compute

$$
C_{\mathrm{joint}}
=
1-
\frac{
N_{\mathrm{rej}}
}{
N_{\mathrm{decl}}
}.
$$

This term prevents an implausible decomposition from receiving support merely through arbitrary joint declarations: a declared connection contributes positively only when its placement and relationship to the incident parts are geometrically supported. When no applicable joint declaration exists, this term is treated as unavailable rather than as evidence for or against the decomposition. The raw computed decomposition signal is the mean of the programmatic terms available for the design:

$$
C_{3.1}
=
\operatorname{mean}
\left(
C_{\mathrm{gran}},
C_{\mathrm{consol}},
C_{\mathrm{joint}}
\right)_{\mathrm{available}}.
$$

\paragraph{Calibration and aggregation.}
The perceptual signal $J_{3.1}$ and computed signal $C_{3.1}$ are independently calibrated to the human rating scale using the common calibration procedure described above, producing $J'*{3.1}$ and $C'*{3.1}$. The calibration parameters and blending ratio are determined during the pre-declared calibration stage and remain fixed during benchmark evaluation. As established by this procedure, D.1 assigns equal weight to the perceptual and computed signals:

$$
D_{3.1}
=
\frac{
J'_{3.1}
+
C'_{3.1}
}{2}.
$$

When only one signal is available, its calibrated value is used directly. This hybrid formulation deliberately avoids treating the segmentation of a single reference object as the unique correct decomposition. Instead, the perceptual component evaluates whether the proposed organization forms meaningful and independently realizable parts, while the programmatic component provides complementary evidence for specific structural failure modes, including severe granularity mismatch, unnecessary fragmentation, and unsupported joint placement.

\end{Metrics_3}

\begin{Metrics_3}{D.2 Aesthetics}

Aesthetics evaluates the intrinsic visual quality of the generated structure independently of its exact correspondence to a particular reference instance. This distinction is important because buildable objects may admit substantial variation in valid design: a generated artifact may differ from the reference while still exhibiting well-resolved proportions, coherent form, and high geometric refinement. Conversely, approximate reference similarity does not by itself imply a visually well-resolved design. To isolate geometric form from superficial appearance, all generated components are rendered using the same neutral gray material. Colors, textures, semantic labels, and material assignments therefore do not contribute to this metric, and individual part boundaries are intentionally deemphasized. The judges receive only the generated clay renderings and are not shown the reference object or task condition image. Aesthetics is therefore evaluated as a reference-free property of the generated design. Each judge evaluates three criteria: \textbf{proportion}, \textbf{coherence}, and \textbf{refinement}. Proportion measures whether the relative dimensions and scales of the visible forms produce a plausible and balanced whole. Coherence measures whether the geometry reads as a unified design rather than as an arbitrary collection of independently generated shapes. Refinement measures whether the major forms, transitions, and characteristic geometric details are sufficiently resolved rather than remaining coarse or unfinished. Let

$$
\mathcal{Q}=\{\mathrm{Qwen},\mathrm{Gemma}\},
\qquad
\mathcal{C}_{\mathrm{aest}}
=
\{\mathrm{proportion},\mathrm{coherence},\mathrm{refinement}\},
$$

and let $x_{q,c}\in{1,\ldots,10}$ denote the score assigned by judge $q\in\mathcal{Q}$ to criterion $c\in\mathcal{C}_{\mathrm{aest}}$. The raw perceptual signal is

$$
J_{3.2}
=
\frac{1}{6}
\sum_{q\in\mathcal{Q}}
\sum_{c\in\mathcal{C}_{\mathrm{aest}}}
\frac{x_{q,c}-1}{9}.
$$

Unlike D.1 and D.3, Aesthetics deliberately contains no separate programmatic component. Simple geometric proxies such as reference overlap, symmetry, or part count would impose assumptions about what a visually successful design should look like and would partially conflate intrinsic design quality with reference fidelity. D.2 therefore follows the same Calibration-then-Evaluation protocol using only its perceptual signal: $J_{3.2}$ is calibrated to the human rating scale using the pre-declared calibration set, and the resulting calibration parameters are subsequently frozen for benchmark evaluation. The final score is

$$
D_{3.2}
=
J'_{3.2}.
$$

This metric is consequently reference-free by design. It rewards generated artifacts whose geometry is proportionally plausible, visually integrated, and sufficiently refined, without requiring reproduction of the particular design choices present in the benchmark reference.

\end{Metrics_3}

\begin{Metrics_3}{D.3 Structural Alignment}

Structural Alignment evaluates whether the generated artifact preserves the characteristic organization and design-defining structure expressed by the task condition. Unlike D.2 Aesthetics, which evaluates the generated object independently of any reference, D.3 is explicitly reference-conditioned. At the same time, it does not require exact geometric reconstruction: multiple valid realizations may preserve the same characteristic design while differing in detailed geometry, construction strategy, or decomposition. We therefore combine perceptual reference-based assessment with conservative programmatic signals of structural correspondence. D.3 is also intentionally distinct from A.1 Geometry. A.1 primarily measures general geometric correspondence at the whole-object and component levels, asking whether the generated geometry broadly matches the intended object and occupies appropriate spatial regions. In contrast, D.3 focuses on finer-grained, design-specific correspondence that is difficult to capture with geometry-only rules, including whether characteristic components are represented, whether their relative arrangement and scale preserve the target design, and whether distinctive structural details remain recognizable. These properties often depend on semantic and perceptual interpretation rather than distance alone. Accordingly, D.3 uses VLM judgment as its primary signal, while programmatic measurements provide complementary geometric evidence rather than replacing perceptual assessment.

\paragraph{Perceptual structural alignment.}
For the perceptual component, the judges receive both the task condition image and the neutral gray clay rendering of the generated structure. Each judge evaluates four criteria: \textbf{inventory}, \textbf{placement}, \textbf{relative scale}, and \textbf{detail fidelity}. Inventory measures whether the major characteristic components of the target are represented. Placement measures whether these components occupy appropriate relative locations. Relative scale measures whether their sizes and proportions preserve the target organization. Detail fidelity measures whether distinctive geometric features and structural details remain recognizable in the generated design. Let

$$
\mathcal{Q}=\{\mathrm{Qwen},\mathrm{Gemma}\},
\qquad
\mathcal{C}_{\mathrm{align}}
=
\{\mathrm{inventory},\mathrm{placement},\mathrm{scale},\mathrm{detail}\},
$$

and let $x_{q,c}\in{1,\ldots,10}$ denote the score assigned by judge $q\in\mathcal{Q}$ to criterion $c\in\mathcal{C}_{\mathrm{align}}$. The raw perceptual signal is

$$
J_{3.3}
=
\frac{1}{8}
\sum_{q\in\mathcal{Q}}
\sum_{c\in\mathcal{C}_{\mathrm{align}}}
\frac{x_{q,c}-1}{9}.
$$

\paragraph{Programmatic structural alignment.}
We complement the perceptual assessment with up to two programmatic signals. These signals are intended as supporting evidence for structural correspondence and do not impose exact reconstruction of the reference.

The first signal measures part-level volumetric correspondence. The generated design and reference object are independently normalized for scale using their respective bounding-box diagonals and placed in the common normalized comparison frame. Each generated and reference part is then voxelized from surface samples followed by hole filling, using a voxel pitch equal to $1\%$ of the corresponding object's diagonal. For each reference part $r$, we identify the generated part $p$ with maximum volumetric intersection-over-union. The resulting score is

$$
C_{\mathrm{part}}
=
\frac{
\sum_r
|r|
\max_p
\operatorname{IoU}(r,p)
}{
\sum_r |r|
},
$$

where $|r|$ denotes the occupied volume of reference part $r$.

The matching is intentionally many-to-one rather than one-to-one. This allows alternative valid decompositions in which multiple reference segments are represented by a single generated component. Weighting by reference-part volume gives greater influence to major structural components while preventing very small decorative pieces from dominating the score. No additional rotation search is performed for this term, so correspondence is evaluated in the shared task-aligned coordinate frame.

The second programmatic signal evaluates consistency in global mirror symmetry when the corresponding reference symmetry measurement is available. For each object, occupied geometry is represented on a $1$~cm voxel grid. We evaluate 36 vertical mirror planes passing through the object's centroid, separated by $5^\circ$ increments, and define the object's symmetry value as the maximum IoU between its occupied cells and their reflected counterparts. Let $\sigma\in[0,1]$ and $\sigma_{\mathrm{ref}}\in[0,1]$ denote the resulting symmetry values for the generated and reference structures, respectively. We define

$$
C_{\mathrm{sym}}
=
1-
\frac{
|\sigma-\sigma_{\mathrm{ref}}|
}{
\max
\left(
\sigma_{\mathrm{ref}},
1-\sigma_{\mathrm{ref}}
\right)
}.
$$

Importantly, $C_{\mathrm{sym}}$ does not reward symmetry itself. It measures agreement with the symmetry characteristics of the target. A strongly asymmetric reference therefore does not require the generated artifact to become symmetric, whereas a highly symmetric reference provides evidence that comparable global organization is characteristic of the target design.

The raw computed signal is the mean of whichever programmatic terms are available:

$$
C_{3.3}
=
\operatorname{mean}
\left(
C_{\mathrm{part}},
C_{\mathrm{sym}}
\right)_{\mathrm{available}}.
$$

For tasks where reference geometry is available but no valid symmetry measurement is provided, $C_{\mathrm{part}}$ remains defined and constitutes the computed component on its own.

\paragraph{Calibration and aggregation.}
The perceptual signal $J_{3.3}$ and computed signal $C_{3.3}$ are independently calibrated to the human rating scale using the common Calibration-then-Evaluation procedure, producing $J'*{3.3}$ and $C'*{3.3}$. The calibration parameters and blending ratio are determined during the pre-declared calibration stage and remain fixed during benchmark evaluation. As established by this procedure, D.3 assigns a $3{:}1$ perceptual-to-computed weighting:

$$
D_{3.3}
=
\frac{
3J'_{3.3}
+
C'_{3.3}
}{
4
}.
$$

This weighting reflects the role of D.3 as a fine-grained structural and design-alignment measure: perceptual judgment captures semantic organization and distinctive design features that are difficult to express through deterministic geometric rules alone, while the computed component provides complementary evidence of part-level correspondence and, where available, symmetry consistency. In this way, D.3 complements rather than duplicates A.1: A.1 measures broad geometric correspondence, whereas D.3 evaluates whether the characteristic structural identity of the specific target design is preserved.

\end{Metrics_3}

\begin{table}[t]
\centering
\small
\setlength{\tabcolsep}{4pt}
\begin{tabular}{lcccccc}
\toprule
 & \multicolumn{2}{c}{\textbf{Decomposition}} & \multicolumn{2}{c}{\textbf{Aesthetics}} & \multicolumn{2}{c}{\textbf{Alignment}} \\
\cmidrule(lr){2-3} \cmidrule(lr){4-5} \cmidrule(lr){6-7}
\textbf{System} & Human & Ours & Human & Ours & Human & Ours \\
\midrule
Claude Sonnet 5 & 92.9 \scriptsize{(1)} & 84.9 \scriptsize{(1)} & 76.8 \scriptsize{(2)} & 79.6 \scriptsize{(1)} & 80.4 \scriptsize{(1)} & 74.4 \scriptsize{(2)} \\
GPT-6 Astra     & 83.9 \scriptsize{(2)} & 80.0 \scriptsize{(2)} & 85.7 \scriptsize{(1)} & 72.0 \scriptsize{(3)} & 78.6 \scriptsize{(2)} & 78.9 \scriptsize{(1)} \\
PartPacker      & 69.6 \scriptsize{(3)} & 59.4 \scriptsize{(5)} & 60.7 \scriptsize{(3)} & 78.1 \scriptsize{(2)} & 60.7 \scriptsize{(3)} & 71.4 \scriptsize{(3)} \\
Qwen3.5-27B     & 50.0 \scriptsize{(5)} & 76.6 \scriptsize{(3)} & 44.6 \scriptsize{(4)} & 66.0 \scriptsize{(4)} & 46.4 \scriptsize{(4)} & 65.8 \scriptsize{(4)} \\
Gemma-4-31B     & 60.7 \scriptsize{(4)} & 60.2 \scriptsize{(4)} & 32.1 \scriptsize{(5)} & 54.9 \scriptsize{(5)} & 37.5 \scriptsize{(5)} & 62.0 \scriptsize{(5)} \\
BrickGPT        &  3.6 \scriptsize{(6)} & 13.0 \scriptsize{(6)} & 23.2 \scriptsize{(6)} & 12.2 \scriptsize{(6)} &  8.9 \scriptsize{(6)} & 35.8 \scriptsize{(6)} \\
\midrule
System Spearman $\rho$ $\uparrow$ & \multicolumn{2}{c}{0.77} & \multicolumn{2}{c}{0.83} & \multicolumn{2}{c}{0.94} \\
System MAE $\downarrow$           & \multicolumn{2}{c}{9.8}  & \multicolumn{2}{c}{14.8} & \multicolumn{2}{c}{14.6} \\
Item Spearman $\rho$ $\uparrow$   & \multicolumn{2}{c}{0.52} & \multicolumn{2}{c}{0.60} & \multicolumn{2}{c}{0.64} \\
Human--human $\rho$ (ref.)        & \multicolumn{2}{c}{0.69} & \multicolumn{2}{c}{0.52} & \multicolumn{2}{c}{0.73} \\
\bottomrule
\end{tabular}
\caption{Alignment between calibrated Design (D) scores and human ratings on 7 held-out objects (6 systems, 42 designs per dimension; 4 annotators). Cells report system mean scores with ranks in parentheses. We report system-level Spearman $\rho$ and MAE, item-level Spearman $\rho$, and human--human $\rho$ as a reference for annotator agreement.}
\label{tab:human_alignment}
\end{table}


\textbf{Physical Realization: Evaluation Basis.}
The preceding metric levels evaluate whether a generated structure is geometrically sound, contains the functional components expected of the target object, and exhibits coherent design quality. Physical Realization asks a complementary question: whether the generated specification provides a plausible path from a digital assembly to an artifact that could be constructed and used. We evaluate three complementary aspects. \textbf{Assembly Sequence} examines whether the parts can be inserted in the declared order while maintaining physically supportable intermediate states. \textbf{Material} evaluates whether declared material choices are compatible with the functional role of each component. \textbf{Operability} evaluates whether the completed structure supports the required functional subsystems, articulated motions, and human-use conditions.

These metrics therefore evaluate information contained in the generated assembly specification rather than only the final rendered geometry. Missing assembly instructions, missing materials, absent semantic roles, or unsupported articulation may reduce realization performance even when the final shape appears visually plausible.

\begin{Metrics_4}{R.1 Assembly Sequence}

A physically plausible final geometry is not necessarily assemblable. A component may be trapped by previously placed geometry, an intermediate subassembly may remain unsupported or become unstable before later components are added, or the declared construction sequence may omit parts that appear in the final design. Assembly Sequence therefore evaluates both whether the declared ordering admits physically feasible insertion operations and whether the resulting intermediate configurations remain supportable throughout construction. Let the declared sequence contain $m$ placement operations,

$$
\{(p_k,d_k)\}_{k=1}^{m},
$$

where $p_k$ denotes the component placed at step $k$ and $d_k$ its declared insertion direction. Let $n$ denote the total number of parts in the final generated design, including any parts omitted from the declared sequence. A design for which no usable assembly sequence is provided receives a score of zero. After each placement, the currently assembled structure is partitioned into connected subassemblies according to the shared \textbf{physical-connectivity} relation defined above, which recognizes geometrically interlocked parts, valid fastening relations, and valid brick--stud connections. Mere geometric contact alone is therefore insufficient to merge two parts into the same subassembly. Each step is then evaluated along two dimensions: \textbf{insertability}, which tests whether the newly introduced component can reach its final pose without passing through previously assembled geometry, and \textbf{intermediate support}, which tests whether the resulting configuration remains physically grounded and, where applicable, stable against tipping.

\paragraph{Insertability.}
For component $p_k$, the declared direction $d_k$ is evaluated first as the preferred insertion hypothesis. We additionally test the fixed canonical directions

$$
-z,\quad \pm x,\quad \pm y,\quad +z
$$

as deterministic fallbacks. Thus, the metric evaluates whether the declared assembly \emph{order} admits a physically plausible insertion path, rather than requiring exact agreement with a single insertion-direction annotation. For each candidate direction, the component is swept from outside the current object extent toward its final pose in discrete odd-cell increments over a search distance extending to the object bounding-box diagonal plus $5$~cm. Let $O_t$ denote its voxel overlap with previously placed geometry at an intermediate sweep position and $O_{\mathrm{final}}$ the corresponding overlap at the final pose. The insertion path is accepted only when

$$
O_t
\le
O_{\mathrm{final}}+2
$$

throughout the sweep. This relative criterion tolerates small discretization or interface overlap already present in the intended final configuration while rejecting paths that require the component to pass substantially through existing geometry. Because the supporting floor is not represented as collision geometry, upward insertion directions with vertical component $d_z>0.3$ are excluded when the component's final pose lies above the ground-contact band. This prevents artificial assembly paths in which a component would effectively enter the structure from below the supporting surface. The first component in the sequence is considered insertable by definition.

\paragraph{Intermediate support and stability.}
After a successful insertion test, the connected subassembly containing the newly placed component must reach the common ground-contact band,

$$
z_{\min}
\le
z_0+\beta,
$$

where $z_0$ and $\beta$ follow the same ground-contact convention used in S.3. For free-standing tasks, each intermediate subassembly must additionally satisfy positive quasi-static tipping stability. Let $X_k$ denote the set of connected subassemblies present after step $k$, let $\theta_B$ denote the critical tilt angle of subassembly $B\in X_k$, and let $\theta^\star$ denote the task-specific stability target defined in S.3. We define the intermediate-state margin as

$$
s_k
=
\min_{B\in X_k}
\min
\left(
1,
\frac{\theta_B}{\theta^\star}
\right).
$$

The minimum over subassemblies reflects the fact that a single unsupported or unstable intermediate component is sufficient to compromise the current assembly state. If step $k$ is not executable, or if any resulting subassembly is floating or self-tipping, we set

$$
s_k=0.
$$

For tasks whose references are not free-standing, the stability requirement reduces to binary grounding, and therefore $s_k\in{0,1}$. A component associated with a failed step remains in the simulated assembly, and later declared steps are still evaluated. This is intentional: the executable-prefix term described below records the first break in sequential feasibility, whereas the intermediate-state term retains information about whether subsequent portions of the declared construction trajectory remain locally plausible.

\paragraph{Aggregation.}
Let $k^\star$ denote the number of consecutive assembly steps successfully completed before the first failure. The executable-prefix score is

$$
E
=
\frac{k^\star}{n}.
$$

The intermediate-state score is

$$
S
=
\frac{1}{n}
\sum_{k=1}^{m}
s_k.
$$

Both terms are normalized by the total number of parts $n$ in the generated design rather than by the number of declared steps $m$. Consequently, omitting difficult components from the assembly sequence cannot improve the score.

The final Assembly Sequence metric is

$$
R_{4.1}
=
\frac{1}{2}E
+
\frac{1}{2}S.
$$

The two terms capture complementary aspects of constructability: $E$ measures how much of the intended artifact can be assembled sequentially before the first infeasible operation, while $S$ measures the quality and supportability of the intermediate configurations encountered across the declared construction process.

\end{Metrics_4}

\begin{Metrics_4}{R.2 Material}

Physical realization additionally requires material assignments that are compatible with the structural and functional role of each component. Because real products frequently admit multiple valid material choices, Material does not evaluate exact agreement with the material of a single reference model. Instead, for each component we construct a set of acceptable material classes from multiple independent evidence sources and evaluate whether the generated assignment is compatible with that set. This formulation is intentionally permissive with respect to legitimate material variation while remaining restrictive toward assignments that are unsupported by either documented usage, observed population statistics, reference correspondence, or functional plausibility. Our material library contains 23 materials organized into nine classes. For compatibility reasoning, related classes are additionally grouped into broader families, including metal, polymer, soft material, wood, glass, and ceramic. For a generated part $i$ with canonical semantic role $r_i$, the accepted material set is

$$
\mathcal{A}_i
=
W(r_i)
\cup
P(r_i)
\cup
R(i),
$$

where $W$, $P$, and $R$ denote textual evidence, population statistics, and reference-overlap evidence, respectively. We use the union because the three sources are complementary and individually incomplete: a material may be well supported by one source while being absent from another simply because that source does not exhaust all valid realizations. Requiring agreement across sources would therefore be unnecessarily restrictive. If this union is empty, we instead use the functional material judgment

$$
\mathcal{A}_i
=
J(r_i).
$$

The first source, $W(r_i)$, is derived from verified material claims in the cited Wikipedia sources. We retain only claims that explicitly name the corresponding component. Per-part material extraction is performed from the cited text, and an extracted answer is accepted only when accompanied by a verbatim supporting quotation containing each relevant material expression. This requirement makes the textual evidence auditable and prevents unsupported material inference. Material claims associated with clearly historical usage, such as descriptions explicitly marked by historical years or terms such as `originally'' or `traditionally,'' are excluded so that the metric better reflects plausible contemporary construction rather than historically possible but currently atypical realizations. The second source, $P(r_i)$, is a population prior estimated from Artiverse instances in the corresponding object category. A material class is included when it accounts for at least $10\%$ of the observed class mentions for the role, provided that the role has at least five observations. The $10\%$ threshold is intended to retain recurring alternatives while excluding isolated or exceptionally rare assignments, and the minimum observation count avoids drawing population-level conclusions from insufficient evidence. This permits multiple commonly observed materials to be treated as valid rather than forcing the design toward a single modal material. The third source, $R(i)$, transfers material evidence from geometrically corresponding reference components. The generated and reference objects are independently normalized by their diagonals and voxelized at a pitch equal to $2\%$ of the corresponding diagonal. For every reference component that covers at least $25\%$ of the generated part's occupied cells, we add the reference component's interior and exterior material classes together with the population prior associated with its semantic role. The overlap threshold requires substantial geometric correspondence before material evidence is transferred, reducing the chance that incidental contact or weak spatial proximity incorrectly determines the acceptable material set. When none of these three sources provides an applicable class, we use $J(r_i)$, a fixed task-and-part functional material table. This fallback represents material classes that are functionally plausible for the corresponding component role. It is used only when evidence from text, population statistics, and reference correspondence is unavailable, thereby providing coverage without allowing the fallback to override stronger task-specific evidence. Let $c_i$ denote the library material class assigned to generated part $i$. Its per-part compatibility score is

$$
a_i
=
\begin{cases}
1,
&
c_i\in\mathcal{A}_i,
\\[4pt]
0.5,
&
c_i
\text{ belongs to the same material family as a class in }
\mathcal{A}_i,
\\[4pt]
0,
&
\text{otherwise}.
\end{cases}
$$

A part with no declared library material receives zero whenever an evaluation basis exists. Exact class agreement therefore receives full credit, while a materially related choice receives partial credit. The family-level score of $0.5$ should be interpreted as coarse compatibility rather than equivalence: it acknowledges that materials within the same broad family may share relevant physical characteristics while still distinguishing them from explicitly supported classes. For example, two metal subclasses may be treated as more similar than a metal and a textile even when the exact preferred class differs. The final Material score is

$$
R_{4.2}
=
\frac{
1
}{
\left|
\left\{
i:
\mathcal{A}_i\neq\varnothing
\right\}
\right|
}
\sum_{
i:
\mathcal{A}_i\neq\varnothing
}
a_i.
$$

Parts for which no evaluation basis can be established are excluded rather than automatically treated as incorrect. This conservative treatment avoids penalizing the generator when the benchmark itself lacks sufficient evidence to determine material suitability. If no part has an available material basis, the metric is considered not applicable. Conversely, a generated design in which no part declares a material from the supported library receives a score of zero. Overall, this evidence-based formulation is designed to balance two competing requirements: it accommodates genuine material diversity across real products while still rejecting choices that are incompatible with the structural or functional role of the corresponding component. The numerical thresholds used for evidence inclusion and partial compatibility are fixed benchmark-wide heuristic choices, selected to provide stable and interpretable behavior across heterogeneous object categories rather than to reproduce any single reference construction exactly.

\end{Metrics_4}

\begin{Metrics_4}{R.3 Operability}

A completed structure may be geometrically plausible and physically assemblable while still failing to support its intended use. Operability therefore evaluates whether the generated artifact can realize the task-level capabilities associated with the corresponding real object. Rather than treating functionality as a single visual or geometric property, each capability is represented as an explicit combination of required components, structural relationships, articulated behavior, and, where applicable, human-use constraints. R.3 consequently evaluates whether these individually necessary conditions are integrated into a structure that can plausibly operate as intended.

\paragraph{Capability specification.}
For each task, we construct a fixed capability sheet from the curated functional and kinematic claims together with a pre-declared human-use layer. The capability specification is task-level and system-independent: it is defined independently of the generated outputs and applied uniformly to all evaluated systems. Capabilities draw from four sources. Functional capabilities, denoted $F$, correspond to required functional subsystems of the object. Motion capabilities, denoted $M$, represent articulated behaviors required by the task. Human-use capabilities, denoted $H$, capture geometric requirements for interaction, including reachability, clearance, graspability, and usable support surfaces. Finally, $P$ contains required components not already represented by a functional subsystem. A functional subsystem may consist of multiple required links. Each named component must be present, and explicitly specified component relationships must be supported by a contact or joint path of at most four hops. When the subsystem includes a control component, the control must additionally be reachable from the estimated user position and connected through a path of at most six hops, containing an appropriate moving joint, to as many as three associated output components. The slightly longer control path accommodates mechanisms in which user input is transmitted through intermediate linkage before reaching the functional output.

\paragraph{Component and structural requirements.}
Component presence receives score $1$ when a generated role matches the required component, including accepted lexical relations and components rigidly fixed to the corresponding matched part, and $0$ otherwise. For a required relationship between two components, a structural path receives full credit when an appropriate path exists through a moving joint whose moving set contains the required endpoint component. When the required components are structurally connected but the expected articulation is absent, the relationship receives partial credit of $1/2$. If no valid structural path exists, the score is $0$. This distinction separates structural presence from functional articulation: connectivity alone provides evidence of partial realization, but full credit requires the motion mechanism needed by the capability.

\paragraph{Human-use geometry.}
Human-use terms are treated as standardized geometric proxies for practical interaction rather than as universal ergonomic standards. Their purpose is to provide consistent benchmark-wide tests of whether controls and usable regions occupy plausible interaction spaces. For reachability, let $d$ denote the Euclidean distance between the relevant control and the estimated eye position. For seated objects, the eye point is placed $0.60$~m above the seat top; otherwise, a standing eye point is placed $1.40$~m above ground. We define

$$
s_{\mathrm{reach}}
=
\max
\left(
0,
1-
\frac{
\max(0,d-0.70)
}{
1.4
}
\right).
$$

Controls within $0.70$~m therefore receive full credit, after which the score decreases continuously with distance rather than introducing a brittle binary reachability boundary. Clearance is evaluated using task-specific median human occupancy boxes sampled on a $10^3$ point grid. Let $f_{\mathrm{free}}$ denote the fraction of the corresponding occupancy volume that remains unobstructed. We define

$$
s_{\mathrm{clear}}
=
\min
\left(
1,
\frac{
f_{\mathrm{free}}
}{
0.9
}
\right).
$$

Thus, clearance is evaluated volumetrically rather than through isolated collision tests. A component expected to be graspable receives full grip credit when its two smallest principal extents lie between $15$~mm and $70$~mm. A required usable surface receives full credit when its area is at least

$$
0.05~\mathrm{m}^2
$$

and, where a task-specific expected height interval is available, the surface lies within the corresponding band. For components expected to be spatially co-located, let $g$ denote their geometric gap. We define

$$
s_{\mathrm{coloc}}
=
\operatorname{clip}
\left(
1-
\frac{
g-0.10
}{
0.20
},
0,1
\right).
$$

Because these human-use rules are intentionally general geometric proxies, we avoid imposing them when they are inconsistent with the task itself. A human-geometry link is disabled whenever the corresponding benchmark reference receives a score below $0.5$ on that criterion. For tasks without reference-based human calibration, such links are disabled except for ground contact. This reference gating prevents a generic anthropometric heuristic from penalizing generated designs for failing a condition that is not satisfied by the corresponding real reference.

\paragraph{Dynamic motion feasibility.}
Motion capabilities are evaluated through rigid-body simulation in MuJoCo. The simulation is intended as a standardized test of geometric and articulated feasibility rather than a high-fidelity reconstruction of material-specific real-world dynamics. All systems are evaluated under the same simulation assumptions.

Simulation uses a $1$~ms timestep and the implicit-fast integrator. Parts connected by physical contact or fixed joints are grouped into rigid bodies and represented using convex-hull collision geometry. A grounded wheel attached through an approximately horizontal and centered hinge is instead represented using cylindrical collision geometry to better approximate rolling behavior. Rigid-body masses are constructed from the S.3 geometric volumes using the common density

$$
1000~\mathrm{kg/m^3},
$$

and friction coefficients are fixed to

$$
(1,\;0.02,\;0.002).
$$

The uniform density is used as a standardized inertial proxy rather than an estimate of the true material density of each generated component. This isolates motion feasibility from errors in material prediction and provides comparable simulation conditions across generated artifacts. Before commanded actuation, the completed artifact is allowed to settle for $0.6$~s. If the object tilts by at least $20^\circ$ or translates by at least $0.3$ times its object diagonal during this period, all motion capabilities receive zero. Such motion indicates that the completed artifact is insufficiently stable for subsequent actuation to provide a meaningful test of intended functionality.

For articulated motion claims, candidate joints are driven by a proportional-derivative servo toward their declared limits. We use the standardized mass- and scale-dependent proportional gain

$$
k_p
=
100MgL,
$$

where $M$ and $L$ denote the moving mass and characteristic length, respectively. This scaling provides comparable actuation strength across structures of different sizes; it is used as a simulation control heuristic rather than as a model of a specific physical actuator. When a generated joint does not provide an explicit motion limit, the target is defined as $\pm90^\circ$ for a revolute joint, $\pm360^\circ$ for a continuous joint, or $\pm$ half of the corresponding moving-part extent for a prismatic joint. Actuation uses a $1.5$~s ramp followed by a $0.4$~s holding period.

Let $\Delta q$ denote the achieved generalized displacement and $q_{\mathrm{target}}$ the commanded displacement. We define

$$
s_{\mathrm{motion}}
=
\min
\left(
1,
\frac{
|\Delta q|
}{
|q_{\mathrm{target}}|
}
\right)
\mathbf{1}
[
\mathrm{tilt}<15^\circ
].
$$

The displacement ratio measures how much of the commanded motion is physically realized. The additional tilt condition prevents a mechanism from receiving motion credit when the apparent displacement is achieved by destabilizing the entire artifact rather than by successful articulation. The actuation stability threshold is intentionally stricter than the initial settling threshold because successful operation requires the artifact to remain stable while the mechanism is being used. The better of the two actuation directions is retained. For each claimed motion, we consider up to the first four candidate joints on the structural path between the moving component and the corresponding relative component.

\paragraph{Rolling capabilities.}
Rolling capabilities associated with wheels, casters, or tracks are tested separately because the existence of a rotational joint alone does not establish that the completed object can translate through rolling. The corresponding wheel joints are left free and a horizontal force equal to $10\%$ of the object's weight is applied for $1.5$~s. If $d_{\mathrm{travel}}$ denotes the resulting translation, rolling receives

$$
s_{\mathrm{roll}}
=
\min
\left(
1,
\frac{
d_{\mathrm{travel}}
}{
0.276~\mathrm{m}
}
\right)
\mathbf{1}
[
\mathrm{tilt}<15^\circ
].
$$

The fixed forcing and normalization conditions provide a common benchmark-wide test of whether the assembled wheel or track mechanism can produce meaningful translation without destabilizing the object.

\paragraph{Capability aggregation.}
Each task-level capability $c$ consists of a set of active requirement links $\mathcal{L}_c$. Because these links represent jointly necessary conditions for the capability to be usable, capability performance is defined by its weakest active requirement:

$$
w_c
=
\min_{\ell\in\mathcal{L}_c}
s_\ell.
$$

The weakest-link formulation is intentional. Averaging the requirements could assign substantial credit to a capability whose components are present but disconnected, whose mechanism cannot move, or whose control cannot be reached. In each of these cases, the capability as a whole is not operational despite success on its remaining conditions. Let $\mathcal{C}$ denote the set of applicable task-level capabilities. The final Operability score is

$$
R_{4.3}
=
\frac{1}{|\mathcal{C}|}
\sum_{c\in\mathcal{C}}
w_c.
$$

R.3 therefore evaluates functionality compositionally rather than through isolated part or joint checks. It asks whether the generated artifact integrates the required components, structural relationships, articulated mechanisms, and human-use geometry into capabilities that can plausibly function as intended. The geometric and simulation thresholds used in these tests are fixed benchmark-wide evaluation parameters and are applied identically to all systems; they are intended as consistent comparative proxies for operability rather than claims of exact real-world physical behavior.

\end{Metrics_4}

{
\renewcommand{\arraystretch}{1.12}
\begin{table*}[t]
\centering
\caption{Level scores, computed as the mean of the three metrics within each level with N/A metrics excluded, are reported on the full 200-object core set and on two disjoint stratified 50-object samples. All three columns are derived from the same evaluation episodes. Their close agreement across subsets demonstrates strong consistency across object types, suggesting that LMBuild primarily captures intrinsic differences in model capability rather than artifacts of a particular object sample.}
\label{tab:resample_results}
\resizebox{\textwidth}{!}{
\begin{tabular}{l|cccc|cccc|cccc}
\toprule
\multirow{2}{*}{\textbf{Models}} & \multicolumn{4}{c|}{\textbf{Full core (200)}} & \multicolumn{4}{c|}{\textbf{Sample A (50)}} & \multicolumn{4}{c}{\textbf{Sample B (50)}} \\
\cmidrule(lr){2-5}\cmidrule(lr){6-9}\cmidrule(lr){10-13}
 & \textbf{S} & \textbf{A} & \textbf{D} & \textbf{R} & \textbf{S} & \textbf{A} & \textbf{D} & \textbf{R} & \textbf{S} & \textbf{A} & \textbf{D} & \textbf{R} \\
\midrule
\rowcolor{gray!20}
\multicolumn{13}{c}{\textbf{State-of-the-art Closed-Source APIs}} \\
GPT-6 Astra & \score{88.5} & \score{50.0} & \score{74.0} & \score{53.5} & \score{86.4} & \score{50.6} & \score{72.5} & \score{55.7} & \score{91.5} & \score{48.2} & \score{73.1} & \score{52.0} \\
GPT-5.6 Sol & \score{88.8} & \score{41.9} & \score{71.5} & \score{57.4} & \score{88.0} & \score{43.8} & \score{70.4} & \score{56.8} & \score{89.5} & \score{41.1} & \score{72.9} & \score{57.1} \\
Claude Opus 5 & \score{88.8} & \score{46.6} & \score{71.3} & \score{55.1} & \score{88.0} & \score{48.9} & \score{71.5} & \score{57.7} & \score{87.4} & \score{43.3} & \score{72.6} & \score{50.9} \\
Claude Fable 5.1 & \score{89.1} & \score{47.3} & \score{71.8} & \score{55.3} & \score{86.1} & \score{49.7} & \score{70.9} & \score{56.0} & \score{85.7} & \score{42.7} & \score{72.5} & \score{53.7} \\
Claude Sonnet 5 & \score{86.1} & \score{41.5} & \score{66.3} & \score{56.2} & \score{86.0} & \score{43.3} & \score{66.1} & \score{58.7} & \score{86.8} & \score{37.5} & \score{66.9} & \score{53.7} \\
Claude Haiku 4.5 & \score{80.8} & \score{30.5} & \score{57.8} & \score{51.3} & \score{82.0} & \score{29.2} & \score{57.9} & \score{50.3} & \score{81.1} & \score{29.8} & \score{58.6} & \score{51.1} \\
\midrule
\rowcolor{gray!20}
\multicolumn{13}{c}{\textbf{Open-Source (Multi-modal) Large Language Models}} \\
Qwen3.5-27B & \score{82.9} & \score{28.3} & \score{52.7} & \score{57.2} & \score{84.6} & \score{28.2} & \score{53.1} & \score{60.1} & \score{84.2} & \score{27.3} & \score{53.2} & \score{55.2} \\
Qwen3.5-35B-A3B & \score{72.1} & \score{24.0} & \score{47.0} & \score{44.9} & \score{69.2} & \score{23.6} & \score{45.8} & \score{44.1} & \score{78.7} & \score{25.4} & \score{48.9} & \score{47.4} \\
GPT-OSS-120B & \score{78.5} & \score{25.8} & \score{49.3} & \score{51.7} & \score{79.6} & \score{25.4} & \score{48.8} & \score{51.1} & \score{76.4} & \score{24.0} & \score{49.8} & \score{50.7} \\
Gemma-4-31B-IT & \score{59.7} & \score{20.1} & \score{41.2} & \score{36.7} & \score{61.6} & \score{21.1} & \score{43.2} & \score{34.3} & \score{63.5} & \score{22.2} & \score{43.1} & \score{39.7} \\
Ministral-3-8B & \score{65.7} & \score{18.8} & \score{46.8} & \score{39.9} & \score{67.5} & \score{18.9} & \score{46.7} & \score{40.1} & \score{67.3} & \score{19.5} & \score{49.6} & \score{37.0} \\
InternVL3.5-38B & \score{49.3} & \score{11.4} & \score{34.8} & \score{21.5} & \score{50.2} & \score{9.6} & \score{33.8} & \score{17.2} & \score{45.7} & \score{12.9} & \score{36.7} & \score{22.6} \\
Qwen3-VL-30B-A3B & \score{48.7} & \score{13.8} & \score{32.1} & \score{33.0} & \score{51.9} & \score{15.0} & \score{36.9} & \score{37.1} & \score{50.2} & \score{13.0} & \score{35.0} & \score{30.2} \\
MiniCPM-V-4.5 & \score{40.0} & \score{9.5} & \score{27.1} & \score{18.6} & \score{45.2} & \score{10.4} & \score{27.8} & \score{19.6} & \score{38.7} & \score{8.9} & \score{25.1} & \score{19.5} \\
Qwen3-4B-Instruct & \score{65.5} & \score{15.6} & \score{30.3} & \score{38.8} & \score{67.8} & \score{13.8} & \score{30.3} & \score{36.6} & \score{66.1} & \score{18.5} & \score{34.0} & \score{40.7} \\
Qwen3-VL-32B & \score{48.5} & \score{13.5} & \score{30.2} & \score{31.8} & \score{43.0} & \score{11.6} & \score{27.3} & \score{27.9} & \score{49.6} & \score{14.3} & \score{32.8} & \score{31.1} \\
ERNIE-4.5-VL-28B-A3B & \score{20.0} & \score{4.4} & \score{15.0} & \score{9.6} & \score{22.7} & \score{4.9} & \score{15.0} & \score{8.1} & \score{20.6} & \score{6.3} & \score{17.2} & \score{10.4} \\
InternVL3.5-8B & \score{39.1} & \score{7.4} & \score{26.2} & \score{18.1} & \score{36.1} & \score{6.5} & \score{22.3} & \score{17.2} & \score{45.2} & \score{9.0} & \score{29.0} & \score{18.2} \\
Qwen3-VL-8B & \score{27.9} & \score{6.9} & \score{15.1} & \score{17.8} & \score{31.8} & \score{8.9} & \score{18.0} & \score{22.8} & \score{19.5} & \score{4.0} & \score{10.6} & \score{11.3} \\
\midrule
\rowcolor{gray!20}
\multicolumn{13}{c}{\textbf{State-of-the-art Domain-Specific Models}} \\
BrickGPT & \score{94.7} & \score{34.5} & \score{24.2} & \scoreNA & \score{94.3} & \score{34.5} & \score{24.2} & \scoreNA & \score{94.9} & \score{40.0} & \score{23.4} & \scoreNA \\
LegoACE & \score{78.0} & \score{29.6} & \score{27.7} & \scoreNA & \score{79.1} & \score{28.8} & \score{28.4} & \scoreNA & \score{81.6} & \score{31.5} & \score{25.9} & \scoreNA \\
PartCrafter-15 & \score{61.3} & \score{32.2} & \score{57.5} & \scoreNA & \score{62.6} & \score{33.1} & \score{57.1} & \scoreNA & \score{65.6} & \score{36.5} & \score{60.4} & \scoreNA \\
PartCrafter-8 & \score{57.1} & \score{28.2} & \score{53.0} & \scoreNA & \score{58.1} & \score{27.7} & \score{52.8} & \scoreNA & \score{60.5} & \score{32.8} & \score{56.4} & \scoreNA \\
PartPacker & \score{75.1} & \score{36.4} & \score{61.3} & \scoreNA & \score{75.7} & \score{34.9} & \score{61.3} & \scoreNA & \score{79.7} & \score{40.8} & \score{63.3} & \scoreNA \\
Cube3D+CubePart & \score{69.4} & \score{28.7} & \score{56.5} & \score{13.4} & \score{70.4} & \score{29.9} & \score{55.2} & \score{16.6} & \score{70.8} & \score{30.4} & \score{58.2} & \score{13.5} \\
PartCrafter15$^{\dagger}$(P) & \score{61.2} & \score{18.0} & \score{37.4} & \scoreNA & \score{61.8} & \score{18.4} & \score{37.7} & \scoreNA & \score{65.8} & \score{19.9} & \score{40.1} & \scoreNA \\
PartCrafter8$^{\dagger}$(P) & \score{60.2} & \score{17.3} & \score{43.1} & \scoreNA & \score{61.8} & \score{17.5} & \score{42.9} & \scoreNA & \score{63.5} & \score{18.4} & \score{44.3} & \scoreNA \\
PartPacker$^{\dagger}$(P) & \score{77.9} & \score{21.7} & \score{55.0} & \scoreNA & \score{77.0} & \score{20.7} & \score{55.3} & \scoreNA & \score{81.1} & \score{24.7} & \score{54.3} & \scoreNA \\
Cube3D$^{\dagger}$(P) & \score{79.8} & \score{17.6} & \score{61.0} & \scoreNA & \score{82.7} & \score{19.6} & \score{61.5} & \scoreNA & \score{81.7} & \score{19.1} & \score{62.6} & \scoreNA \\
PhysX-Anything & \score{63.7} & \score{11.3} & \score{44.7} & \score{31.6} & \score{67.1} & \score{11.7} & \score{43.8} & \score{34.3} & \score{62.8} & \score{10.6} & \score{46.1} & \score{30.5} \\
\bottomrule
\end{tabular}
}
\end{table*}
}

\begin{table}[t]
\centering
\caption{Agreement of system-level scores across task sets: Pearson $r$ (Spearman $\rho$ in parentheses) over the 30 systems' means. A vs.\ B compares two disjoint 50-object samples; the last two columns repeat the stratified draw 1{,}000 times (mean, 95\% interval in brackets). Sample-vs-full correlations are an upper bound because each sample is part of the full set.}
\label{tab:resample_corr}
\resizebox{\columnwidth}{!}{
\begin{tabular}{l|ccc|cc}
\toprule
\multirow{2}{*}{\textbf{Metric}} & \multirow{2}{*}{\textbf{A vs.\ 200}} & \multirow{2}{*}{\textbf{B vs.\ 200}} & \multirow{2}{*}{\textbf{A vs.\ B}} & \multicolumn{2}{c}{\textbf{1{,}000 resamples}} \\
\cmidrule(lr){5-6}
 & & & & \textbf{50 vs.\ 200} & \textbf{disjoint pairs} \\
\midrule
S.1 & 0.982 (0.97) & 0.986 (0.97) & 0.960 (0.95) & 0.987 [0.98, 0.99] & 0.967 [0.94, 0.98] \\
S.2 & 0.990 (0.96) & 0.995 (0.98) & 0.981 (0.96) & 0.989 [0.98, 1.00] & 0.972 [0.95, 0.99] \\
S.3 & 0.991 (0.99) & 0.978 (0.97) & 0.960 (0.96) & 0.983 [0.97, 0.99] & 0.957 [0.93, 0.98] \\
\midrule
A.1 & 0.994 (0.99) & 0.981 (0.98) & 0.967 (0.96) & 0.990 [0.98, 1.00] & 0.975 [0.95, 0.99] \\
A.2 & 0.991 (0.98) & 0.986 (0.99) & 0.966 (0.95) & 0.991 [0.98, 1.00] & 0.977 [0.96, 0.99] \\
A.3 & 0.992 (0.94) & 0.995 (0.98) & 0.987 (0.92) & 0.992 [0.99, 1.00] & 0.980 [0.96, 0.99] \\
\midrule
D.1 & 0.992 (0.99) & 0.995 (0.99) & 0.988 (0.99) & 0.991 [0.98, 1.00] & 0.977 [0.96, 0.99] \\
D.2 & 0.993 (0.98) & 0.989 (0.98) & 0.975 (0.97) & 0.988 [0.98, 0.99] & 0.970 [0.95, 0.98] \\
D.3 & 0.994 (1.00) & 0.990 (0.98) & 0.979 (0.97) & 0.990 [0.98, 1.00] & 0.975 [0.96, 0.99] \\
\midrule
R.1 & 0.984 (0.96) & 0.988 (0.98) & 0.959 (0.94) & 0.989 [0.98, 1.00] & 0.970 [0.94, 0.99] \\
R.2 & 0.982 (0.95) & 0.979 (0.95) & 0.937 (0.86) & 0.982 [0.96, 0.99] & 0.953 [0.91, 0.98] \\
R.3 & 0.987 (0.95) & 0.994 (0.98) & 0.980 (0.93) & 0.987 [0.98, 0.99] & 0.967 [0.94, 0.98] \\
\bottomrule
\end{tabular}
}
\end{table}

\begin{table}[t]
\centering
\caption{Failure modes of the six closed-source APIs on the 200-object core set (\% of tasks; thresholds in parentheses). All models submit a design on every task.}
\label{tab:sota_errors}
\resizebox{\columnwidth}{!}{
\begin{tabular}{ll|cccccc}
\toprule
\textbf{Level} & \textbf{Failure mode} & \textbf{GPT-6} & \textbf{GPT-5.6} & \textbf{Opus 5} & \textbf{Fable 5.1} & \textbf{Sonnet 5} & \textbf{Haiku 4.5} \\
\midrule
Soundness & Disconnected assembly (S.1) & 21.0 & 13.0 & 6.5 & 6.5 & 19.5 & 38.0 \\
 & Interpenetrating parts (S.2) & 13.0 & 16.5 & 4.0 & 3.0 & 15.5 & 38.5 \\
 & Unstable under gravity (S.3 $<$ 0.5) & 8.5 & 10.5 & 10.0 & 6.5 & 12.5 & 24.0 \\
\midrule
Affordance & Required parts missing or misplaced (A.2 $<$ 0.5) & 54.5 & 68.5 & 59.0 & 56.0 & 68.5 & 79.0 \\
 & A required moving part is never built (A.3) & 56.5 & 61.5 & 51.0 & 49.5 & 57.5 & 65.5 \\
 & Part built but no joint moves it (A.3) & 24.5 & 27.5 & 17.5 & 30.5 & 29.0 & 34.5 \\
 & Joint present but wrong type/axis/range (A.3) & 36.0 & 38.0 & 44.0 & 39.5 & 33.5 & 37.0 \\
\midrule
Design & Part count off by $>2\times$ vs.\ reference (D.1) & 23.0 & 31.0 & 31.0 & 28.5 & 31.0 & 37.5 \\
 & Low fidelity to the photo (D.3 $<$ 0.4) & 1.0 & 8.5 & 5.0 & 4.5 & 13.0 & 19.5 \\
\midrule
Realization & Assembly sequence not fully executable (R.1) & 60.0 & 51.0 & 67.5 & 50.5 & 48.0 & 57.5 \\
 & No functional capability works (R.3) & 35.5 & 34.0 & 30.0 & 32.5 & 40.0 & 51.0 \\
\midrule
Tool use & At least one failed \texttt{create\_part} call & 15.5 & 16.5 & 27.5 & 8.5 & 11.5 & 10.5 \\
\bottomrule
\end{tabular}
}
\end{table}

\subsection{Interpretation of Evaluation Metrics and Results}
\label{app:interpretation}

\paragraph{Stable Metrics Across Object Sub-samples.}
Because several of our evaluation metrics involve design choices and hyperparameters, we further examine whether they capture underlying system capabilities rather than being overly sensitive to the particular objects included in the benchmark. To validate this, we draw two disjoint stratified 50-object samples from the 200-object core set and compare the resulting scores with each other and with those obtained on the full set. Table~\ref{tab:resample_results} reports the level scores for the full core set and the two sub-samples. Aggregate performance remains highly stable: closed-source APIs score 63.4 on the full set, 63.6 on sample~A, and 62.5 on sample~B, while open-source models score 33.9, 34.1, and 34.4, respectively. At the individual-system level, the mean absolute deviation from the full-set score is only 1.0 point for sample~A and 1.7 points for sample~B. Table~\ref{tab:resample_corr} further shows that this consistency holds at the metric level. Across the 30 systems, the two disjoint samples achieve Pearson correlations of $r=0.937$--$0.988$ and Spearman correlations of $\rho=0.86$--$0.99$. Across 1{,}000 repeated stratified resamplings, the mean correlation between disjoint sample pairs remains between 0.953 and 0.980, with all 95\% intervals at or above 0.91. These results indicate that the evaluation metrics produce highly consistent relative system performance across substantially different object subsets, suggesting that LMBuild primarily measures persistent differences in model capability rather than artifacts of a particular object sample. The full 200-object evaluation therefore provides the most stable estimate of overall system performance. The full results in Table~\ref{tab:main_results} also reveal clear differences across system families. Closed-source APIs achieve the strongest overall performance across the four levels: GPT-6 Astra reaches 88.5 / 50.0 / 74.0 / 53.5 on Soundness / Affordance / Design / Realization, with Claude Opus~5 and Claude Fable~5.1 following closely. The strongest open model, Qwen3.5-27B, is competitive on Soundness and Realization (82.9 and 57.2) but remains substantially lower on Affordance (28.3) and Design (52.7). Domain-specific generators show the widest variation across dimensions, performing well when their output representation directly enforces particular structural properties but less consistently on tasks that require richer part decomposition and functional behavior.

\paragraph{Interpreting the Scores Where a Closed-source API Does Not Lead.}
For a small number of metrics, we observe that another system matches or exceeds GPT-6 Astra. These isolated results, however, should not be interpreted as evidence that the corresponding system is a stronger general-purpose builder. Each metric captures only one aspect of a generated structure, and high performance on a particular dimension may arise from properties enforced by the system's representation, from producing simpler structures, or from optimizing visual appearance at the expense of functionality. We therefore examine these cases individually and explain why such scores can be misleading when interpreted in isolation. 

\textbf{Soundness of Brick Generators.} BrickGPT and LegoACE achieve the strongest results on the connectivity- and collision-related Soundness metrics, outperforming all API-based systems on these dimensions. This advantage is largely explained by their representations and decoding procedures rather than by broader design capability. Both systems generate structures on a discrete stud lattice, and BrickGPT further rejects placements that would collide with existing bricks and rolls back unstable construction steps. As a result, connectivity and collision avoidance are strongly enforced by construction. However, the same systems perform substantially worse on metrics that assess whether the generated structure corresponds to the requested object and contains the appropriate functional components. Their Design and Affordance scores remain comparatively low, indicating that a structure can be physically coherent without faithfully realizing the intended object or its functionality. These results therefore illustrate that high Soundness is necessary, but not sufficient, for successful structure generation, and it is not suprising that several systems could 'outperform' the best LLM-based solutions.

\textbf{Soundness of Part-based Mesh Generators.}
A related effect appears among part-based mesh generators. Particulate and Cube3D+CubePart achieve strong scores on selected Soundness metrics because their generation procedures naturally encourage geometric compatibility between parts. Particulate first generates a complete mesh and subsequently segments it into components, meaning that the resulting parts inherit a shared surface and are unlikely to interpenetrate. CubePart similarly generates parts within a common object volume, which encourages contact between adjacent components. This geometric consistency, however, does not necessarily imply structural robustness. For example, PartCrafter performs substantially worse on load-bearing stability, suggesting that its components are not always connected in mechanically meaningful ways. These results distinguish geometric compatibility from structural reasoning: parts may fit together spatially without being designed to support one another under physical constraints.

\textbf{Soundness and realization of simpler designs.}
Several open and smaller general-purpose models also match or outperform GPT-6 Astra on selected Soundness and Realization metrics, particularly those related to supported connections, assembly executability, and material plausibility. A key factor is that these systems tend to produce considerably smaller assemblies. For example, Qwen3.5-27B and GPT-OSS-120B generate designs with substantially fewer parts than GPT-6 Astra and Claude Opus~5. This difference directly affects metrics such as S.1, R.1, and R.2, all of which become easier to satisfy as assembly complexity decreases. Fewer parts imply fewer interfaces at which unsupported or incorrect connections can occur, shorter assembly sequences reduce the number of ordering dependencies, and a smaller set of generic components simplifies material assignment. The corresponding tradeoff appears on metrics that evaluate completeness, decomposition quality, and functional structure. Systems that produce simpler assemblies tend to omit required components and mechanisms more frequently, resulting in lower Affordance and Design-fidelity scores. This pattern suggests that strong performance on local structural or realization criteria can partly reflect reduced design complexity rather than stronger end-to-end generation capability. Moreover, many failures captured by these metrics are locally repairable: a geometric gap can often be closed, a fastener added, or a material reassigned without changing the overall design. In contrast, a missing mechanism, incomplete decomposition, or incorrect functional structure usually requires more fundamental redesign. We therefore interpret Soundness and low-level Realization metrics as important necessary conditions, while Affordance, Design fidelity, and Operability provide stronger evidence of whether a model can generate the object as a complete functional system.

\textbf{Effective-part generation of Cube3D+CubePart.}
Cube3D+CubePart performs notably better than other mesh-based generators on effective-part generation and also surpasses several open general-purpose models. This result should be interpreted in the context of its input representation. CubePart receives an explicit part schema, and in our evaluation protocol this schema corresponds to the set of components required by the task. The system therefore begins with substantial information about which parts should be present, rather than having to infer the decomposition entirely from the conditioning input. Our metric still requires those components to occupy appropriate spatial locations, so the system does not receive full credit merely for reproducing the supplied labels. Nevertheless, part of its advantage reflects access to the expected component inventory. Its simulation results further show that the presence of correctly named parts does not necessarily translate into functional articulation, as the generated components rarely produce meaningful motion.

\textbf{Aesthetic quality of mesh generators.}
Mesh-based generators show a different form of specialization on the aesthetic metric. Cube3D and related systems produce smooth, coherent, and visually refined surfaces and consequently perform very well on reference-free aesthetic evaluation. This behavior is consistent with their training on large collections of 3D assets, where generating a polished and watertight surface is a central objective. However, aesthetic quality alone does not establish that the generated object matches the requested instance. A text-conditioned system may generate a visually convincing generic member of a category while missing the specific structural configuration required by the task. This distinction is reflected in the gap between reference-free aesthetics and reference-conditioned Design fidelity. Conversely, image-to-3D reconstruction methods such as PartPacker achieve stronger visual fidelity to the target object but still lack the articulation and functional decomposition required for high Affordance and Operability. Aesthetic quality should therefore be interpreted together with fidelity and functionality rather than as an independent indicator of overall design capability.

\textbf{Material generation of PhysX-Anything.}
PhysX-Anything is distinctive among the specialized generators because it explicitly predicts a material for each generated component. As a result, it achieves meaningful performance on material plausibility, unlike most domain-specific baselines. This result demonstrates a useful capability, but it should not be conflated with end-to-end product realizability. The system still produces relatively coarse part decompositions, and its generated objects exhibit almost no successful functional operation in simulation. Material plausibility therefore captures one aspect of realization quality, but does not by itself indicate that the generated structure is complete, mechanically functional, or manufacturable as a coherent product.

\textbf{Overall interpretation.}
Taken together, these cases reveal three recurring reasons why a specialized or open system may match or exceed the strongest closed-source API on an individual metric. First, some properties are guaranteed or strongly encouraged by the output representation itself, as in lattice-based brick generation or mesh segmentation. Second, some metrics become easier to satisfy when a system generates smaller and less complex assemblies. Third, systems optimized for visual quality can perform well on appearance-based criteria without necessarily reproducing the correct object structure or functionality. These observations motivate interpreting LMBuild as a multidimensional benchmark rather than comparing systems through isolated metric maxima. Soundness, material plausibility, and aesthetic quality capture important properties of a generated structure, but none alone establishes that the system has identified the correct components, organized them into the intended design, and produced the required functionality. Across the metrics that more directly require these capabilities, including effective-part generation, functional decomposition, reference fidelity, and operabilitym the closed-source APIs remain strongest overall. The consistency of the level-level results across object sub-samples further suggests that these differences reflect persistent variation in end to end model capability rather than advantages tied to a small set of objects or individual metrics.

\paragraph{Errors of state-of-the-art models.}
Table~\ref{tab:sota_errors} summarizes the dominant failure modes of the six closed-source APIs. All systems produce a design for every task, indicating that their primary limitations are functional rather than procedural. Physical defects do occur, but they are generally secondary: between 6.5\% and 21\% of generated designs contain at least one disconnected part, while 3\%--16.5\% contain interpenetrating parts, with Claude Haiku~4.5 representing a notable outlier at 38\% and 38.5\%, respectively. Most of these defects are small gaps or overlaps, particularly on LEGO and product-CAD objects, and could plausibly be corrected through local geometric adjustments. The more consequential failures arise from incomplete functional decomposition. A.3 is the weakest Affordance metric across these systems, with an average score of 37.9. Among failed motion targets, 57\% fail because the required moving component, such as a lid, steering link, or handle, is never generated; a further 17\% include the relevant part but lack a joint capable of moving it. The remaining failures are associated with incorrect joint type, axis, range, or offset. A similar pattern appears in A.2: in 54.5\%--68.5\% of designs, at least half of the required components are missing or incorrectly positioned, with many omissions corresponding to internal parts that are not directly visible from the exterior. These decomposition failures propagate directly to functional execution. R.3, which evaluates each declared capability according to its weakest required element, averages only 18.3 across the six systems. Nearly every design omits at least one component referenced by its functional description, such as a timer, thermal sensor, or lever. As a result, no evaluated capability succeeds at all on roughly 30\%--40\% of tasks, increasing to 51\% for Claude Haiku~4.5. Assembly planning introduces an additional source of error: between 48\% and 68\% of generated assembly sequences are not fully executable. The first failure is often a floating component, such as a main body placed before the legs or supports required to hold it, even when the final geometry could in principle be assembled under a different ordering. In contrast, low-level tool interaction is rarely the limiting factor. Only 2\%--7\% of \texttt{create\_part} calls fail, and the models typically recover from these failures within the same episode. Performance also varies systematically across object sources. Product-level CAD and Fusion~360 objects are the most difficult, with 81\% and 73\% of designs, respectively, missing or misplacing required components, compared with 48\% for Artiverse furniture and appliances. Qualitative examples reinforce the same pattern: Claude Opus~5 can generate a toaster with a recognizable bread slot while omitting the lever, timer, thermal sensor, and ejection mechanism; it can produce a toilet with a plausible seat hinge but no lid or flush handle; and GPT-6 Astra can generate an off-road vehicle with rotating wheels while failing to provide articulated steering links. Taken together, these results suggest that current frontier models are comparatively effective at producing recognizable external geometry, but remain substantially weaker at recovering the internal component structure, articulation, and mechanisms that make an object function. This gap is precisely what the Affordance and Realization levels of LMBuild are intended to capture.

\subsection{Limitations and Future Work}
\label{Limitation}

\paragraph{Evaluation scope and interpretation.} LMBuild is designed as a standardized framework for comparing generated structures, rather than as a certification procedure for real-world engineering readiness. Its metrics operationalize buildability and functionality through reproducible geometric, semantic, kinematic, and simulation-based tests. These tests provide substantially stronger evidence than visual plausibility alone, but they do not establish that a generated artifact can be fabricated, assembled, or operated safely in every physical setting. In particular, the current evaluation does not model manufacturing tolerances, fastener strength, elastic deformation, fracture, fatigue, tool access, wiring or fluid routing, actuator limits, or sustained loading. LMBuild scores should therefore be interpreted as measures of performance under fixed benchmark criteria and as comparative evidence of physical and functional plausibility, not as proof of deployability or safety.

\textbf{Approximate physical models.} Several evaluation components necessarily use simplified physical assumptions. Stability focuses on  tipping; assembly feasibility uses discretized insertion paths and intermediate-support checks; and operability uses rigid-body simulation with standardized density, friction, collision geometry, and actuation parameters. These choices make evaluation deterministic and comparable across heterogeneous structures, while isolating geometric and kinematic feasibility from uncertainty in inferred physical properties. Nevertheless, different parameterizations or higher-fidelity models may change individual scores, especially for structures whose operation depends on deformable components, material-specific behavior, or complex contact. Future work should study metric sensitivity and ranking stability under alternative thresholds and simulation parameters, and validate a representative subset through physical fabrication, robotic assembly, or expert mechanical assessment.
Benchmark coverage. The experiments in this paper use the 50-object LMBuild-Core split. This split is deliberately coverage-oriented: it spans multiple structural families, articulation regimes, construction paradigms, and difficulty levels rather than reproducing the empirical frequency distribution of any one source dataset. Consequently, the reported results characterize performance across the selected construction regimes but should not be interpreted as population-level estimates over all real-world objects. The 2,399-object LMBuild-Full split is intended to support broader evaluation, including category-level analysis, rare mechanisms, and more difficult product-scale assemblies. Extending evaluation to the full split and to additional domains such as compliant mechanisms, electronics, and fluid-powered systems is an important direction.

\textbf{Automated evaluation and human calibration.} Design Quality combines deterministic measurements with two VLM judges calibrated against ratings from four human annotators. On held-out designs, the resulting metrics preserve system-level human rankings well, while item-level correlations and human-human agreement show that individual designs remain inherently more ambiguous. The present calibration covers a limited set of objects and systems and may not capture every valid design style or decomposition strategy. In addition, Wikipedia-derived capability sheets provide an auditable source of real-world knowledge but may omit uncommon components or alternative realizations. Future versions should broaden the annotator and system pool, include domain experts where appropriate, report uncertainty for agreement statistics, and continue expanding accepted component synonyms, material alternatives, and functional realizations.

\textbf{System comparability and statistical uncertainty.} General-purpose agents receive the same task specification, environment semantics, tool definitions, and budgets, but provider-specific execution interfaces and modality support necessarily differ. Domain-specific systems additionally retain their native inputs and outputs and are therefore comparable only on shared applicable metrics. Unless otherwise stated, the current results use one final generation trajectory per model-task pair. Confidence intervals obtained by resampling benchmark objects quantify variation across tasks, but not within-model stochasticity across repeated generations. Future evaluation should include repeated runs where computationally feasible, standardized native tool-calling interfaces, and reporting of both task-level and run-level uncertainty.

\textbf{Outlook.}
Beyond scaling models and tool budgets, future work should improve the knowledge, skills, and creative physical reasoning that models bring to structure generation. First, models could internalize reusable construction knowledge, including geometric constraints, assembly rules, component affordances, and functional properties, rather than rediscovering them at inference time. Recent work on textual and multimodal policy internalization suggests that such structured knowledge can be incorporated more directly into model behavior~\citep{liu2026internalizing,wang2026multimodalpolicy}. Similar methods could provide stronger priors for joints, interfaces, load-bearing structures, clearances, and other recurring physical constraints. Second, building agents could acquire reusable procedural skills through exploration and interaction. Work on environment exploration, recursive self-improvement, and reasoning skills shows that agents can discover and retain useful strategies from experience~\citep{liu2026osexpert,zhu2026rsiagent,zhao2026reasoningskills}. In LMBuild, agents could similarly explore alternative construction strategies, learn from failed checks or simulations, and accumulate reusable skills for creating parts, joints, and mechanisms. Third, future systems should support more creative physical construction, including inventing functional parts, repurposing components, and refining designs through multi-turn interaction. Recent work highlights the difficulty of affordance-based creative reasoning in multimodal models~\citep{qian2026creativitybench,qian2026creativephysical}. Combining such reasoning with iterative tool use, execution feedback, and code-based construction interfaces could enable agents to propose, test, and revise increasingly functional designs~\citep{wang2024mint,yang2024wizardcode}. A complementary direction is to close the loop between digital generation and physical fabrication. Future extensions could incorporate additive-manufacturing constraints such as minimum feature size, wall thickness, overhangs, tolerances, material choice, and build volume, and evaluate whether generated structures can be successfully printed, assembled, and operated. Physical validation would enable direct measurement of print success, dimensional accuracy, assembly feasibility, structural integrity, and functional performance, while supporting sim-to-real calibration of LMBuild's evaluation metrics. Beyond 3D printing, future work could consider additional manufacturing processes, richer material and contact mechanics, fasteners, compliant components, and electrical or fluid routing. Expanding evaluation to LMBuild-Full, repeated generations, and evaluator-sensitivity analysis would further improve coverage and reliability. Ultimately, LMBuild could serve not only as a benchmark for buildable generation, but also as an environment for studying how agents acquire the knowledge, skills, and creative physical intelligence required to construct functional real-world artifacts.

\end{document}